\documentclass[10pt,journal,compsoc]{IEEEtran}

\usepackage{marginnote}

\usepackage{amsmath,graphicx}
\usepackage[american]{babel}
\usepackage{makeidx}
\usepackage{color}
\usepackage{lineno,hyperref}
\usepackage{amssymb,amsmath,amsthm}
\usepackage{multirow}
\usepackage{bm}
\usepackage{subfigure}
\usepackage{booktabs}
\usepackage{threeparttable,multirow}
\usepackage{makecell}
\usepackage[ruled,linesnumbered]{algorithm2e}
\usepackage{microtype}

\DeclareMathOperator*{\argmin}{argmin}
\DeclareMathOperator*{\sign}{sign}

\DeclareMathOperator*{\F}{F}
\usepackage{schemata}
\newcommand\AB[2]{\schema{\schemabox{#1}}{\schemabox{#2}}}
\usepackage{enumitem}
\usepackage{pifont}
\newcommand{\cmark}{\text{\ding{51}}}
\newcommand{\xmark}{\text{\ding{55}}}

\usepackage{ragged2e}
\usepackage{times}  
\usepackage{helvet}  
\usepackage{courier}  
\usepackage{url}  
\usepackage{graphicx}  
\usepackage{chronology}
\newtheorem{definition}{Definition}

\newtheorem{theorem}{Theorem}

\newtheorem{proposition}{Proposition}

\newtheorem{assumption}{Assumption}

\usepackage{tikz}
\usetikzlibrary{calc,fit,arrows.meta,calc,positioning,decorations.pathreplacing}
\usepackage{xifthen}
\usetikzlibrary{arrows}

\tikzset{
	>=stealth', 
	grid/.style={step=1cm,gray!30,very thin},
	axis/.style={thick,<->},
	vect/.style={ultra thick,->},
	vnode/.style={midway,font=\scriptsize},
	proj/.style={dashed,color=gray!50,->},
	poly/.style={fill=blue!30,opacity=0.3}
}

\colorlet{A}{gray}
\colorlet{B}{lightgray}
\colorlet{C}{white}

\pgfdeclarelayer{background}
\pgfdeclarelayer{foreground}
\pgfsetlayers{background,main,foreground}

\tikzset{
        timeline/.style={arrows={}}%
        ,timeline style/.style={timeline/.append style={#1}}%
        ,year label/.style={font=\small\bfseries,below}
        ,year label style/.style={year label/.append style={#1}}%
        ,year tick/.style={tick size=0pt}%
        ,year tick style/.style={year tick/.append style={#1}}%
        ,minor tick/.style={tick size=0pt, very thin}%
        ,minor tick style/.style={minor tick/.append style={#1}}%
        ,period/.style={solid,line width=\timelinewidth,line cap=square}%
        ,periodbox/.style={font=\small\bfseries,text=black}
        ,eventline/.style={draw,red,thick,line cap=round,line join=round}%
        ,eventbox/.style={rectangle,rounded corners=3pt,inner sep=3pt,fill=red!25!white,text width=3cm,anchor=west,text=black,align=left,font=\small}%
        ,tick size/.code={\def\ticksize{#1}}%
        ,labeled years step/.code={\def\yearlabelstep{#1}}%
        ,minor tick step/.code={\def\minortickstep{#1}}%
        ,year tick step/.code={\def\yeartickstep{#1}}%
        ,enlarge timeline/.code={\def\enlarge{#1}}%
        ,eventboxa/.style={eventbox,text width=#1,draw=A,fill=none}%
        ,eventboxb/.style={eventbox,text width=#1,draw=A,fill=none}%
}

\newcommand*{\drawtimeline}[5][]{%
        \def\fromyear{#2}%
        \def\toyear{#3}%
        \def\timelinesize{#4}%
        \def\timelinewidth{#5}%
        \pgfmathsetmacro{\timelinesizept}{\timelinesize}%
        \pgfmathsetmacro{\timelinewidthpt}{\timelinewidth}%
        \pgfmathsetmacro{\timelineoffset}{\timelinewidth/2}
        \pgfmathsetmacro{\timelineoffsetpt}{\timelineoffset}
        \begin{scope}[x=1pt, y=1pt, 
                labeled years step=1,
                minor tick step=0.25,%
                enlarge timeline=0cm,%
                year tick step=1,#1]
                \pgfmathsetmacro{\enlargept}{\enlarge}
                \pgfmathsetmacro{\yearticksep}{\timelinesize/((\toyear-\fromyear)/\yeartickstep)}
                \pgfmathsetmacro{\minorticksep}{\timelinesize/((\toyear-\fromyear)/\minortickstep)}
                \pgfmathsetmacro{\minorticklast}{\minorticksep/\minortickstep}
                \foreach \y[remember=\y as \lasty (initially 0), count=\i from \fromyear] in {0,\yearticksep,...,\timelinesizept}{
                        \coordinate (Y-\i) at (\y,0);
                        \draw[year tick] (\y,-\ticksize/2) --  ++(0,\ticksize);
                        \ifnum\i=\toyear\breakforeach\else
                        \foreach \q[count=\j from 0] in {0,\minorticksep,...,\minorticklast}
                        {
                                \coordinate (Y-\i-\j) at (\q+\y,0);
                                \draw[minor tick] (\q+\y,-\ticksize/2) -- ++(0,\ticksize);
                        };\fi};%
                \pgfmathsetmacro{\nextyear}{int(\fromyear+\yearlabelstep)}
                \draw[timeline] (0,0) -- ++(-\enlargept,0) (0,0) -- ++(\timelinesizept,0) coordinate (end) -- ++(\enlargept,0);
        \end{scope}%

}

\newcommand{\period}[5]{\draw[period,#1] (Y-#2) -- (Y-#3) node[periodbox,#5,midway,text=white] {#4};}

\newcommand{\vevent}[9]{
        \pgfmathtruncatemacro{\syr}{#2}
        \pgfmathtruncatemacro{\smth}{#3-1}
        \pgfmathsetmacro{\dim}{#4/31}
        \ifthenelse{#3=12}{%
                \pgfmathtruncatemacro{\fyr}{#2+1}
                \pgfmathtruncatemacro{\fmth}{0}
        }{%
                \pgfmathtruncatemacro{\fyr}{#2}
                \pgfmathtruncatemacro{\fmth}{#3}
        }
        \draw[eventline,#1]($(Y-\syr-\smth)!\dim!(Y-\fyr-\fmth)$) -- ++(#5) -- ++(#6) node[#7] (#8) {#9};
}

\tikzset{
        block/.style 2 args = {
                draw=none, inner sep=0, outer sep=0,
                rounded corners=3pt,
                fit=(#1) (#2)}
}

\usetikzlibrary{calc,matrix,backgrounds,fit,positioning}
\tikzset{%
	nodeStyleGreen/.style={
		draw=green!30!black,
		fill=green!50!lime!30,
		align=left,
		very thick,
		rounded corners
	},
	nodeStyleRed/.style={
		draw=red,
		fill=red!50!lime!30,
		align=left,
		very thick,
		rounded corners
	},
	nodeStyleBlue/.style={
		draw=blue,
		fill=blue!50!lime!30,
		align=left,
		very thick,
		rounded corners
	},
	nodeStyleYellow/.style={
	draw=yellow!80!black,
	fill=yellow!50!lime!30,
	align=left,
	very thick,
	rounded corners
	},
	lineStyleRed/.style={
		color=red,opacity=0.75,line width=2pt,
	},
	lineStyleGreen/.style={
		color=green!40!black,opacity=0.75,line width=2pt,
	},
	lineStyleBlue/.style={
		color=blue,opacity=0.75,line width=2pt,
	},
	lineStyleYellow/.style={
	color=yellow!80!black,opacity=0.75,line width=2pt,
},
}

\usetikzlibrary{arrows,chains,matrix,positioning,scopes}
\makeatletter
\tikzset{join/.code=\tikzset{after node path={%
			\ifx\tikzchainprevious\pgfutil@empty\else(\tikzchainprevious)%
			edge[every join]#1(\tikzchaincurrent)\fi}}}
\makeatother
\tikzset{>=stealth',every on chain/.append style={join},
	every join/.style={->}}
\tikzstyle{labeled}=[execute at begin node=$\scriptstyle,
execute at end node=$]

\ifCLASSOPTIONcompsoc
  \usepackage[nocompress]{cite}
\else
  \usepackage{cite}
\fi
\ifCLASSINFOpdf
\else
\fi

\definecolor{red}{rgb}{0.8,0,0}
\definecolor{blue}{rgb}{0,0,0.8}
\definecolor{green}{rgb}{0,0.4,0}

\begin{document}

%
\title{Random Features for Kernel Approximation: A Survey on Algorithms, Theory, and Beyond}

\author{Fanghui Liu, Xiaolin Huang, Yudong Chen, Johan A.K. Suykens
\thanks{
F. Liu and J.A.K. Suykens are with the Department of Electrical Engineering
(ESAT-STADIUS), KU Leuven, B-3001 Leuven, Belgium (email: \{fanghui.liu;johan.suykens\}@esat.kuleuven.be).
 }
\thanks{
	X. Huang is with Institute of Image Processing and Pattern Recognition, and also with Institute of Medical Robotics, Shanghai Jiao Tong University, Shanghai 200240, P.R. China (e-mail: xiaolinhuang@sjtu.edu.cn).
}
\thanks{
	Y. Chen is with School of Operations Research and Information Engineering, Cornell University, Ithaca, NY 14850 USA (e-mail: yudong.chen@cornell.edu).
}
}


\IEEEtitleabstractindextext{%
\begin{abstract}
	\justifying  
The class of random features is one of the most popular techniques to speed up kernel methods in large-scale problems. Related works have been recognized by the NeurIPS Test-of-Time award in 2017 and the ICML Best Paper Finalist in 2019. The body of work on random features has grown rapidly, and hence it is desirable to have a comprehensive overview on this topic explaining the connections among various algorithms and theoretical results. In this survey, we systematically review the work on random features from the past ten years. First, the motivations, characteristics and contributions of representative random features based algorithms are summarized according to their sampling schemes, learning procedures, variance reduction properties and how they exploit training data. Second, we review theoretical results that center around the following key question: how many random features are needed to ensure a high approximation quality or no loss in the empirical/expected risks of the learned estimator. Third, we provide a comprehensive evaluation of popular random features based algorithms on several large-scale benchmark datasets and discuss their  approximation quality and prediction performance for classification. Last, we discuss the relationship between random features and modern over-parameterized deep neural networks (DNNs), including the use of high dimensional random features in the analysis of DNNs as well as the gaps between current theoretical and empirical results. This survey may serve as a gentle introduction to this topic, and as a users' guide for practitioners interested in applying the representative algorithms and understanding theoretical results under various technical assumptions. We hope that this survey will facilitate discussion on the open problems in this topic, and more importantly, shed light on future research directions.
\justifying  
\end{abstract}

}

\maketitle

\IEEEdisplaynontitleabstractindextext

\IEEEpeerreviewmaketitle

\ifCLASSOPTIONcompsoc
\IEEEraisesectionheading{\section{Introduction}\label{sec:introduction}}
\else
\section{Introduction}
\fi
\IEEEPARstart{K}{ernel} methods \cite{Sch2003Learning,suykens2002least,kafai2018croification} are one of the most powerful techniques for nonlinear statistical learning problems with a wide range of successful applications.
Let $\bm x, \bm x' \in \mathcal{X} \subseteq \mathbb{R}^d$ be two samples and $\phi: \mathcal{X} \rightarrow \mathcal{H}$ be a nonlinear feature map transforming each element in $\mathcal{X}$ into a reproducing kernel Hilbert space (RKHS) $\mathcal{H}$, in which the inner product between $\phi(\bm x)$ and $\phi(\bm x')$ endowed by $\mathcal{H}$ can be computed using a kernel function $k(\cdot, \cdot): \mathbb{R}^d \times \mathbb{R}^d \rightarrow \mathbb{R}$ as
$\langle \phi(\bm x), \phi(\bm x') \rangle_{\mathcal{H}} = k(\bm x, \bm x')$.
In practice, the kernel function $k$ is directly given to obtain the inner product $\langle \phi(\bm x), \phi(\bm x') \rangle_{\mathcal{H}}$ instead of finding the explicit expression of $\phi$, which is known as the \emph{kernel trick}. 
Benefiting from this scheme, kernel methods are effective for learning nonlinear structures but often suffer from scalability issues in large-scale problems due to high space and time complexities.
For instance, given $n$ samples in the original $d$-dimensional space $\mathcal{X}$, kernel ridge regression (KRR) requires $\mathcal{O}(n^3)$ training time and $\mathcal{O}(n^2)$ space to store the kernel matrix, which is often computationally infeasible when $ n $ is large.

To overcome the poor scalability of kernel methods, kernel approximation is an effective technique by constructing an explicit mapping $\Psi: \mathbb{R}^d \rightarrow \mathbb{R}^s$ such that $k(\bm x, \bm y) \approx \Psi(\bm x)^{\!\top} \Psi(\bm y)$.
By doing so, an efficient linear model can be well learned in the transformed space with $\mathcal{O}(ns^2)$ time and $\mathcal{O}(ns)$ memory while retaining the expressive power of nonlinear methods.
A series of kernel approximation algorithms have been developed in the past years, including divide-and-conquer approaches \cite{hsieh2014divide,zhang2013divide,liu2020learning}, greedy basis selection techniques \cite{smola2000sparse} and Nystr\"{o}m methods \cite{williams2001using}. These methods provide a data dependent vector representation of the kernel.
Random Fourier features (RFF) \cite{rahimi2007random}, on the other hand, is a typical data-independent technique to approximate the kernel function using an explicit feature mapping.
This survey focuses on RFF and its variants for kernel approximation.
RFF applies in particular to shift-invariant (also called ``stationary'') kernels that satisfy $k(\bm x, \bm x') = k(\bm x - \bm x')$.
By virtue of the correspondence between a shift-invariant kernel and its Fourier spectral density, the kernel can be approximated by $k(\bm x, \bm x') \approx \langle \varphi(\bm x), \varphi(\bm x') \rangle$, where the explicit mapping $\varphi: \mathbb{R}^d \rightarrow \mathbb{R}^s$ is obtained by sampling from a distribution defined by the inverse Fourier transform of $k$.
To scale kernel methods in the large sample case (e.g., $n \gg d$), the number of random features $s$ is often taken to be larger than the original sample dimension $d$ but much smaller than the sample size $n$ to achieve computational efficiency in practice.\footnote{
	Random features model can be regarded as an over-parameterized model allowing for $s \gg n$, refer to Section~\ref{sec:DNNs} for details.}
Accordingly, the random features model is a powerful tool for scaling up traditional kernel methods \cite{lopez2014randomized,sun2018but}, neural tangent kernel \cite{jacot2018neural,arora2019exact,zandieh2021scaling}, graph neural networks~\cite{du2019graph,zambon2020graph}, and attention in Transformers \cite{choromanski2021rethinking,peng2021random}.
Interestingly, the random features model can be viewed as a class of two-layer neural networks with fixed weights in the first layer. This connection has important theoretical implications.
It has been observed that deep neural networks (DNNs) exhibit certain intriguing phenomena such as the ability to fit random labels \cite{zhang2016understanding} and double descent \cite{belkin2019reconciling} in the \emph{over-parameterized} regime. Theoretical results \cite{arora2019exact,cao2019generalization,arora2019fine,ji2020polylogarithmic} for random features can be leveraged to explain these phenomena and provide an analysis of two-layer \emph{over-parameterized} neural networks.
Partly due to its far-reaching repercussions, the seminal work by Rahimi and Recht on RFF~\cite{rahimi2007random} won the Test-of-Time Award in the \emph{Thirty-first Advances in Neural Information Processing Systems} (NeurIPS 2017).

RFF spawns a new direction for kernel approximation, and the past ten years has witnessed a flurry of research papers devoted to this topic.
On the algorithmic side, subsequent work has focused on improving the kernel approximation quality~\cite{Yu2016Orthogonal,Avron2016Quasi} and decreasing the time and space complexities \cite{dao2017gaussian,munkhoeva2018quadrature}.
Implementation of RFF has in fact been taken to the hardware level \cite{saade2016random,ohana2019kernel}.
On the theoretical side, a series of works aim to address the following two key questions:
\begin{enumerate}
        \item \textbf{Approximation:} how many random features are needed to ensure high quality of kernel approximation?
        \item \textbf{Generalization:} how many random features are needed to incur no loss in the expected risk of a learned estimator?
\end{enumerate}
Here ``no loss" means how large $s$ should be for the (approximated) kernel estimator with $s$ random features to be almost as good as the exact one.
Much research effort has been devoted to this direction, including analyzing the kernel approximation error (the first question above) \cite{rahimi2007random,sutherland2015error}, and studying the risk and generalization properties (the second question above) \cite{sun2018but,li2019towards}. Increasingly refined and general results have been obtained over the years.
In the \emph{Thirty-sixth International Conference on Machine Learning} (ICML 2019), Li et al.\cite{li2019towards} were recognized by the Honorable Mentions (best paper finalist) for their unified theoretical analysis of RFF.

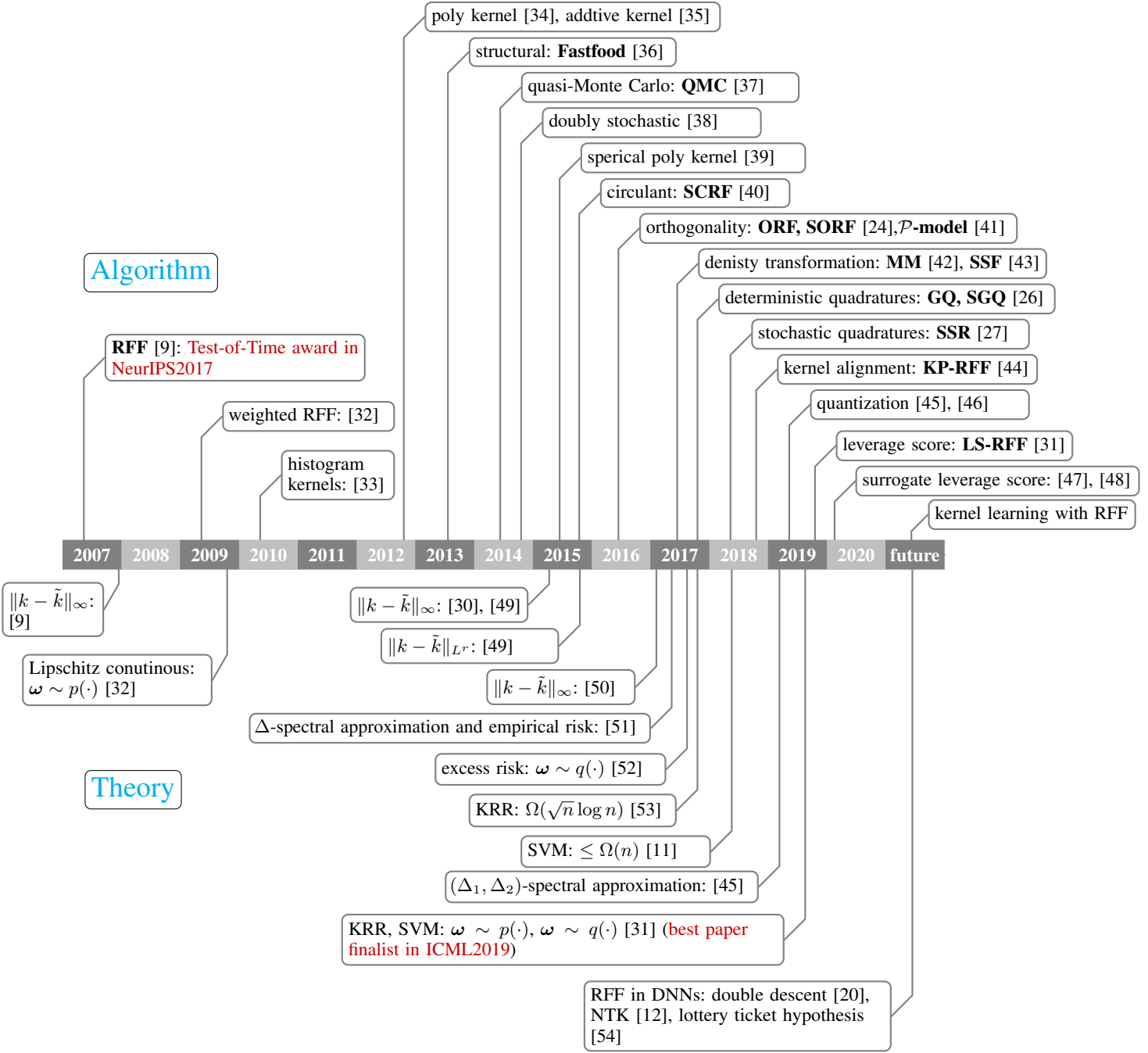
\begin{figure*}
	\centering
	\begin{tikzpicture}{scale=0.5}
		\drawtimeline[
		labeled years step=1,
		minor tick step=0.083333,
		timeline style={draw=gray,line width=\timelinewidthpt},
		minor tick style={-,lightgray,tick size=0pt,line width=0pt,yshift=-\timelineoffsetpt},
		]%
		{2007}{2022}{15cm}{0.5cm};
		\period{A}{2007}{2008}{2007}{}
		\period{B}{2008}{2009}{2008}{}
		\period{A}{2009}{2010}{2009}{}
		\period{B}{2010}{2011}{2010}{}
		\period{A}{2011}{2012}{2011}{}
		\period{B}{2012}{2013}{2012}{}
		\period{A}{2013}{2014}{2013}{}
		\period{B}{2014}{2015}{2014}{}
		\period{A}{2015}{2016}{2015}{}
		\period{B}{2016}{2017}{2016}{}
		\period{A}{2017}{2018}{2017}{}
		\period{B}{2018}{2019}{2018}{}
		\period{A}{2019}{2020}{2019}{}
		\period{B}{2020}{2021}{2020}{}
		\period{A}{2021}{2022}{future}{}
		\begin{pgfonlayer}{background}
			\vevent{A}{2007}{5}{10}{90:3cm}{45:0.5cm}{eventboxa=4.2cm,anchor=west}{H}{{\bf RFF}\cite{rahimi2007random}: {\color{red}Test-of-Time award in NeurIPS2017}}
			
			\vevent{A}{2009}{5}{10}{90:2cm}{45:0.5cm}{eventboxa=2.7cm,anchor=west}{H}{weighted RFF: \cite{rahimi2009weighted}}
			
			\vevent{A}{2010}{5}{10}{90:1cm}{45:0.5cm}{eventboxa=1.7cm,anchor=west}{H}{histogram kernels: \cite{li2010random}}
			
			\vevent{A}{2012}{10}{20}{90:8.8cm}{45:0.5cm}{eventboxa=4.8cm,anchor=west}{H}{poly kernel \cite{kar2012random}, addtive kernel \cite{Vedaldi2012Efficient}}
			\vevent{A}{2013}{7}{20}{90:8.2cm}{45:0.5cm}{eventboxa=3.3cm,anchor=west}{H}{structural: {\bf Fastfood}\cite{le2013fastfood}}
			\vevent{A}{2014}{6}{10}{90:7.6cm}{45:0.5cm}{eventboxa=4.5cm,anchor=west}{H}{quasi-Monte Carlo: {\bf QMC}\cite{yang2014quasi}}
			\vevent{A}{2014}{10}{18}{90:7cm}{45:0.5cm}{eventboxa=3.5cm,anchor=west}{H}{doubly stochastic\cite{dai2014scalable}}
			
			\vevent{A}{2015}{6}{14}{90:6.4cm}{45:0.5cm}{eventboxa=3.6cm,anchor=west}{H}{sperical poly kernel \cite{pennington2015spherical}}
			\vevent{A}{2015}{10}{14}{90:5.8cm}{45:0.5cm}{eventboxa=3cm,anchor=west}{H}{circulant: {\bf SCRF}\cite{feng2015random}}
			
			\vevent{A}{2016}{6}{14}{90:5.2cm}{45:0.5cm}{eventboxa=6.2cm,anchor=west}{H}{orthogonality: {\bf ORF, SORF}\cite{Yu2016Orthogonal},{\bf $\mathcal{P}$-model}\cite{choromanski2016recycling}}
			
			\vevent{A}{2017}{6}{14}{90:4.6cm}{45:0.5cm}{eventboxa=5.7cm,anchor=west}{H}{denisty transformation: {\bf MM}\cite{shen2017random}, {\bf SSF}\cite{lyu2017spherical}}
			\vevent{A}{2017}{10}{18}{90:4cm}{45:0.5cm}{eventboxa=5.5cm,anchor=west}{H}{deterministic quadratures: {\bf GQ, SGQ}\cite{dao2017gaussian}}
			
			\vevent{A}{2018}{5}{10}{90:3.4cm}{45:0.5cm}{eventboxa=4.5cm,anchor=west}{H}{stochastic quadratures: {\bf SSR}\cite{munkhoeva2018quadrature}}
			\vevent{A}{2018}{10}{18}{90:2.8cm}{45:0.5cm}{eventboxa=4.2cm,anchor=west}{H}{kernel alignment: {\bf KP-RFF} \cite{shahrampour2018data}}
			\vevent{A}{2019}{5}{10}{90:2.2cm}{45:0.5cm}{eventboxa=3.5cm,anchor=west}{H}{quantization \cite{zhang2019f}, \cite{agrawal2019data}}
			\vevent{A}{2019}{10}{18}{90:1.5cm}{45:0.5cm}{eventboxa=3.8cm,anchor=west}{H}{leverage score: {\bf LS-RFF}\cite{li2019towards}}
			\vevent{A}{2020}{5}{-75}{90:0.9cm}{45:0.5cm}{eventboxa=4.6cm,anchor=west}{H}{surrogate leverage score: \cite{liu2020random,erdelyi2020fourier}}
			\vevent{A}{2021}{6}{8}{90:0.4cm}{45:0.4cm}{eventboxa=3.3cm,anchor=west}{H}{kernel learning with RFF}
			
			\vevent{A}{2007}{12}{14}{-90:0.5cm}{60:-0.5cm}{eventboxa=1.5cm,anchor=east}{H}{$\| k - \tilde{k} \|_{\infty}$: \cite{rahimi2007random}}
			\vevent{A}{2009}{10}{18}{-90:1.7cm}{60:-0.5cm}{eventboxa=3.0cm,anchor=east}{H}{Lipschitz conutinous: $\bm \omega \sim p(\cdot)$ \cite{rahimi2009weighted}}
			
			\vevent{A}{2015}{4}{10}{-90:0.5cm}{45:-0.5cm}{eventboxa=2.8cm,anchor=east}{H}{$\| k - \tilde{k} \|_{\infty}$: \cite{sutherland2015error,sriperumbudur2015optimal}}
			\vevent{A}{2015}{10}{18}{-90:1.2cm}{45:-0.5cm}{eventboxa=2.8cm,anchor=east}{H}{$\| k - \tilde{k} \|_{L^r}$: \cite{sriperumbudur2015optimal}}
			\vevent{A}{2017}{2}{6}{-90:1.9cm}{45:-0.5cm}{eventboxa=2.3cm,anchor=east}{H}{$\| k - \tilde{k} \|_{\infty}$: \cite{honorio2017error}}
			\vevent{A}{2017}{5}{10}{-90:2.6cm}{45:-0.5cm}{eventboxa=6.6cm,anchor=east}{H}{$\Delta$-spectral approximation and empirical risk: \cite{avron2017random}}
			\vevent{A}{2017}{8}{14}{-90:3.3cm}{45:-0.5cm}{eventboxa=3.7cm,anchor=east}{H}{excess risk:  $\bm \omega \sim q(\cdot)$ \cite{bach2017equivalence} }
			\vevent{A}{2017}{10}{18}{-90:4cm}{45:-0.5cm}{eventboxa=3.3cm,anchor=east}{H}{KRR: $\Omega(\sqrt{n} \log n)$ \cite{Rudi2017Generalization}}
			\vevent{A}{2018}{5}{18}{-90:4.7cm}{45:-0.5cm}{eventboxa=3.0cm,anchor=east}{H}{SVM: $\leq \Omega(n)$ \cite{sun2018but}}
			\vevent{A}{2019}{3}{10}{-90:5.3cm}{45:-0.5cm}{eventboxa=5.1cm,anchor=east}{H}{$(\Delta_1, \Delta_2)$-spectral approximation: \cite{zhang2019f}}
			\vevent{A}{2019}{8}{18}{-90:6.2cm}{45:-0.5cm}{eventboxa=7.3cm,anchor=east}{H}{KRR, SVM: $\bm \omega \sim p(\cdot)$, $\bm \omega \sim q(\cdot)$ \cite{li2019towards} ({\color{red}best paper finalist in ICML2019})}
			\vevent{A}{2021}{6}{14}{-90:7.5cm}{45:-0.5cm}{eventboxa=5cm,anchor=east}{H}{RFF in DNNs: double descent \cite{belkin2019reconciling}, NTK \cite{jacot2018neural}, lottery ticket hypothesis \cite{malach2020proving}}
		\end{pgfonlayer}
		%
		
		\node [rectangle,rounded corners=1mm, draw, inner sep=3pt] (O) at
		(1.5,4.8) {\color{cyan}{\Large Algorithm}};
		\node [rectangle,rounded corners=1mm,draw, inner sep=3pt] (O) at
		(1.2,-4) {\color{cyan}{\Large Theory}};
	\end{tikzpicture}
	\caption{Timeline of representative work on the algorithms and theory of random features.}
	\label{fig:timeline}
	\vspace{-0.1cm}
\end{figure*}

RFF has proved effective in a broad range of machine learning tasks. Given its remarkable empirical success and the rapid growth of the related literature, we believe it is desirable to have a comprehensive overview on this topic summarizing the progress in algorithm design and applications, and elucidating existing theoretical results and their underlying  assumptions.
With this goal in mind, in this survey we systematically review the work from the past ten years on the algorithms, theory and applications of random features methods.
Figure~\ref{fig:timeline} shows a schematic overview of the history of the work on random features in recent years. 
The main contributions of this survey include:
\begin{enumerate}
        \item We provide an overview of a wide range of random features based algorithms, re-organize the formulation of representative approaches under a unifying framework for a direct understanding and comparison.
        \item We summarize existing theoretical results on the kernel approximation error measured in various metrics, as well as results on generalization risk of kernel estimators. The underlying assumptions in these results are discussed in detail. In particular, we (partly) answer an open question in this topic: why good kernel approximation performance cannot lead to good generalization performance?
        \item We systematically evaluate and compare the empirical performance of representative random features based algorithms under different experimental settings.
        \item We discuss recent research trends on (high dimensional) random features in over-parameterized settings for understanding generalization properties of over-parameterized neural networks as well as the gaps in existing theoretical analysis. We view this topic as a promising research direction.
\end{enumerate}
 
The remainder of this paper is organized as follows. Section~\ref{sec:overviewRFF} presents the preliminaries and a taxonomy of random features based algorithms. We review \emph{data-independent} algorithms in Section~\ref{sec:data-independent} and \emph{data-dependent} approaches in Section~\ref{sec:data-dependent}.
In Section~\ref{sec:theoretical}, we survey existing theoretical results on kernel approximation and generalization performance.
Experimental comparisons of representative random features based methods are given
in Section~\ref{sec:experiments}.
In Section~\ref{sec:DNNs}, we discuss recent results on random features in over-parameterized regimes.
The paper is concluded in  Section~\ref{sec:conclusion} with a discussion on future directions.

\begin{table}[t]
	\centering
	\caption{\footnotesize Commonly used parameters and symbols.}
	\label{tabparam}
	\begin{threeparttable}
		\begin{tabular}{p{0.4cm}c|p{0.4cm}cccccccccccc}
			\toprule
			\centering{Notation} & Definition &  \centering{Notation} & Definition  \\
			\midrule
			\centering{$n$} & number of samples & \centering{$d$} & feature dimension \\
			\hline 
			\centering{$s$} & number of random features & \centering{$\lambda$} & regularization parameter
			\\
			\hline
			\centering{$k$} & (original) kernel function & \centering{$\tilde{k}$} & (approximated) kernel function
			\\
			\hline
			\centering{$\bm \omega_i$} & random feature & \centering{$\bm \beta_{\lambda}$} & optimization variable
			\\
			\hline
			\centering{$\bm x$} & data point & \centering{${\bm y}$} & label vector \\
			\hline
			\centering{$\varsigma$} & Gaussian kernel width & \centering{${\sigma}$} & activation function \\
				\hline
			\centering{$\bm e_i$} & standard basis vector & \centering{$u$} & $u:= \langle \bm x, \bm x^{\prime}\rangle/ ( \|\bm x\|\|\bm x^{\prime}\|)$ \\
			\hline
			\centering{$\bm K$} & (original) kernel matrix & \centering{$\widetilde{\bm K}$} & (approximated) kernel matrix \\
			\hline
			\centering{$\bm \tau$} & $\bm \tau := \bm x - \bm x'$ & \centering{$\tau$} & $\tau := \| \bm \tau \|_2$ \\
			\hline
			\centering{$\bm Z$} & random feature matrix & \centering{${\bm W}$} & transformation matrix \\
			\hline
			\centering{$f_{\rho}$} & target function & \centering{$\ell$} & loss function \\
			\hline
			\centering{$f_{\bm z,\lambda}$} & (original) empirical functional & \centering{$\tilde{f}_{\bm z,\lambda}$} & (approximated) functional \\
			\hline
			\centering{$\mathcal{E}_{\bm z}$} & empirical risk & \centering{$\mathcal{E}$} & expected risk \\
			\hline
			\centering{$l_{\lambda}(\bm \omega)$} & ridge leverage function & \centering{$d_{\bm K}^{\lambda}$} & effective dimension (matrix) \\
			\hline
			\centering{$\Sigma$} & integral operator & \centering{$\mathcal{N}(\lambda)$} & effective dimension (operator) \\
			\hline
			\centering{$ \otimes $} & tensor product & \centering{$\lesssim $} & $\leq$ with a constant $C$ times \\
			\hline \hline
			\centering{$\alpha$} & convergence rate for $\lambda$ & \centering{$\gamma$} & rate for effective dimension \\
			\bottomrule
		\end{tabular}
	\end{threeparttable}
	\vspace{-0.1cm}
\end{table}

\vspace{-0.3cm}
\section{Preliminaries and Taxonomies}
\label{sec:overviewRFF}

In this section, we introduce preliminaries on the problem setting and theoretical foundation of random features.
We then present a taxonomy of existing random features based algorithms, which sets the stage for the subsequent discussion. A set of commonly used parameters is summarized in Table~\ref{tabparam}.

\vspace{-0.3cm}
\subsection{Problem Settings}
\label{sec:setting}


Consider the following standard supervised learning setup. Let $\mathcal{X} \subset \mathbb{R}^d$ be a compact metric space of samples, and $\mathcal{Y}=\{ -1, 1\}$ (in classification) or $\mathcal{Y} \subseteq \mathbb{R}$ (in regression) be the label space. We assume that a sample set $ \{  \bm z_i = (\bm x_i, y_i) \}_{i=1}^n $ is drawn from a non-degenerate unknown Borel probability measure $\rho$ on $\mathcal{X} \times \mathcal{Y}$.
Let $\mathcal{H}$ be a RKHS endowed with a positive definite kernel function $k(\cdot,\cdot)$, and $\bm K = [k(\bm x_i, \bm x_j)]_{i,j=1}^{n}$ be the kernel matrix associated with the samples.
The \emph{target function} of $\rho$ is defined as
$f_{\rho}(\bm x) = \int_\mathcal{Y} y \mathrm{d} \rho(y|\bm x)$ for $ \bm x \in \mathcal{X}$, where $\rho(\cdot|\bm x)$ is the conditional distribution of $y$ given $\bm x $.
The typical empirical risk minimization problem is considered as
\begin{equation}\label{ermp}
f_{\bm{z},\lambda} := \argmin_{f \in \mathcal{H}  } \left\{\! \frac{1}{n} \sum_{i=1}^{n} \ell \big(y_i,f(\bm x_i) \big) + \lambda \| f \|^2_{\mathcal{H}} \! \right\}\,,
\end{equation}
where $\ell: \mathcal{Y} \times \mathcal{Y} \rightarrow \mathbb{R}$ is a loss function and $\lambda \equiv \lambda(n) > 0$ is a regularization parameter. In learning theory, one typically assumes that $\lim_{n \rightarrow \infty} \lambda(n) = 0$ and adopts $\lambda := n^{-\alpha}$ with $\alpha \in (0,1]$.

The loss function $\ell(y, f(\bm x))$ in Eq.~\eqref{ermp} measures the quality of the prediction $f(\bm x)$ at $\bm x \in \mathcal{X}$ with respect to the observed response $y$.
Popular choices of $\ell$ include the squared loss $\ell(y, f(\bm x)) = (y - f(\bm x))^2$ in kernel ridge regression (KRR) and the hinge loss  $\ell(y, f(\bm x)) = \max(0,1-yf(\bm x))$ in support vector machines (SVM), etc.
For a given $\ell$, the empirical risk functional on the sample set is defined as $\mathcal{E}_{\bm z}(f) = \frac{1}{n} \sum_{i=1}^{n} \ell(y_i, f(\bm x_i))$, and the corresponding expected risk is defined as $\mathcal{E}(f) = \int_{\mathcal{X} \times \mathcal{Y}} \ell(y, f(\bm x)) \mathrm{d} \rho$.
The statistical theory of supervised learning in an approximation theory view aims to understand the generalization property of $f_{\bm{z},\lambda}$ as an approximation of the true target function $f_{\rho}$,  which can be quantified by the excess risk $\mathcal{E}(f_{\bm z, \lambda}) - \mathcal{E}(f_{\rho})$, or the estimation error $\| f_{\bm z, \lambda} - f_{\rho}\|^2$ in an appropriate norm $\| \cdot \|$.

Using an explicit randomized feature mapping $\varphi: \mathbb{R}^d \rightarrow \mathbb{R}^s$, one may approximate the kernel function $k(\bm x, \bm x')$ by $ \tilde{k}(\bm x, \bm x') = \langle \varphi(\bm x), \varphi(\bm x') \rangle$.
In this case, the approximate kernel $\tilde{k}(\cdot , \cdot)$ defines an RKHS $\widetilde{\mathcal{H}}$ (not necessarily contained in the RKHS $\mathcal{H}$ associated with the original kernel function $k$).
With the above approximation, one solves the following approximate version of problem~\eqref{ermp}:
\begin{equation}\label{ermpa}
        \tilde{f}_{\bm{z},\lambda} := \argmin_{f \in \widetilde{\mathcal{H}}  } \left\{\! \frac{1}{n} \sum_{i=1}^{n} \ell \big(y_i,f(\bm x_i) \big) + \lambda \| f \|^2_{\widetilde{\mathcal{H}}} \! \right\}\,.
\end{equation}
By the representer theorem \cite{Sch2003Learning}, the above problem can be rewritten as a finite-dimensional empirical risk minimization problem\vspace{-0.1cm}
\begin{equation}\label{ermrff}
\bm \beta_{\lambda} :=\underset{\bm \beta \in \mathbb{R}^{s}}{\argmin } ~\frac{1}{n} \sum_{i=1}^n \ell \left(y_i, \bm \beta^{\!\top}\varphi(\bm x_i)\right) +\lambda \|\bm \beta\|_{2}^{2}\,.
\end{equation}
For example, in least squares regression where $\ell$ is the squared loss, the first term in problem~\eqref{ermrff} is equivalent to $\left\|\bm y- \bm {Z} \bm \beta\right\|_{2}^{2}$, where  $\bm y = [y_1, y_2, \cdots, y_n]^{\!\top}$ is the label vector and  $\bm Z =[\varphi(\bm x_1), \cdots, \varphi(\bm x_n)]^{\!\top} \in \mathbb{R}^{n \times s}$ is the random feature matrix.
This is a linear ridge regression problem in the space spanned by the random features, with the optimal prediction given by $\tilde{f}_{\bm{z},\lambda}(\bm x') = \bm \beta_{\lambda}^{\!\top}\varphi(\bm x')$ for a new data point $\bm x'$, where $\bm \beta_{\lambda}$ has the explicit expression  
$ \bm \beta_{\lambda} = (\bm Z^{\!\top} \bm Z + n \lambda \bm I)^{-1} \bm Z^{\!\top} \bm y$.
For classification, one may take the sign to output the binary classification labels.
Note that problem~\eqref{ermrff} also corresponds to fixed-size kernel methods with feature map approximation (related to Nystr\"{o}m approximation) and estimation in the primal \cite{suykens2002least}.

\vspace{-0.1cm}
\subsection{Theoretical Foundation of Random Features}
\label{sec:theorf}
The theoretical foundation of RFF builds on Bochner's celebrated characterization of positive definite functions.

\begin{theorem}[Bochner's Theorem \cite{bochner2005harmonic}]
	\label{bochner}
	A continuous and shift-invariant function $k: \mathbb{R}^d \times \mathbb{R}^d \rightarrow \mathbb{R}$ is positive definite if and only if it can be represented as
	\begin{equation*}
		k(\bm x- \bm x') = \int_{\mathbb{R}^d}  \exp\left(\mathrm{i}{\bm \omega}^{\!\top}(\bm x- \bm x')\right) \mu_k(\mathrm{d} {\bm \omega})\,,
	\end{equation*}
	where $\mu_k $ is a positive finite measure on the frequencies $\bm \omega$.
\end{theorem}
According to Bochner's theorem, the spectral distribution $\mu_k$ of a stationary kernel $k$ is the finite measure induced by a Fourier transform. 
By setting $k(0)=1$, we may normalize $\mu_k$ to a probability density $p$ (the Fourier transform associated with $k$), hence
\begin{equation}\label{rffdef}
\begin{split}
k(\bm x- \bm x') & = \int_{\mathbb{R}^d} \exp\big(\mbox{i}{\bm \omega}^{\!\top}(\bm x- \bm x')\big) \mu (\mbox{d} {\bm \omega}) \\
& = \mathbb{E}_{{\bm \omega} \sim p(\cdot) } \big[\exp(\mbox{i}{\bm \omega}^{\!\top}\bm x) \exp(\mbox{i}{\bm \omega}^{\!\top}\bm x')^*\big]\,,
\end{split}
\end{equation}
where the symbol $\bm z^*$ denotes the complex conjugate of $\bm z$. 
The kernels used in practice are typically real-valued and thus the imaginary part in Eq.~\eqref{rffdef} can be discarded.
According to Eq.~\eqref{rffdef}, RFF makes use of the standard Monte Carlo sampling scheme to approximate $k(\bm x , \bm x')$. In particular, one uses the approximation $$k(\bm x, \bm x') \!=\! \mathbb{E}_{{\bm \omega} \sim p} [\varphi_p( \bm x)^{\!\top} \varphi_p(\bm x')] \approx \tilde{k}_p(\bm x , \bm x') \!:=\! \varphi_p(\bm x)^{\!\top} \varphi_p(\bm x') $$ with the explicit feature mapping\footnote{The subscript in $\varphi_p$, $\bm Z_p$, $k_p$ (and other symbols) emphasizes the dependence on the distribution $p(\cdot)$ but can be omitted for notational simplicity.} 
\begin{equation}\label{mapping}
\varphi_p(\bm x) \! := \! \frac{1}{\sqrt{s}}
\big[\exp(-\mbox{i}{\bm \omega}^{\!\top}_1 \bm x), \cdots,\exp(-\mbox{i}{\bm \omega}^{\!\top}_s \bm x)]^{\!\top}\,,
\end{equation}
where $\{ {\bm \omega}_i\}_{i=1}^s$ are sampled from $p(\cdot)$ \emph{independently} of the training set.
Consequently, the original kernel matrix $\bm K = [k(\bm x_i, \bm x_j)]_{n \times n} $ can be approximated by $\bm K \approx \widetilde{\bm K}_p =\bm Z_p \bm Z_p^{\! \top} $ with $\bm Z_p =[\varphi_p(\bm x_1), \cdots, \varphi_p(\bm x_n)]^{\!\top} \in \mathbb{R}^{n \times s}$.
It is convenient to introduce the shorthand  $z_p({\bm \omega}_i, \bm x_j) := \exp(-\mbox{i}{\bm \omega}^{\!\top}_i \bm x_j)$ such that $\varphi_p(\bm x)=1/\sqrt{s}[z_p({\bm \omega}_1, \bm x), \cdots, z_p({\bm \omega}_s, \bm x)]^{\!\top}$. With this notation, the approximate kernel $ \tilde{k}_p(\bm x , \bm x')$ can be rewritten as
$\tilde{k}_p(\bm x , \bm x') = \frac{1}{s} \sum_{i=1}^s z_p({\bm \omega}_i, \bm x) z_p({\bm \omega}_i, \bm x')
$.

A similar characterization in Eq.~\eqref{rffdef} is available for rotation-invariant kernels, where the Fourier basis functions are \emph{spherical harmonics} \cite{schoenberg1942positive,smola2001regularization}.
Here rotation-invariant kernels are dot-product kernels defined on the unit sphere $\mathcal{X}=\mathbb{S}^{d-1} := \{ \bm x \in \mathbb{R}^{d} : \| \bm x \|_2 = 1 \}$, and can be represented as a non-negative expansion with spherical harmonics, refer to the book \cite{muller2006spherical} for details.
\begin{theorem}[\hspace{1sp}\cite{schoenberg1942positive}]\label{harmonic}
	A rotation-invariant continuous function $k: \mathbb{S}^{d-1} \times \mathbb{S}^{d-1} \rightarrow \mathbb{R}$ is positive definite if and only if it has a symmetric non-negative expansion into spherical harmonics $Y_{\ell,m}^d$, that is
	\begin{equation*}
		k(\bm x, \bm x') \equiv k(\langle \bm x, \bm x' \rangle)=\sum_{i=0}^{\infty} \Lambda_{i} \sum_{j=1}^{N(d, i)} Y_{i, j}(\bm x) Y_{i, j}(\bm x')\,,
	\end{equation*}
	where $\Lambda_i \geq 0$ are the Fourier coefficients, $Y_{i, j}$ is the spherical harmonics, and $N(d, i)=\frac{2 i+d-2}{i}\left(\begin{array}{c}i+d-3 \\ d-2\end{array}\right)$.
\end{theorem}

Note that, dot product kernels defined in $\mathbb{R}^{d}$ do not belong to the \emph{rotation-invariant} class. Nevertheless, by virtue of the neural network structure under Gaussian initialization, some dot product kernels defined on $\mathbb{R}^{d}$ are able to benefit from the sampling framework behind RFF.
Given a two-layer network of the form $f(\bm x; \bm \theta) = \sqrt{\frac{2}{s}} \sum_{j=1}^s a_j \sigma(\bm \omega_j^{\!\top} \bm x)$ with $s$ neurons (notation chosen to be consistent with the number of random features), for some activation function $\sigma$ and $\bm x \in \mathbb{R}^d$, when $\bm \omega \sim \mathcal{N}(\bm 0, \bm I_d)$ are fixed and only the second layer (parameters $\bm a$) are optimized\footnote{Extreme learning machine \cite{huang2006extreme} is another structure in a two-layer feedforward neural network by randomly hidden nodes.}, this actually corresponds to random features approximation\vspace{-0.05cm}
\begin{equation}\label{kernelfor}
k\left(\bm x, \bm x' \right)=\mathbb{E}_{\bm \omega \sim \mathcal{N}(\bm 0, \bm I_d)}[\sigma(\bm \omega^{\top} \bm x) \sigma(\bm \omega^{\!\top} \bm x')]\,,
\end{equation}
where the nonlinear activation function $\sigma(\cdot)$ depends on the kernel type such that $\varphi(\bm x_i) := \sigma(\bm W \bm x_i)$ in Eq.~\eqref{mapping}, by denoting the transformation matrix $\bm W := [{\bm \omega}_1, {\bm \omega}_2, \cdots, {\bm \omega}_s]^{\!\top} \in \mathbb{R}^{s \times d}$.
The formulation in~\eqref{kernelfor} is quite general to cover a series of kernels by various activation functions.
For example, if we take $\sigma(x) = [\cos(x), \sin(x)]^{\!\top}$, Eq.~\eqref{kernelfor} corresponds to the Gaussian kernel, which is the standard RFF model \cite{rahimi2007random} for Gaussian kernel approximation.
If we consider the commonly used ReLU activation $\sigma(x) = \max \{ 0, x \}$ in neural networks, Eq.~\eqref{kernelfor} corresponds to the first order arc-cosine kernel, termed as $k(\bm x, \bm x') \equiv \kappa_{1}(u)=\frac{1}{\pi}(u(\pi-\arccos (u))+\sqrt{1-u^{2}})$ by setting $u:= \langle \bm x, \bm x^{\prime}\rangle/ ( \|\bm x\|\|\bm x^{\prime}\|)$.
If the Heaviside step function $\sigma(x) = \frac{1}{2}(1+\sign(x))$ is used, Eq.~\eqref{kernelfor} corresponds to the zeroth order arc-cosine kernel, termed as $k(\bm x, \bm x') \equiv \kappa_0(u) = 1 - \frac{1}{\pi} \arccos(u)$ by setting $u:= \langle \bm x, \bm x^{\prime}\rangle/ ( \|\bm x\|\|\bm x^{\prime}\|)$, refer to arc-cosine kernels \cite{cho2009kernel} for details.
If we take other activation functions used in neural networks, e.g., erf activations \cite{williams1997computing}, GELU \cite{hendrycks2016gaussian} in Eq.~\eqref{kernelfor}, such two-layer neural network also corresponds to a kernel.
In this case, the standard RFF model is still valid (via Monte Carlo sampling from a Gaussian distribution) for these non-stationary kernels.

Further, for a fully-connected deep neural network (more than two layers) and fixed random weights before the output layer, if the hidden layers are wide enough, one can still approach a kernel obtained by letting the widths tend to infinity \cite{daniely2016toward,lee2017deep}.
If both intermediate layers and the output layer are trained by (stochastic) gradient descent, for the network $f(\bm x; \bm \theta)$ with large enough $s$, the model remains close to its linearization around its random initialization throughout training, known as \emph{lazy training} regime \cite{chizat2019lazy}.
Learning is then equivalent to a kernel method with another architecture-specific kernel, known as \emph{neural tangent kernel} (NTK, \cite{jacot2018neural}).
Interestingly, NTK for two-layer ReLU networks \cite{bietti2019inductive} can be constructed by arc-cosine kernels, i.e., $k\left(\bm x, \bm x^{\prime}\right) = \|\bm x\| \|\bm x^{\prime} \| [  u\kappa_0(u) + \kappa_1 (u) ]$.
In fact, there is an interesting line of work showing insightful connections between kernel methods and (over-parameterized) neural networks, but this is out of scope of this survey on random features. We suggest the readers refer to some recent literature \cite{ghorbani2019linearized,bietti2020deep,arora2019exact} for details.

Further, if we consider the general non-stationary kernels \cite{genton2001classes,remes2017non}, the spectral representation can be generalized by introducing two random variables $\bm \omega$ and $\bm \omega'$.
\begin{theorem}\label{doublethe} (\cite{ton2018spatial,yaglom1987basic,remes2017non})
        A non-stationary kernel $k$ is positive definite if and only if it admits
        \begin{equation*}
                k(\bm x, \bm x^{\prime}) = \int_{\mathbb{R}^d \times \mathbb{R}^d} \exp\left(\mathrm{i} \left( {\bm \omega}^{\!\top}\bm x- {\bm \omega'}^{\!\top}\bm x' \right) \right) \mu_{\Psi_k} ( \mathrm{d} {\bm \omega}, \mathrm{d} {\bm \omega'} ) \,,
        \end{equation*}
        where $\mu_{\Psi_k}$ is the Lebesgue-Stieltjes measure on the product space $\mathbb{R}^d \times \mathbb{R}^d$ associated to some positive definite function $\Psi_k(\bm \omega, \bm \omega^{\prime})$ with bounded variations.
\end{theorem}

\subsection{Commonly used kernels in Random Features}
\label{sec:common_kernel}
Random features based algorithms often consider the following kernels:

i) Gaussian kernel: Arguably the most important member of shift-invariant kernels, the Gaussian kernel is given by
\begin{equation*}
	k(\bm{x}, \bm x')=\exp \left(-\frac{\|\bm{x}-\bm x' \|_{2}^{2}}{2 \varsigma^{2}} \right)\,,
\end{equation*}
where $\varsigma>0$ is the kernel width.
The density (see Theorem~\ref{bochner} or Eq.~\eqref{kernelfor}) associated with the Gaussian kernel is Gaussian $\bm \omega \sim \mathcal{N}(0, \varsigma^{-2} \bm I_d)$.

ii) arc-cosine kernels: This class admits Eq.~\eqref{kernelfor} by sampling from the Gaussian distribution $\mathcal{N}(0, \bm I_d)$, that can be connected to a two-layer neural networks with various activation functions. Following \cite{cho2009kernel}, we define the $b$-order arc-cosine kernel by
\begin{equation*}
	k(\bm x, \bm x') = \frac{1}{\pi} \| \bm x \|^b_2 \| \bm x' \|^b_2 J_b(\theta)\,,
\end{equation*}
where $\theta = \cos^{-1}\left( \frac{\bm x^{\!\top} \bm x'}{\| \bm x \|_2 \| \bm x' \|_2} \right)$ and
\begin{equation*}
	J_{b}(\theta)=(-1)^{b}(\sin \theta)^{2 b+1}\left(\frac{1}{\sin \theta} \frac{\partial}{\partial \theta}\right)^{b}\left(\frac{\pi-\theta}{\sin \theta}\right) \,.
\end{equation*}
Most common in practice are the zeroth order ($b=0$) and first order ($b=1$) arc-cosine kernels.
The zeroth order  kernel is given explicitly by
\begin{equation*}
	k(\bm x, \bm x') = 1 - \frac{\theta}{\pi} \,,
\end{equation*}
and the first order kernel is
\begin{equation*}
	k(\bm x, \bm x') =  \frac{1}{\pi} \| \bm x \|_2 \| \bm x' \|_2 \left(\sin \theta + (\pi - \theta) \cos \theta \right)\,.
\end{equation*}

iii) Polynomial kernel: This is a widely used  family of non-stationary kernels given by
\begin{equation*}
	k(\bm x, \bm x') = (1+ \langle \bm x, \bm x' \rangle)^b \,,
\end{equation*}
where $b$ is the order of the polynomial.

Note that, dot-product kernel defined in $\mathbb{R}^d$ admit neither \emph{spherical harmonics} nor Eq.~\eqref{kernelfor}. As a result, random features for polynomial kernels work in different theoretical foundations and settings, and have been studied in a smaller number of papers, including Maclaurin expansion \cite{kar2012random}, the tensor sketch technique \cite{Pham2013Fast,meister2019tight}, and oblivious subspace embedding \cite{avron2014subspace,woodruff2020near}.
Interestingly, if the data are $\ell_2$ normalized, dot product kernels defined in $\mathbb{R}^d$ can be transformed as stationary but indefinite (real, symmetric, but not positive definite) on the unit sphere\footnote{This setting cannot ensure the data are i.i.d on the unit sphere, which is different from the setting of previously discussed rotation invariant kernels.}.
The related random features based algorithms under this setting provide biased estimators \cite{pennington2015spherical,liu2019double}, or unbiased estimation \cite{liu2020fast}.

\subsection{Taxonomy of random features based algorithms}

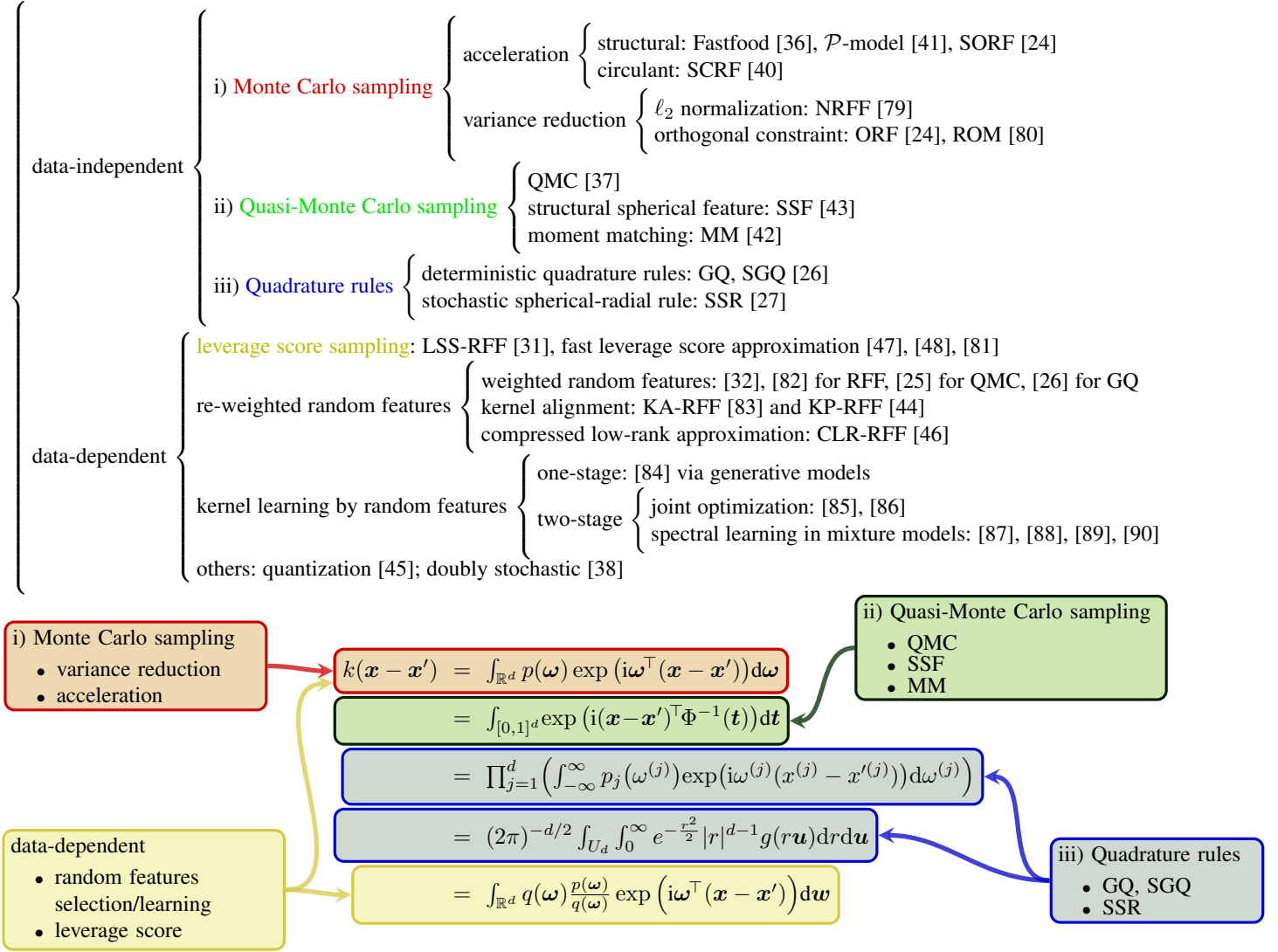
\begin{figure*}[!htb]
	\centering
	\AB{}
{
	\AB{data-independent}
	{
		\AB{i) {\color{red}Monte Carlo sampling}}
		{
			\AB{acceleration}
			{
				structural: Fastfood \cite{le2013fastfood}, $\mathcal{P}$-model \cite{choromanski2016recycling}, SORF \cite{Yu2016Orthogonal} \\
				circulant: SCRF \cite{feng2015random}
			}\\
		\AB{variance reduction}
		{
			$\ell_2$ normalization: NRFF \cite{li2017linearized} \\
			orthogonal constraint: ORF \cite{Yu2016Orthogonal}, ROM \cite{choromanski2017unreasonable}
		}
		}\\
		\AB{ii) \textcolor[RGB]{0,226,0}{Quasi-Monte Carlo sampling}}
		{
			QMC \cite{yang2014quasi} \\
			structural spherical feature: SSF \cite{lyu2017spherical}\\
			moment matching: MM \cite{shen2017random}
		}\\
	\AB{iii) {\color{blue}Quadrature rules}}
	{
		deterministic quadrature rules: GQ, SGQ \cite{dao2017gaussian} \\
		stochastic spherical-radial rule: SSR \cite{munkhoeva2018quadrature}
	}
	}\\
	\AB{data-dependent}
	{
		\textcolor[RGB]{187,187,0}{leverage score sampling}: LSS-RFF \cite{li2019towards}, fast leverage score approximation \cite{rudi2018fast,liu2020random,erdelyi2020fourier} \\
		\AB{re-weighted random features}
		{
			weighted random features: \cite{rahimi2009weighted,chang2017data} for RFF, \cite{Avron2016Quasi} for QMC, \cite{dao2017gaussian} for GQ \\
			kernel alignment: KA-RFF \cite{AmanNIPS2016} and KP-RFF \cite{shahrampour2018data}\\
			compressed low-rank approximation: CLR-RFF \cite{agrawal2019data}
		}\\
		\AB{kernel learning by random features}
		{
			one-stage: \cite{li2019implicit} via generative models \\
			\AB{two-stage}
			{
				joint optimization: \cite{Yu2015Compact,bullins2017not} \\
				spectral learning in mixture models: \cite{Wilson2013Gaussian,yang2015carte,shen2019harmonizable,oliva2016bayesian}
			} 
		}\\
	others: quantization \cite{zhang2019f}; doubly stochastic \cite{dai2014scalable}
	}
}
\begin{tikzpicture}{scale=0.2}
\matrix (m) [matrix of math nodes, row sep=2.5pt, column 3/.style={anchor=mid west}, column 2/.style={anchor=mid}, column 1/.style={anchor=mid east}]
{
	k(\bm x- \bm x') & =&  \int_{\mathbb{R}^d} p({\bm \omega}) \exp\big(\mbox{i}{\bm \omega}^{\!\top}(\bm x- \bm x')\big) \mbox{d} {\bm \omega}\\
	{\color{green!50!lime!30}k(\bm x- \bm x')} & =&  \int_{[0,1]^{d}} \! \exp\big(\mbox{i} (\bm x \!-\! \bm x')^{\!\top}\! \Phi^{-1}(\bm{t}) \big) \mbox{d} \bm t \\
	{\color{blue!50!lime!30} k(\bm x \!-\! \bm x')} & =& \prod_{j=1}^{d}\! \left( \int_{-\infty}^{\infty} p_j \big(\omega^{(j)} \big)\! \exp \! \big(\mathrm{i} \omega^{(j)} (x^{(j)} - x'^{(j)}) \big) \mathrm{d} \omega^{(j)} \right) \\
	{\color{blue!50!lime!30} k(\bm x - \bm x')} & =& (2\pi)^{-d/2} \int_{U_{d}} \int_{0}^{\infty} e^{-\frac{r^{2}}{2}}|r|^{d-1} g(r \bm u ) \mathrm{d} r \mathrm{d} \bm{u} \\
	{\color{yellow!50!lime!30} k(\bm x,\bm x')} & =& \int_{\mathbb{R}^d} {q(\bm \omega)} \frac{p(\bm \omega)}{{q(\bm \omega)}} \exp\Big(\mbox{i}\bm \omega^{\!\top}(\bm x- \bm x')\Big) \mbox{d} \bm w \\
};
\begin{scope}[on background layer, every node/.style={inner sep=0pt}]
\node (firsteq) [nodeStyleRed, fit=(m-1-1) (m-1-3)] {};
\node (sixdeq) [nodeStyleRed, fit=(m-1-1) (m-1-3)] {};
\node (secondeq) [nodeStyleGreen, fit=(m-2-1) (m-2-3)] {};
\node (thirdeq) [nodeStyleBlue, fit=(m-3-1) (m-3-3)] {};
\node (fourdeq) [nodeStyleBlue, fit=(m-4-1) (m-4-3)] {};
\node (fivedeq) [nodeStyleYellow, fit=(m-5-1) (m-5-3)] {};
\end{scope}
\node [nodeStyleRed, text width=37.5mm, left=of firsteq.mid west] (firsteq-aa)  {
	i) Monte Carlo sampling
	\begin{itemize}
	\item variance reduction
	\item acceleration
	\end{itemize}
};
\path [lineStyleRed] (firsteq-aa.east) edge[out=0,in=180,-stealth] (firsteq);
\node [nodeStyleGreen, text width=45mm, right=of firsteq.north east] (secondeq-aa) {
	ii) Quasi-Monte Carlo sampling
	\begin{itemize}
	\itemsep-0.25em
	\item QMC
	\item SSF
	\item MM
	\end{itemize}
};
\path [lineStyleGreen] (secondeq-aa.west) edge[out=180,in=0,-stealth] (secondeq);
\node [nodeStyleBlue, text width=30mm, below right=of thirdeq.mid east] (thirdeq-aa)  {
	iii) Quadrature rules 
	\begin{itemize}
	\itemsep-0.2em
	\item GQ, SGQ
	\item SSR
	\end{itemize}
};
\path [lineStyleBlue] (thirdeq-aa.west) edge[out=180,in=0,-stealth] (thirdeq);
\path [lineStyleBlue] (thirdeq-aa.west) edge[out=180,in=0,-stealth] (fourdeq);
\node [nodeStyleYellow, text width=40.5mm, left=of fivedeq.mid west] (fivedeq-aa)  {
	data-dependent
	\begin{itemize}
	\item random features selection/learning
	\item leverage score
	\end{itemize}
};
\path [lineStyleYellow] (fivedeq-aa.east) edge[out=0,in=180,-stealth] (fivedeq);
\path [lineStyleYellow] (fivedeq-aa.east) edge[out=0,in=183,-stealth] (firsteq);
\end{tikzpicture}
\caption{A taxonomy of representative random features based algorithms.
}
\vspace{-0.05cm}
\label{figtax}
\end{figure*}

The key step in random features based algorithms is constructing the following random feature mapping
\begin{equation}\label{phiweight}
\varphi(\bm x)  := \frac{1}{\sqrt{s}}
\big[a_1 \exp(-\mbox{i}{\bm \omega}^{\!\top}_1 \bm x), \cdots, a_s\exp(-\mbox{i}{\bm \omega}^{\!\top}_s \bm x)]^{\!\top}
\end{equation}
so as to approximate the integral~\eqref{rffdef}. Random features $\{\bm \omega_i \}_{i=1}^s$ can be formulated as the feature matrix $\bm W = [\bm \omega_1, \cdots, \bm \omega_s]^{\!\top} \in \mathbb{R}^{s \times d}$ in a compact form.
Existing algorithms differ in how they select the points ${\bm \omega}_i$ (the transformation matrix $\bm W$) and weights $a_i$.
Figure~\ref{figtax} presents a taxonomy of some representative random features based algorithms.
They can be grouped into two categories,  \emph{data-independent} algorithms and \emph{data-dependent} algorithms, based on whether or not  the selection of ${\bm \omega}_i$ and $a_i$ is independent of the training data.

Data-independent random features based algorithms can be further categorized into three classes according to their sampling strategy:

i) \emph{Monte Carlo sampling}: The points $\{ {\bm \omega}_i \}_{i=1}^s$ are sampled from $p(\cdot)$ in Eq.~\eqref{rffdef} (see the red box in Figure~\ref{figtax}).
In particular, to approximate the Gaussian kernel by RFF~\cite{rahimi2007random}, these points are sampled from the Gaussian distribution $p= \mathcal{N}(0, \varsigma^{-2} \bm I_d)$, with the weights being equal, i.e., $a_i \equiv  1$ in Eq.~\eqref{phiweight}.
To reduce the storage and time complexity, one may replace the dense Gaussian matrix in RFF by structural matrices; see, e.g., Fastfood \cite{le2013fastfood} using Hadamard matrices as well as its general version $\mathcal{P}$-model \cite{choromanski2016recycling}. An alternative approach is  using circulant matrices; see, e.g., Signed Circulant Random Features (SCRF) \cite{feng2015random}.
To improve the approximation quality, a simple and effective approach is to use an $\ell_2$-normalization scheme, which leads to Normalized RFF (NRFF) \cite{li2017linearized}. 
Another powerful technique for variance reduction is orthogonalization to decrease the randomness in Monte Carlo sampling. Typical algorithms include Orthogonal Random Features (ORF) \cite{Yu2016Orthogonal} by employing an orthogonality constraint to the random Gaussian matrix, Structural ORF (SORF) \cite{Yu2016Orthogonal,bojarski2017structured}, and Random Orthogonal Embeddings (ROM) \cite{choromanski2017unreasonable}.

ii) \emph{Quasi-Monte Carlo sampling}:
This is a typical sampling scheme in sampling theory \cite{niederreiter1992random} to reduce the randomness in Monte Carlo sampling for variance reduction.
It can significantly improve the convergence of Monte Carlo sampling by virtue of a low-discrepancy sequence $\bm t_1, \bm t_2, \cdots, \bm t_s \in [0,1]^d$ instead of a uniform sampling sequence over the unit cube to construct the sample points; see the integral representation in the green box in Figure~\ref{figtax}. 
Based on this representation, it can be used for kernel approximation, as conducted by \cite{Avron2016Quasi}. Subsequently, Lyu \cite{lyu2017spherical} proposes Spherical Structural Features (SSF), which generates asymptotically uniformly distributed points on $\mathbb{S}^{d-1}$ to achieve better convergence rate and approximation quality.
The Moment Matching (MM) scheme \cite{shen2017random} is based on the same integral representation but uses  a $d$-dimensional refined uniform sampling sequence $\{ \bm t_i \}_{i=1}^s$ instead of a low discrepancy sequence.
Strictly speaking, SSF and MM go beyond the QMC framework.
Nevertheless, these methods share the same integration formulation with QMC over the unit cube and thus we include them here for a streamlined presentation.

iii) \emph{Quadrature based methods}:
Numerical integration techniques can be also used to approximate the integral representation in Eq.~\eqref{rffdef}. These techniques may involve \emph{deterministic} selection of the points and weights, e.g., by using Gaussian Quadrature (GQ) \cite{dao2017gaussian} or Sparse Grids Quadrature (SGQ) \cite{dao2017gaussian} over each dimension (their integration formulation can be found in the first blue box in Figure~\ref{figtax}).
The selection can also be \emph{randomized}.
For example, in the work \cite{munkhoeva2018quadrature}, the $d$-dimensional integration in Eq.~\eqref{rffdef} is transformed to a double integral, and then approximated by using the Stochastic Spherical-Radial (SSR) rule (see the second blue box in Figure~\ref{figtax}).

Data-dependent algorithms use the training data to guide the selection of points and weights in the random features for better approximation quality and/or generalization performance.
These algorithms can be grouped into three classes according to how the random features are generated.

\emph{i) Leverage score sampling:} Built upon the importance sampling framework, this class of algorithm replaces the original distribution $p(\bm \omega)$ by a carefully chosen distribution $q(\bm \omega)$ constructed using leverage scores \cite{bach2017equivalence,avron2017random} (see the yellow box in Figure~\ref{figtax}).
The representative approach in this class is Leverage Score based RFF (LS-RFF) \cite{li2019towards}, and its accelerated version \cite{rudi2018fast,liu2020random}.

\emph{ii) Re-weighted random feature selection:} Here the basic idea is to re-weight the random features by solving a constrained optimization problem. Examples of this approach include  weighted RFF \cite{rahimi2009weighted,chang2017data}, weighted QMC \cite{Avron2016Quasi}, and weighted GQ \cite{dao2017gaussian}.
Note that these algorithms directly learn the weights of pre-given random features. Another line of methods re-weight the random features using a two-step procedure: i) ``up-projection": first generate a large set of random features $\{ \bm \omega_i \}_{i=1}^J$ ; ii) ``compression": then reduce these features to a small number (e.g., $10^2\sim 10^3$) in a data-dependent manner, e.g., by using kernel alignment \cite{AmanNIPS2016}, kernel polarization \cite{shahrampour2018data}, or compressed low-rank approximation \cite{agrawal2019data}.

\emph{iii) Kernel learning by random features:}
This class of methods aim to learn the spectral distribution of kernel \emph{from the data} so as to achieve better similarity representation and prediction. Note that these methods learn both the weights and the distribution of the features, and hence differ from the other random features selection methods mentioned above, which assume that the candidate features are generated from a pre-given distribution and only learn the weights of these features. Representative approaches for kernel learning involve a \emph{one-stage} \cite{li2019implicit} or \emph{two-stage} procedure \cite{Yu2015Compact,bullins2017not,Wilson2013Gaussian,yang2015carte,shen2019harmonizable,oliva2016bayesian}.
From a more general point of view, the aforementioned \emph{re-weighted random features selection} methods can also be classified into  this class. Since these methods belong to the broad area of kernel learning instead of kernel approximation, we do not detail them in this survey. 

Besides the above three main categories, other data-dependent approaches include the following. i) Quantization random features \cite{zhang2019f}: Given a memory budget, this method quantizes RFF for Gaussian kernel approximation.
A key observation from this work is that random features achieve better generalization performance than Nystr{\"o}m approximation \cite{Yang2012Nystr} under the same memory space.
ii) Doubly stochastic random features \cite{dai2014scalable}: This method uses two sources of stochasticity, one from sampling data points by stochastic gradient descent (SGD), and the other from using RFF to approximate the kernel. 
This scheme has been used for Kernel PCA approximation \cite{xie2015scale}, and can be further extended to triply stochastic scheme for multiple kernel approximation \cite{li2017triply}.

\section{Data-independent Algorithms}
\label{sec:data-independent}
\vspace{-0.0cm}
\begin{table*}[tp]
        \centering
        \fontsize{7}{8}\selectfont
        \begin{threeparttable}
                \caption{Comparison of different kernel approximation methods on space and time complexities to obtain $\bm W \bm x$.}
                \label{tabrff}
                \begin{tabular}{cccccccccccccccccccc}
                        \toprule
                        Method &Kernels (in theory) &Extra Memory  & Time & Lower variance than RFF  \cr
                        \midrule
                        Random Fourier Features (RFF) \cite{rahimi2007random} &shift-invariant kernels & $\mathcal{O}(sd)$ & $\mathcal{O}(sd)$ &- \cr
                        \midrule
                        Quasi-Monte Carlo (QMC) \cite{yang2014quasi} &shift-invariant kernels & $\mathcal{O}(sd)$ & $\mathcal{O}(sd)$ &Yes \cr
                        \midrule
                        Normalized RFF (NRFF) \cite{li2017linearized} &Gaussian kernel & $\mathcal{O}(sd)$ & $\mathcal{O}(sd)$ &Yes \cr
                        \midrule
                        Moment matching (MM) \cite{shen2017random} &shift-invariant kernels & $\mathcal{O}(sd)$ & $\mathcal{O}(sd)$ &Yes \cr
                        \midrule
                        Orthogonal Random Feature (ORF)\cite{Yu2016Orthogonal} &Gaussian kernel & $\mathcal{O}(sd)$ & $\mathcal{O}(s d)$ & Yes \cr
                        \midrule
                        Fastfood \cite{le2013fastfood} &Gaussian kernel & $\mathcal{O}(s)$ & $\mathcal{O}(s \log d)$ & No \cr
                        \midrule
                        Spherical Structured Features (SSF) \cite{lyu2017spherical}  &shift and rotation-invariant kernels & $\mathcal{O}(s)$ & $\mathcal{O}(s \log d)$ & Yes \cr
                        \midrule
                        Structured ORF (SORF) \cite{Yu2016Orthogonal,bojarski2017structured} &shift and rotation-invariant kernels & $\mathcal{O}(s)$  & $\mathcal{O}(s \log d)$ & Unknown \cr
                        \midrule
                        Signed Circulant (SCRF) \cite{feng2015random}  &shift-invariant kernels & $\mathcal{O}(s)$ & $\mathcal{O}(s \log d)$ & The same \cr
                        \midrule
                        $\mathcal{P}$-model \cite{choromanski2016recycling}  &shift and rotation-invariant kernels & $\mathcal{O}(s)$ & $\mathcal{O}(s \log d)$ & No \cr
                        \midrule
                        Random Orthogonal Embeddings (ROM) \cite{choromanski2017unreasonable} & rotation-invariant kernels & $\mathcal{O}(d)$ & $\mathcal{O}(d \log d)$ & Yes \cr
                        \midrule
                        Gaussian Quadrature (GQ), Sparse Grids Quadrature (SGQ)  \cite{dao2017gaussian} &shift invariant kernels & $\mathcal{O}(d)$ & $\mathcal{O}(d \log d)$ & Yes \cr
                        \midrule
                        Stochastic Spherical-Radial rules (SSR) \cite{munkhoeva2018quadrature} &shift and rotation-invariant kernels & $\mathcal{O}(d)$ & $\mathcal{O}(d \log d)$ & Yes \cr
                        \bottomrule
                \end{tabular}
        \end{threeparttable}
\end{table*}

In this section, we discuss data-independent algorithms in a unified framework based on the transformation matrix $\bm W $, that plays an important role in constructing the mapping $\varphi(\cdot)$ in Eq.~\eqref{phiweight} and determining how well the estimated kernel converges to the actual kernel.
Table~\ref{tabrff} reports various random features based algorithms in terms of the class of kernels they apply to (in theory) as well as their space and time complexities for computing the feature mapping $\bm W \bm x$ for a given $\bm x \in \mathcal{X}$.
In Table~\ref{tabrff}, we also summarize the \emph{variance reduction} properties of these algorithms, i.e., whether the variance of the resulting kernel estimator is smaller than the standard RFF.
Before proceeding, we introduce some notations and 
definitions.
When discussing a stationary kernel function $k(\bm x, \bm x')=k(\bm x - \bm x')$, we use the convenient shorthands $\bm \tau := \bm x - \bm x'$ and  $\tau :=\| \bm \tau \|_2$.
For a random features algorithm $A$ with frequencies $\{ \bm \omega_i \}_{i=1}^s$ sampled from a distribution $\mu(\cdot)$, we define its expectation $\mathbb{E}(A) := \mathbb{E}[k(\bm \tau)] = \mathbb{E}_{\bm \omega \sim \mu} \left[ 1/s \sum_{i=1}^s \cos(\bm \omega_i^{\!\top} \bm \tau) \right]$ and variance $\mathbb{V}[A]:= \mathbb{V}[k(\bm \tau)] = \mathbb{V}\left[ \frac{1}{s} \sum_{i=1}^s \cos(\bm \omega^{\!\top} \bm \tau) \right]$.

\subsection{Monte Carlo sampling based approaches}

We describe several representative data-independent algorithms based on Monte Carlo sampling, using  the Gaussian kernel $k(\bm x, \bm x') = k(\bm \tau) = \exp(-{\|\bm \tau \|^2_2}/{2 \varsigma^2})$ as an example.
Note that these algorithms often apply to more general classes of kernels, as summarized in Table~\ref{tabrff}.

{\bf RFF} \cite{rahimi2007random}: For Gaussian kernels, RFF directly samples the random features from a Gaussian distribution (corresponds to the inverse Fourier transform): $\{\bm \omega \}_{i=1}^s \sim p(\bm \omega)$. In particular, the corresponding transformation matrix is given by
\begin{equation}\label{wrff}
  {\bm W}_{\text{RFF}} = \frac{1}{\varsigma} \bm G \,,
\end{equation}
where $\bm G \in \mathbb{R}^{s \times d}$ is a (dense) Gaussian matrix with $G_{ij} \sim \mathcal{N}(0,1)$.
For other stationary kernels, the associated $p(\cdot)$ corresponds to the specific distribution given by the Bochner's Theorem. For example, the Laplacian kernel $k(\bm \tau) = \exp(-\| \bm \tau \|_1/\varsigma)$ is associated with a Cauchy distribution.
RFF is unbiased, i.e., $\mathbb{E}[\text{RFF}] = \exp(-{\| \bm \tau \|_2^2}/{2\varsigma^2})$, and the corresponding variance is $\mathbb{V}[\text{RFF}] = {(1-e^{-\tau^2})^2}/{2s}$.

{\bf Fastfood} \cite{le2013fastfood}: By observing the similarity between the dense Gaussian matrix and Hadamard matrices with diagonal Gaussian matrices, Le et al.~\cite{le2013fastfood} firstly introduce Hadamard and diagonal matrices to speed up the construction of dense Gaussian matrices in RFF, especially in high dimensions (e.g., $d \geq 1000$). In particular, $\bm W$ used in Eq.~\eqref{wrff} is substituted by
\begin{equation}\label{wfast}
{\bm W}_{\text{Fastfood}} = \frac{1}{\varsigma } \bm B_1 \bm H \bm G \bm \Gamma \bm H \bm B_2\,,
\end{equation}
where $\bm H$ is the Walsh-Hadamard matrix admitting fast multiplication in $\mathcal{O}(d \log d)$ time, and $\bm \Gamma \in \{0,1 \}^{d \times d}$ is a permutation matrix that decorrelates the eigen-systems of two Hadamard matrices.
The three \emph{diagonal} random matrices $\bm G$, $\bm B_1$  and $\bm B_2$ are specified as follows: $\bm G$ has independent Gaussian entries drawn from $ \mathcal{N}(0,1)$; $\bm B_1$ is a random scaling matrix with $(\bm B_1)_{ii}= \| \bm \omega_i \|_2 / \| \bm G \|_{\text{F}}$, which encodes the spectral properties of the associated kernel; $\bm B_2$ is a binary decorrelation matrix with  independent random $\{ \pm 1 \}$ entries.
FastFood is an unbiased estimator, but may have a larger variance than RFF: 
\begin{equation*}
	\mathbb{V}[\text{Fastfood}] - \mathbb{V}[\text{RFF}] \leq \frac{6 \tau^4}{s} \left(e^{-\tau^2}+\frac{\tau^2}{3}\right) \,,
\end{equation*}
which converges at an $\mathcal{O}(1/s)$ rate.

{\bf $\mathcal{P}$-model} \cite{choromanski2016recycling}: A general version of Fastfood, the $\mathcal{P}$-model constructs the transformation matrix as
\begin{equation*}
        \bm W_{\mathcal{P}} = [\bm g^{\!\top} \bm P_1,\bm g^{\!\top} \bm P_2, \cdots, \bm g^{\!\top} \bm P_s]^{\!\top} \in \mathbb{R}^{s \times d} \,,
\end{equation*}
where $\bm g$ is a Gaussian random vector of length $a$ and $\mathcal{P}=\{ \bm P_i \}_{i=1}^s$ is a sequence of $a$-by-$d$ matrices each  with unit $\ell_2$ norm columns.
Fastfood can viewed as a special case of the $\mathcal{P}$-model: the matrix $\bm H \bm G$ in Eq.~\eqref{wfast}  can be constructed by using a fixed budget of randomness in $\bm g$ and letting each $\bm P_i$ be a random diagonal matrix with diagonal entries of the form $H_{i1}, H_{i2},\ldots, H_{id}$.
The $\mathcal{P}$-model is unbiased and its variance is close to that of RFF with an $\mathcal{O}(1/d)$ convergence rate
\begin{equation*}
\Big| \mathbb{V}[\mbox{$\mathcal{P}$-model}] - \mathbb{V}[\text{RFF}] \Big| = \mathcal{O}\left({1}/{d}\right) \,.
\end{equation*}

{\bf SCRF} \cite{feng2015random}: It accelerates the construction of random features by using circulant matrices. The transformation matrix is
\begin{equation*}
\bm W_{\text{SCRF}} =  [ \bm \nu \otimes \mathcal{C}(\bm \omega_1),\bm \nu \otimes \mathcal{C}(\bm \omega_2), \cdots, \bm \nu \otimes \mathcal{C}(\bm \omega_t)]^{\!\top} \in \mathbb{R}^{td \times d} \,,
\end{equation*}
where $ \otimes $ denotes the tensor product, $\bm \nu = [\nu_1, \nu_2, \ldots, \nu_d]$ is a Rademacher vector with $\mathbb{P}(\nu_i=\pm 1) =1/2$, and $\mathcal{C}(\bm w_i) \in \mathbb{R}^{d \times d}$ is a circulant matrix generated by the vector $\bm \omega_i \sim \mathcal{N}(0, \varsigma^{-2} \bm I_d)$.
Thanks to the circulant structure, we only need $\mathcal{O}(s)$ space to store the feature mapping matrix $\bm W_{\text{SCRF}}$ with $s=td$.
Note that $\mathcal{C}(\bm w_i)$ can be diagonalized using the Discrete Fourier Transform for $\bm \omega_i$.
SCRF is unbiased and has the same variance as RFF.

The above three approaches are designed to accelerate the computation of RFF. We next overview representative methods that aim for better approximation performance than RFF.

{\bf NRFF} \cite{li2017linearized}: It normalizes the input data to have unit $\ell_2$ norm before constructing the random Fourier features. With normalized data, the Gaussian kernel can be computed as
\begin{equation*}
\begin{split}
k(\bm x, \bm x') = \exp \left( -\frac{1}{\varsigma^2} \bigg( 1 - \frac{\bm x^{\!\top} \bm x'}{\| \bm x\|_2 \| \bm x' \|_2} \bigg) \right)\,,
\end{split}
\end{equation*}
which is related to the normalized linear kernel \cite{li2017linearized,pennington2015spherical}.
Albeit simple, NRFF is effective in variance reduction and in particular satisfies
\begin{equation*}
	\mathbb{V}[\text{NRFF}] = \mathbb{V}[\text{RFF}] - \frac{1}{4s} e^{-\tau^2}(3 - e^{-2\tau^2})\,.
\end{equation*}

{\bf ORF} \cite{Yu2016Orthogonal}: It imposes orthogonality on random features for the Gaussian kernel and has the transformation matrix 
\begin{equation*}
	{\bm W}_{\text{ORF}} = \frac{1}{\varsigma} \bm S \bm Q \,,
\end{equation*}
where $\bm Q$ is a uniformly distributed random orthogonal matrix, and $\bm S$ is a diagonal matrix with diagonal entries sampled \emph{i.i.d} from the $\chi$-distribution with $d$ degrees of freedom.
This orthogonality constraint is useful in reducing the approximation error in random features. It is also considered in \cite{choromanski2019unifying} for unifying orthogonal Monte Carlo methods.
ORF is unbiased and with variance bounded by
\begin{equation*}
	\mathbb{V}[\text{ORF}] -  \mathbb{V}[\text{RFF}] \leq \frac{1}{s} \left( \frac{g(\tau)}{d} - \frac{(d-1)e^{-\tau^2}\tau^4}{2d} \right) \,,
\end{equation*}
where we have 
$g(\tau) = {e^{\tau^{2}}\left(\tau^{8}+6 \tau^{6}+7 \tau^{4}+\tau^{2}\right)}/{4}$ $+ {e^{\tau^{2}} \tau^{4}\left(\tau^{6}+2 \tau^{4}\right)}/{2 d}$.
It can be seen that the variance reduction property $\text{Var}[\text{ORF}] < \text{Var}[\text{RFF}]$ holds under some conditions, e.g., when $d$ is large and $\tau$ is small. For a large $d$, the ratio of the variances of ORF and RFF can be approximated by
\begin{equation}\label{stdorf}
	\frac{\mathbb{V}[\text{ORF}]}{\mathbb{V}[\text{RFF}]} \approx 1-\frac{(s-1) e^{-\tau^{2}} \tau^{4}}{d\left(1-e^{-\tau^{2}}\right)^{2}}\,.
\end{equation}
Choromanski et al. \cite{choromanski2018geometry} further improve the variance bound to
\begin{equation}\label{mseorf}
	\begin{split}
		\mathbb{V}[\text{RFF}] - & \mathbb{V}[\text{ORF}] = \\
		& {\frac{s-1}{s} \mathbb{E}_{R_{1}, R_{2}}\left[\frac{J_{\frac{d}{2}-1}(\sqrt{R_{1}^{2}+R_{2}^{2}}\tau) \Gamma(d / 2)}{(\sqrt{R_{1}^{2}+R_{2}^{2}}\tau / 2)^{\frac{d}{2}-1}}\right]} \\
		& -{\frac{s-1}{s} \mathbb{E}_{R_{1}}\left[\frac{J_{\frac{d}{2}-1}\left(R_{1}\tau \right) \Gamma(d / 2)}{\left(R_{1}\tau / 2\right)^{\frac{d}{2}-1}}\right]^{2}}\,,
	\end{split}
\end{equation}
where $J_{d}$ is the Bessel function of the first kind of degree $d$, and $R_1$ and $R_2$ are two independent scalar random variables satisfying $\bm \omega_1 = R_1 \bm v $ and $\bm \omega_2 = R_2 \bm v $ with $\bm \omega_1, \bm \omega_2 \sim \mathcal{N}(0, \varsigma^{-2} \bm I_d)$ and $\bm v \sim \text{Unif}(\mathcal{S}^{d-1})$.
According to Eq.~\eqref{mseorf}, the property $\mathbb{V}[\text{ORF}] < \mathbb{V}[\text{RFF}]$ holds asymptotically in cases: i) a fixed $d$ and a small enough $\tau$ with $\mathbb{E}[\| \bm \omega \|_2^4] \leq \infty$; ii) a fixed $\tau<\frac{1}{4\sqrt{c}}$ with some constant $c$ and a large $d$, in which case we have
\begin{equation*}
	\begin{split}
		\mathbb{V}[\text{RFF}] - \mathbb{V}[\text{ORF}] = \frac{s-1}{s} \left( \frac{1}{2d} \frac{\tau^4}{\varsigma^2} e^{-\frac{\tau^2}{\varsigma^2}} + \mathcal{O}\Big(\frac{1}{d}\Big)\right)\,.
	\end{split}
\end{equation*}

{\bf SORF} \cite{Yu2016Orthogonal,bojarski2017structured}: It replaces the random orthogonal matrices used in ORF by a class of structured matrices akin to those in Fastfood. The transformation matrix of SORF is given by
\begin{equation}\label{sorfm}
\bm{W}_{\text{SORF}}=\frac{\sqrt{d}}{\varsigma} \bm{H D}_{1} \bm{H D}_{2} \bm{H D}_{3}\,,
\end{equation}
where $\bm H$ is the normalized Walsh-Hadamard matrix and $\bm D_i \in \mathbb{R}^{d \times d}$, $i=1,2,3$ are diagonal “sign-flipping” matrices, of which each diagonal entry is sampled from the Rademacher distribution.
Bojarski et al. \cite{bojarski2017structured} consider more general structures for the three blocks of matrices $\bm H \bm D_i$ in Eq.~\eqref{sorfm}.
Note that each block plays a different role.
The first block $\bm H \bm D_1$ satisfies $\text{Pr}\left[\| \bm H \bm D_1 \bm x \|_{\infty} > \frac{\log d}{\sqrt{d}}\right] \leq 2de^{-\frac{\log^2 d}{8}}$ for any $\bm x \in \mathbb{R}^d$ with $\| \bm x \|_2 = 1$, termed as $(\log d, 2de^{-\frac{\log^2 d}{8}})$-balanced, hence no dimension carries too much of the $\ell_2$ norm of the vector $\bm x$.
The second block $\bm H \bm D_2$ ensures that vectors are close to orthogonal. The third block $\bm H \bm D_3$ controls the capacity of the entire structured transform by providing a vector of parameters.
SORF is not an unbiased estimator of the Gaussian kernel, but it satisfies an asymptotic unbiased property
\begin{equation*}
	\left|\mathbb{E}\left[ {\text{SORF}} \right]-e^{-\tau^{2} / 2}\right| \leq \frac{6 \tau}{\sqrt{d}}\,.
\end{equation*}

{\bf ROM} \cite{choromanski2017unreasonable}: It generalizes SORF to the form
\begin{equation*}\label{romm}
\bm{W}_{\text{ROM}}=\frac{\sqrt{d}}{\varsigma} \prod_{i=1}^t \bm H \bm D_i\,,
\end{equation*}
where $\bm H$ can be the normalized Hadamard matrix or the Walsh matrix, and $\bm D_i$ is the Rademacher matrix as defined in SORF.
Theoretical results in \cite{choromanski2017unreasonable} show that the ROM estimator achieves variance reduction compared to RFF.
Interestingly, odd values of $t$ yield better results than even $t$. 
This provides an explanation for why SORF chooses $t=3$.

\noindent {\bf LP-RFF} \cite{zhang2019f}: It attempts to quantize RFF with the Gaussian kernel under a memory budget, i.e., mapping each $s$-dimensional random feature $z_{p}(\bm x) = \sqrt{2/s}\cos (\bm W_{\text{RFF}} \bm x ) \in [-\sqrt{2/s}, \sqrt{2/s}]$ to an $s$-dimensional low precision vector with $b$ bits via a stochastic rounding scheme.
They divide the interval $[-\sqrt{2/s}, \sqrt{2/s}]$ into $2^b-1$ equal-sized sub-intervals and randomly round each value $\sqrt{2/s}\cos (\bm \omega_i \bm x )$ to either the top or bottom of the corresponding sub-interval.
Strictly speaking, this method does not belong to data-independent algorithms.
But we put it here for ease of description as this approach directly quantizes RFF.
More importantly, a new insight demonstrated by this method is that, under the same memory budget, random features based algorithms achieve better generalization performance than Nystr{\"o}m approximation \cite{Yang2012Nystr}.
Apart from the stochastic quantization scheme used in \cite{zhang2019f}, the authors of  \cite{li2021quantization} employ Lloyd-Max quantization with a smaller number of bits.

From the above description, one can find that orthogonalization is a typical operation for variance reduction, e.g., ORF/SORF/ROM. Here we take the Gaussian kernel as an example to illustrate insights of such scheme. By sampling $\{ \bm \omega_i \}_{i=1}^s \sim \mathcal{N}(\bm 0, \varsigma^{-2}\bm I_d)$, the used Gaussian distribution is isotropic and only depends on the norm $\| \bm \omega\|_2$ instead of $\bm \omega$. The used orthogonal operator makes the direction of $\bm \omega_i$ orthogonal to each other (that means more uniform) while retaining its norm unchanged\footnote{In fact, while orthogonalization only makes the direction of $\{ \bm \omega_i \}_{i=1}^s$ more uniform, one can make the length $\| \bm \omega_i \|_2$ uniform by sampling from the cumulative distribution function of $\| \bm \omega\|_2$.}, which leads to decrease the randomness in Monte Carlo sampling, and thus achieve variance reduction effect.
If we attempt to directly decrease the randomness in Monte Carlo sampling, QMC is a powerful way to achieve this goal and can then be used to kernel approximation. This is another line of random features with variance reduction illustrated as below.

\subsection{Quasi-Monte Carlo Sampling}

Here we briefly review methods based on quasi-Monte Carlo sampling (QMC) \cite{yang2014quasi}, spherical structured feature (SSF) \cite{lyu2017spherical}, and moment matching (MM) \cite{shen2017random}.
These three methods achieve a lower variance or approximation error than RFF.
Strictly speaking, the later two algorithms do not belong to the quasi-Monte Carlo sampling framework.
However, SSF and MM share the same integration formulation with QMC and thus we introduce them here for simplicity.

Classical Monte Carlo sampling generates a sequence of samples randomly and independently, which may lead to an undesired clustering effect and empty spaces between the samples \cite{niederreiter1992random}.
Instead of fully random samples, QMC \cite{yang2014quasi} outputs low-discrepancy sequences.
A typical QMC sequence has a hierarchical structure: the initial
points are sampled  on a coarse scale whereas the subsequent points are sampled more finely.
For approximating a high-dimensional integral, QMC achieves an asymptotic error convergence rate of $\epsilon = \mathcal{O}((\log s)^d/s)$, which is faster than the $\mathcal{O}(s^{-1/2})$ rate of Monte Carlo.
Note however that QMC often requires $s$ to be exponential in $d$ for the improvement to manifest.

{\bf QMC} \cite{yang2014quasi}: It assumes that $p(\cdot)$ factorizes with respect to the dimensions, i.e., $p(\bm{x})=\prod_{j=1}^{d} p_{j}\left(x_{j}\right)$, where each $p_j(\cdot)$ is a univariate density function.
QMC generally transforms an integral on $\mathbb{R}^d$ in Eq.~\eqref{rffdef} to one on the unit cube $[0,1]^d$ as
\begin{equation}\label{qmceq}
k(\bm x - \bm x') =  \int_{[0,1]^{d}} \exp\big(\mbox{i} (\bm x- \bm x')^{\!\top} \Phi^{-1}(\bm{t}) \big) \mbox{d} \bm t \,,
\end{equation}
where $ \Phi^{-1}(\bm{t})=\left(\Phi_{1}^{-1}\left(t_{1}\right), \cdots, \Phi_{d}^{-1}\left(t_{d}\right)\right) \in \mathbb{R}^{d} $
with $\Phi_j$ being the cumulative distribution function (CDF) of $p_j$.
Accordingly, by generating a low \emph{discrepancy} sequence $\bm t_1, \bm t_2, \cdots, \bm t_s \in [0,1]^d$, the random frequencies can be constructed by ${\bm \omega}_i = \Phi^{-1}(\bm{t}_i)$. 
The corresponding transformation matrix for QMC is
\begin{equation}\label{QMCmapping}
        \bm W_{\text{QMC}} = [\Phi^{-1}(\bm{t}_1), \Phi^{-1}(\bm{t}_2), \cdots, \Phi^{-1}(\bm{t}_s)]^{\!\top} \in \mathbb{R}^{s \times d}\,.
\end{equation}

{\bf SSF} \cite{lyu2017spherical}: It improves the space and time complexities of QMC for approximating shift- and rotation-invariant kernels. SSF generates points $\{\bm v_1, \bm v_2, \cdots, \bm v_s \} $ asymptotically uniformly distributed on the sphere $\mathbb{S}^{d-1}$, and construct the transformation matrix as 
\begin{equation*}
	\bm W_{\text{SSF}} = [\Phi^{-1}(t)\bm v_1, \Phi^{-1}(t) \bm v_2, \cdots, \Phi^{-1}(t) \bm v_s]^{\!\top} \in \mathbb{R}^{s \times d}\,,
\end{equation*}
where $\Phi^{-1}(t)$ uses the one-dimensional QMC point.
The structure matrix $\bm V  := [\bm v_1, \bm v_2, \cdots, \bm v_s] \in \mathbb{S}^{(d-1) \times s}$ has the form
\begin{equation*}
	\bm{V}=\frac{1}{\sqrt{d/2}}\left[\begin{array}{cc}{\operatorname{Re} \bm F_{\Lambda}} & {-\operatorname{Im} \bm F_{\Lambda}} \\ {\operatorname{Im} \bm F_{\Lambda}} & {\operatorname{Re} \bm F_{\Lambda}}\end{array}\right] \in \mathbb{R}^{d \times s} \, ,
\end{equation*}
where $\bm F_{\Lambda} \in \mathbb{C}^{\frac{d}{2} \times \frac{s}{2}}$ consists of a subset of the rows of the discrete Fourier matrix $\bm F \in \mathbb{C}^{\frac{s}{2} \times \frac{s}{2}}$.
The selection of $\frac{d}{2}$ rows from $\bm F$  is done by minimizing the discrete Riesz 0-energy \cite{brauchart2015distributing} such that the points spread as evenly as possible on the sphere.

{\bf MM} \cite{shen2017random}: It also uses the transformation matrix in Eq.~\eqref{QMCmapping}, but  generates a $d$-dimensional uniform sampling sequence $\{ \bm t_i \}_{i=1}^s$ by a moment matching scheme instead of using a low discrepancy sequence as in QMC.
In particular, the transformation matrix is
\begin{equation}\label{mmmapping}
\bm W_{\text{MM}} = [\widetilde{\Phi}^{-1}(\bm{t}_1), \widetilde{\Phi}^{-1}(\bm{t}_2), \cdots, \widetilde{\Phi}^{-1}(\bm{t}_s)]^{\!\top} \in \mathbb{R}^{s \times d}\,,
\end{equation}
where one uses moment matching to construct the vectors
 $   \widetilde{\Phi}^{-1}(\bm{t}_i) = \tilde{\bm A}^{-1} (\Phi^{-1}(\bm{t}_i) - \tilde{\bm \mu})$ with the sample mean $\tilde{\bm \mu} = \frac{1}{s}\sum_{i=1}^{s} \Phi^{-1}(\bm{t}_i)$ and the square root of the sample covariance matrix $\tilde{\bm A}$ satisfying $\tilde{\bm A} \tilde{\bm A}^{\!\top} = \text{Cov}(\Phi^{-1}(\bm{t}_i) - \tilde{\bm \mu})$.

To achieve the target of variance reduction, both orthogonalization in Monte Carlo sampling and QMC based algorithms share the similar principle, namely, generating random features that are as independent/uniform as possible.
To be specific, QMC and MM are able to generate more uniform data points to avoid undesirable \emph{clustering} effect, see Figure~1 in \cite{yang2014quasi}. Likewise, SSF aims to generate asymptotically uniformly distributed points on the sphere $\mathbb{S}^{d-1}$, which attempts to encode more information with fewer random features, and thus allows for variance reduction.
In sampling theory, QMC can be further improved by an sub-grouped based rank-one lattice construction \cite{lyu2020subgroup} for computational efficiency, which can be used for the subsequent kernel approximation.

\subsection{Quadrature based Methods}
Quadrature based methods build on a long line of work on numerical quadrature for estimating integrals.
In quadrature methods, the weights are often non-uniform, and the points are usually selected using \emph{deterministic} rules including Gaussian quadrature (GQ) \cite{evans1993practical,dao2017gaussian} and sparse grids quadrature (SGQ) \cite{dao2017gaussian}.
Deterministic rules can be extended to their stochastic versions. For example, Munkhoeva et al. \cite{munkhoeva2018quadrature} explore the stochastic spherical-radial (SSR) rule \cite{genz1998stochastic,genz1999stochastic} in kernel approximation. Below we briefly review these methods.

{\bf GQ} \cite{dao2017gaussian}: It assumes that the kernel function $k$ factorizes with respect to the dimensions and the corresponding distribution $p({\bm \omega}) = p([ \omega^{(1)}, \omega^{(2)}, \ldots, \omega^{(d)}]^{\!\top} )$ in Eq.~\eqref{rffdef} is sub-Gaussian. Therefore, the $d$-dimenionsal integral in Eq.~\eqref{rffdef} can be factorized as
\begin{equation}\label{rffdefga}
	\begin{split}
		k(\bm x \!-\! \bm x') \! =\! \prod_{j=1}^{d}\! \left( \int_{-\infty}^{\infty} p_j \big(\omega^{(j)} \big)\! \exp \! \big(\mathrm{i} \omega^{(j)} (x^{(j)} - x'^{(j)}) \big) \mathrm{d} \omega^{(j)} \!\right) \!.
	\end{split}
\end{equation}
Since each of the factors is a one-dimensional integral, we can approximate them using a one-dimensional quadrature rule. For example, one may use  Gaussian quadrature \cite{evans1993practical} with orthogonal polynomials:
\begin{equation}\label{gaqu}
	\int_{-\infty}^{\infty} p(\omega) \exp(\mathrm{i} \omega (x - x') ) \mathrm{d} \omega \approx \sum_{j=1}^{L} a_j \exp\big(\mathrm{i} \gamma_j^{\!\top}(x- x')\big)\,,
\end{equation}
where $L$ is the accuracy level and each $\gamma_j$ is a univariate point associated with the weight $a_j$.
For a third-point rule with the points $\left\{-\hat{p}_{1}, 0, \hat{p}_{1}\right\}$ and their associated weights $(\hat{a}_1,\hat{a}_0,\hat{a}_1)$, the transformation matrix $\bm W_{\text{GQ}} \in \mathbb{R}^{s \times d}$ has entries  $W_{ij}$ following the distribution
\begin{equation*}
	\operatorname{Pr} \left( W_{ij} \!=\! \pm \hat{p}_{1} \right) \!=\! \hat{a}_1,~ \operatorname{Pr} \left( W_{ij} \!=\! 0 \right) \!=\!  \hat{a}_0,~\forall i \in \! [s], j \in \! [d]\,.
\end{equation*}
In general, the univariate Gaussian quadrature with $L$ quadrature points is exact for polynomials up to $(2L-1)$ degrees.
The multivariate Gaussian quadrature is exact for all polynomials of the form $\omega_1^{i_1} \omega_2^{i_2} \cdots \omega_d^{i_d}$ with $1 \leq i_j \leq 2L -1 $; however the total number of points $s=L^d$ scales exponentially with the dimension $d$ and thus this method suffers from the curse of dimensionality.

{\bf SGQ} \cite{dao2017gaussian}: To alleviate the curse of dimensionality, SGQ uses the Smolyak rule \cite{heiss2008likelihood} to decrease the needed number of points.
Here we consider the third-degree SGQ using the symmetric univariate quadrature points $\left\{-\hat{p}_{1}, 0, \hat{p}_{1}\right\}$ with weights $(\hat{a}_1,\hat{a}_0,\hat{a}_1)$:
\begin{equation*}
	k(\bm x, \bm x') \! \approx\! \left(1\!-\!d\!+\!d \hat{a}_{0}\right) g(\bm{0})+\hat{a}_{1} \sum_{j=1}^{d}\!\big[g\left(\hat{p}_{1} \bm{e}_{j}\right)\!+\!g\left(-\hat{p}_{1} \bm{e}_{j}\right)\!\big]\,,
\end{equation*}
where the function $g(\bm \omega):= \sigma(\bm \omega^{\top} \bm x) \sigma(\bm \omega^{\!\top} \bm x')$ is given by Eq.~\eqref{kernelfor}, and $\bm e_i$ is the $d$-dimensional standard basis vector with the $i$-th element being~1.
The corresponding transformation matrix is
\begin{equation*}
	\bm W_{\text{SGQ}} \!=\! [ \bm 0_d, \hat{p}_{1} \bm{e}_{1}, \cdots, \hat{p}_{1} \bm{e}_{d}, -\hat{p}_{1} \bm{e}_{1}, \cdots, -\hat{p}_{1} \bm{e}_{d}  ]^{\!\top} \! \in \! \mathbb{R}^{(2d+1) \times d} \!,
\end{equation*}
which leads to the explicit feature mapping
\begin{equation*}
	\varphi(\bm x) = [\hat{a}_{0} g(\bm 0), \hat{a}_{1} g(\bm w_2^{\!\top} \bm x), \cdots, g(\bm w_{2d+1}^{\!\top} \bm x)] \,,
\end{equation*}
where $\bm w_i$ is the $i$-th row of $\bm W_{\text{SGQ}}$.
Note that SGQ generates  $2d+1$ points.
To obtain a \emph{dimension-adaptive} feature mapping, Dao et al. \cite{dao2017gaussian} propose to subsample the points according to the distribution determined by their weights such that the mapping feature dimension is equal to $s$.

{\bf SSR} \cite{munkhoeva2018quadrature}: It transforms Eq.~\eqref{kernelfor} (actually a $d$-dimensional integral) to a double integral over a hyper-sphere and the real line.
Let ${\bm \omega} = r \bm u$ with $\bm u^{\!\top} \bm u = 1$ for $r \in [0, \infty)$, we have
\begin{equation}\label{doubint}
\begin{split}
k(\bm x - \bm x') =\frac{C_d}{2} \int_{\mathbb{S}^{d-1}} \int_{-\infty}^{\infty} e^{-\frac{r^{2}}{2}}|r|^{d-1} g(r \bm u ) \mathrm{d} r \mathrm{d} \bm{u}\,,
\end{split}
\end{equation}
where the integrand is $g(\bm \omega):= \sigma(\bm \omega^{\top} \bm x) \sigma(\bm \omega^{\!\top} \bm x')$ given in Eq.~\eqref{kernelfor} and $C_d := (2\pi)^{-d/2}$.
The inner integral in Eq.~\eqref{doubint} can be approximated by stochastic \emph{radial} rules of degree $2l+1$, i.e., $R(g)=\sum_{i=0}^{l} \hat{w}_{i} \frac{g\left(\rho_{i}\right)+g\left(-\rho_{i}\right)}{2}$.
The outer integral over the  $d$-sphere in Eq.~\eqref{doubint} can be approximated by stochastic \emph{spherical} rules: $S_{\mathrm{Q}}(g)=\sum_{j=1}^{q} \widetilde{w}_{j} g \left(\bm{Q} \bm{u}_{j}\right)$, where $\bm Q$ is a random orthogonal matrix and $\widetilde{w}_{j}$ are stochastic weights whose distributions are such that the rule is exact for polynomials of degree $q$ and gives unbiased estimate for other functions.
Combining the above two rules, we have the SSR rule.
Accordingly, the transformation matrix of SSR is 
\begin{equation*}
	\bm W_{\text{SSR}} = \bm \vartheta \otimes \left[\begin{array}{r}
	(\bm{QV})^{\top} \\
	-(\bm{Q} \bm{V})^{\top}
	\end{array}\right] \in \mathbb{R}^{2(d+1) \times d}\,,
\end{equation*}
with $\bm \vartheta = [\vartheta_1, \vartheta_2, \cdots, \vartheta_s]$ and $\bm V = [\bm v_1, \bm v_2, \cdots, \bm v_{d+1} ]$,
where $\vartheta \sim \chi(d+2)$ and $\{\bm v_i\}_{i=1}^{d+1}$ are the vertices of a unit  regular $d$-simplex, which is randomly rotated by $\bm Q$.
To get $s$ features, one may stack $s/(2d+3)$ independent copies of $\bm W$ as suggested by \cite{munkhoeva2018quadrature}.
Finally, the feature mapping by SSR is given by
\begin{equation*}
	\varphi(\bm x) = [{a}_{0} g(\bm 0), {a}_{1} g(\bm w_1^{\!\top} \bm x), \cdots, a_s g(\bm w_{s}^{\!\top} \bm x)] \,,
\end{equation*}
where $a_{0}=\sqrt{1-\sum_{j=1}^{d+1} \frac{d}{\rho_j^{2}}}$, $a_{j}=\frac{1}{\rho_{j}} \sqrt{\frac{d}{2(d+1)}}$ for $j \in [s]$, and $w_j$ is the $j$-th element of the stacked $\bm W$.

In general, according to Eq.~\eqref{kernelfor}, kernel approximation by random features is actually a $d$-dimensional integration approximation problem in mathematics. Sampling methods and quadrature based rules are two typical classes of approaches for high-dimensional integration approximation. Efforts on quadrature based methods focus on developing a high-accuracy, mesh-free, efficiency rule, e.g., \cite{belhadji2019kernel,liu2020towards}. 
Note that, if the integrand $g(\bm \omega) := \sigma(\bm \omega^{\top} \bm x) \sigma(\bm \omega^{\!\top} \bm x')$ in the integration representation~\eqref{kernelfor} belongs to a RKHS, the above quadrature rules can be termed as kernel-based quadrature, e.g., Bayesian quadrature \cite{briol2017sampling,gauthier2018optimal} and leverage-score quadrature \cite{bach2017equivalence}. This approach is in essence different from the previously studied quadrature rules in functional spaces, model formulation, and scope of application.

\vspace{-0.3cm}
\section{Data-dependent algorithms}
\label{sec:data-dependent}

Data-dependent approaches aim to design/learn the random features using the training data so as to achieve better approximation quality or generalization performance.
Based on how the random features are generated, we can group these algorithms into three classes: \emph{leverage score sampling}, \emph{random features selection}, and \emph{kernel learning by random features}.

\vspace{-0.0cm}
\subsection{Leverage score based sampling}

Leverage score based approaches \cite{li2019towards,liu2020random,wang2019general} are built on the \emph{importance sampling} framework.
Here one samples $\{ \bm w_i \}_{i=1}^s$  from a distribution $q(\bm w)$ that needs to be designed, and then uses the following feature mapping in Eq.~\eqref{mapping}:
\begin{equation}\label{fmaping}
\varphi_q(\bm x) = \frac{1}{\sqrt{s}} \left(\sqrt{\frac{p\left(\bm w_1\right)}{q\left(\bm w_1\right)}}  e^{-\mathrm{i}\bm w^{\!\top}_1 \bm x}, \cdots, \sqrt{\frac{p\left(\bm w_s\right)}{q\left(\bm w_s\right)}}  e^{-\mathrm{i}\bm w^{\!\top}_s \bm x}\right)^{\!\!\top}\,.
\end{equation}
Consequently, we have the approximation
$k(\bm x, \bm x')=\mathbb{E}_{\bm{w} \sim q } [\varphi_q( \bm x)^{\!\top} \varphi_q(\bm x')] 
    \approx \sum_{i=1}^{s} z_q(\bm w_i, \bm x) z_q(\bm w_i, \bm x')$, 
where $z_q(\bm w_i, \bm x_j) := \sqrt{p(\bm w_i)/ q(\bm w_i)} z_p(\bm w_i, \bm x_j).$
Thus, the kernel matrix $\bm K$ can be approximated by $\bm K_q = \bm Z_q \bm Z_q^{\!\top}$, where $\bm Z_q :=[\varphi_q(\bm x_1), \cdots, \varphi_q(\bm x_n)]^{\!\top} \in \mathbb{R}^{n \times s}$.
Denoting by $\bm z_{q,\bm w_i}(\bm X)$  the $i$-th column of $\bm Z_q$,
we have $\bm K = \mathbb{E}_{\bm w \sim p}[\bm z_{p,\bm w}(\bm X) \bm z^{\!\top}_{p,\bm w}(\bm X)]=\mathbb{E}_{\bm w \sim q}[\bm z_{q,\bm w}(\bm X) \bm z^{\!\top}_{q,\bm w}(\bm X)]$.

To design the distribution $q$, one makes use of  the ridge leverage function \cite{bach2017equivalence,avron2017random} in KRR:\vspace{-0.1cm}
\begin{equation}\label{llambda}
l_{\lambda}({\bm \omega}_i)=p({\bm \omega}_i) \bm z^{\!\top}_{p,\bm{\omega}_i}(\bm{X})(\bm{K}+n \lambda \bm{I})^{-1} \bm z_{p,\bm{\omega}_i}(\bm{X})\,,
\end{equation}
where $\lambda$ is the KRR regularization parameter.
Define\vspace{-0.1cm}
\begin{equation}\label{dklambda}
d_{\bm{K}}^{\lambda} := \int_{\mathbb{R}^d} l_{\lambda}({\bm \omega}) \mbox{d} {\bm \omega} = \operatorname{tr}\left[\bm{K}(\bm{K}+n \lambda \bm{I})^{-1}\right] \,.
\end{equation}
The quantity $d_{\bm{K}}^{\lambda} \ll n$ determines the number of independent parameters in a learning problem and hence is referred to as the \emph{number of effective degrees of freedom} \cite{zhang2005learning,bach2013sharp}.
With the above notation, the distribution $q$ designed in \cite{avron2017random} is given by
\begin{equation}\label{qwdef}
        q({\bm \omega}):= \frac{l_{\lambda}({\bm \omega}) }{\int l_{\lambda}({\bm \omega}) \text{d} {\bm \omega} } = \frac{l_{\lambda}({\bm \omega}) }{d_{\bm{K}}^{\lambda} }\,.
\end{equation}
Compared to standard Monte Carlo sampling for RFF, leverage score  sampling requires fewer Fourier features and enjoys nice theoretical guarantees \cite{avron2017random,li2019towards} (see the next section for details).
Note that $q(\bm \omega)$ can be also defined by the integral operator \cite{bach2017equivalence,yamasaki2020fast} rather than the Gram matrix used above, but
we do not strictly distinguish these two cases. The typical leverage score based sampling algorithm for RFF is illustrated in \cite{li2019towards} as below.

{\bf LS-RFF} (Leverage Score-RFF) \cite{li2019towards}: It uses a subset of data to approximate the matrix $\bm K$ in Eq.~\eqref{dklambda} so as to compute $d_{\bm{K}}^{\lambda}$.
LS-RFF needs $\mathcal{O}(ns^2+s^3)$ time to generate refined random features, which can be used in KRR \cite{li2019towards} and SVM \cite{sun2018but} for prediction.

{\bf SLS-RFF} (Surrogate Leverage Score-RFF) \cite{liu2020random}: To avoid inverting an $s \times s$ matrix in LS-RFF, SLS-RFF designs a simple but effective surrogate leverage function
\begin{equation}\label{lw}
L_{\lambda}(\bm w)=p(\bm w) \bm z^{\!\top}_{p,\bm{w}}(\bm{X}) \left( \frac{1}{n^2 \lambda} \Big(\bm y \bm y^{\!\top}+n \bm I\Big) \right) \bm z_{p,\bm{w}}(\bm{X})\,,
\end{equation}
where the additional term $n \bm I$ and the coefficient $1/(n^2 \lambda)$ in Eq.~\eqref{lw} ensure that $ L_{\lambda} $ is a \emph{surrogate} function that upper bounds the function $ l_{\lambda} $ in Eq.~\eqref{llambda}.
One then samples random features from the \emph{surrogate} distribution $Q(\bm w) = \frac{L_{\lambda}({\bm \omega}) }{\int L_{\lambda}({\bm \omega}) \text{d} {\bm \omega} } $, which has the same time complexity $\mathcal{O}(ns^2)$ as RFF. SLS-RFF and can be applied to KRR \cite{liu2020random} and Canonical Correlation Analysis \cite{wang2019general}.

Note that leverage scores sampling is a powerful tool used in sub-sampling algorithms for approximating large kernel matrices with theoretical guarantees, in particular in Nystr\"{o}m approximation.
Research on this topic mainly focuses on obtaining fast leverage score approximation due to inversion of an $n$-by-$n$ kernel matrix, e.g., two-pass sampling \cite{alaoui2015fast} (LS-RFF belongs to this), online setting \cite{calandriello2017distributed}, path-following algorithm \cite{rudi2018fast}, or developing various surrogate leverage score sampling based algorithms \cite{liu2020random,wang2019general,erdelyi2020fourier}.

\subsection{Re-weighted random features}
Here we briefly review three re-weighted methods: KA-RFF \cite{AmanNIPS2016} by kernel alignment, KP-RFF \cite{shahrampour2018data} by kernel polarization, and CLR-RFF \cite{agrawal2019data} by compressed low-rank approximation.

{\bf KA-RFF} (Kernel Alignment-RFF) \cite{AmanNIPS2016}: It pre-computes a large number of random features that are generated by RFF, and then select a subset of them by solving a simple optimization problem based on kernel alignment \cite{Cortes2010Two}. In particular, the  optimization problem is
\begin{equation}\label{kap}
	\underset{\bm a \in \mathcal{P}_{J}}{\max}~~ \sum_{i,j=1}^n y_{i} y_{j} \sum_{t=1}^{J} a_{t} z_p\left(\bm x_{i}, \bm \omega_{t}\right) z_p\left(\bm x_{j}, \bm \omega_{t}\right)\,,
\end{equation}
where $J>s$ is the number of the candidate random features by RFF, and $\bm a$ is the weight vector.
Here the maximization is over the set of distributions $\mathcal{P}_{J} := \{ \bm a: D_f(\bm a \| \bm 1/J) \leq c \}$, where $c>0$  is a pre-specified constant and $D_f ({P} \| {Q}) := \int f(\frac{\mathrm{d}{P}}{\mathrm{d}{Q}}) \mathrm{d}{Q}$ with $f(t)=t^2-1$ is the $\chi^2$-divergence between the distributions ${P}$ and ${Q}$ (a special case of the $f$-divergence).
Solving the problem~\eqref{kap} learns a (sparse) weight vector $\bm a$ of the candidate random features, so that the kernel matrix matches the target kernel $\bm y \bm y^{\!\top}$.
Problem~\eqref{kap} can be efficiently solved via bisection over a scalar dual variable, and an $\epsilon$-suboptimal solution can be found in $\mathcal{O}(J \log (1/ \epsilon) )$ time.

{\bf KP-RFF} (Kernel Polarization-RFF) \cite{shahrampour2018data}: It first generates a large number of random features by RFF and then selects a subset from them using an energy-based scheme
\begin{equation*}
	\tilde{S}(\bm \omega) = \frac{1}{n} \sum_{i=1}^{n} y_i z_p(\bm x_i, \bm \omega) \,.
\end{equation*}
Further, the quantity $(1/J) \sum_{i=1}^{J} \tilde{S}^2(\bm \omega_j)$ can be associated with kernel polarization for $\{ \bm w_i \}_{i=1}^{J}$ sampled from $p(\bm \omega)$.
Accordingly, the top $s$ random features with the top $|\tilde{S}(\cdot)|$ values are selected as the refined random features.
This algorithm can in fact be regarded as a version of the kernel alignment method for generating random features.

{\bf CLR-RFF} (Compression Low Rank-RFF) \cite{agrawal2019data}: It first generates a large number of random features and then selects a subset from them by approximately solving the optimization problem
\begin{equation}\label{rej}
	\begin{split}
		& \underset{\bm a \in \mathbb{R}^{J}: \| \bm a \|_0 \leq s }{\min}~ \frac{1}{n^{2}}\left\|\bm Z_J \bm Z_J^{\!\top}- \widetilde{\bm Z}_J(\bm a) \widetilde{\bm Z}_J(\bm a)^{\!\top} \right\|_{\mathrm{F}}^{2} =\\
		& \quad \qquad \mathbb{E}_{i,j \overset{\text{i.i.d.}}{\sim} [J] }\left[ \varphi_p(\bm x_i)^{\!\top}\varphi_p(\bm x_j) - \widetilde{\varphi}_p(\bm x_i)^{\!\top}\widetilde{\varphi}_p(\bm x_j) \right]\,,
	\end{split}
\end{equation}
where $\varphi_p(\bm x) \in \mathbb{R}^J$ uses $J$ random features, and $\widetilde{\varphi}_p(\bm x) $ is
\begin{equation*}
	\widetilde{\varphi}_p(\bm x) \! := \! \frac{1}{\sqrt{J}}
	\big[a_1 \exp(-\mbox{i}{\bm \omega}^{\!\top}_1 \bm x), \cdots,a_J\exp(-\mbox{i}{\bm \omega}^{\!\top}_J \bm x)\big]^{\!\top}\,,
\end{equation*}
which leads to $\widetilde{\bm Z}_J(\bm a) = [\widetilde{\varphi}_p(\bm x_1), \widetilde{\varphi}_p(\bm x_2), \cdots, \widetilde{\varphi}_p(\bm x_n) ] \in \mathbb{R}^{n \times J}$.
We can construct a Monte-Carlo estimate of the optimization objective function in Eq.~\eqref{rej} by sampling some pairs $i,j \overset{\text{i.i.d.}}{\sim} [J]$.
Therefore, this scheme focuses on a subset of pairs, instead of the all data pairs, by seeking a sparse weight vector $\bm a$ with only $s$ nonzero elements.
The problem of building a small, weighted subset of the data that approximates the full dataset, is known as the \emph{Hilbert coreset construction problem}.
It can be approximately solved by greedy iterative geodesic ascent \cite{campbell2018bayesian} or Frank-Wolfe based methods \cite{campbell2019automated}.
Another way to obtain the compact random features is using Johnson-Lindenstrauss random projection \cite{Hamid2014Compact} instead of the above data-dependent optimization scheme. 

\subsection{Kernel learning by random features}

This class of approaches construct random features using sophisticated learning techniques, e.g., by learning the spectral distribution of kernel from the data.

Representative approaches in this class often involve a \emph{one-stage} or \emph{two-stage} process.
The two-stage scheme is common when using random features. It first learns the random features, and then incorporates them into kernel methods for prediction.
Actually, the above-mentioned \emph{leverage sampling} and \emph{random features selection} based algorithms employ this scheme.
The algorithm proposed in \cite{li2019implicit} is a typical method for kernel learning by random features. This method first learns a spectral distribution of a kernel via an implicit generative model, and then trains a linear model by these learned features.

One-stage algorithms aim to simultaneously learn the spectral distribution of a kernel and the prediction model by solving a single joint optimization problem or using a spectral inference scheme. For example, 
Yu et al. \cite{Yu2015Compact} propose to jointly optimize the nonlinear feature mapping matrix $\bm W$ and the linear model with the hinge loss.
The associated optimization problem can be solved in an alternating fashion with SGD.
In \cite{bullins2017not}, the kernel alignment approach in the Fourier domain and SVM are combined into a unified framework, which can be also solved using an alternating scheme by Langevin dynamics and projection gradient descent.
Wilson and Adams \cite{Wilson2013Gaussian} construct stationary kernels as the Fourier transform of a Gaussian mixture based on Gaussian process frequency functions. This approach can be extended to learning with Fastfood \cite{yang2015carte}, non-stationary spectral kernel generalization \cite{remes2017non,ton2018spatial}, and the harmonizable mixture kernel \cite{shen2019harmonizable}.
Moreover, Oliva et al. \cite{oliva2016bayesian} propose a nonparametric Bayesian model, in which $p(\bm \omega)$ is modeled as a mixture of Gaussians with a Dirichlet process prior.
The parameters of the Gaussian mixture and the classifier/regressor model are inferred using MCMC.

\vspace{-0.0cm}
\section{Theoretical Analysis}
\label{sec:theoretical}
\begin{figure}[t]
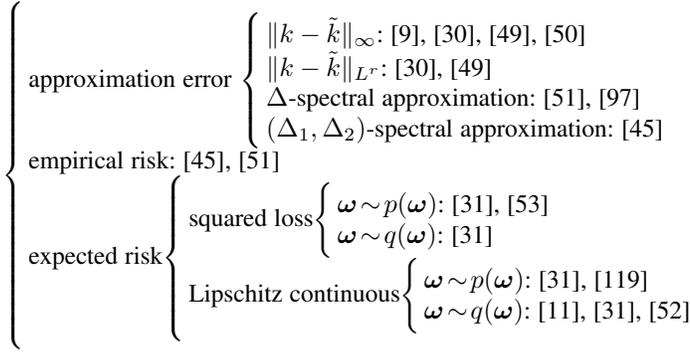

	\AB{}
	{
		\AB{approximation error}
		{
			$\| k - \tilde{k} \|_{\infty}$: \cite{rahimi2007random,sutherland2015error,sriperumbudur2015optimal,honorio2017error} \\
			$\| k - \tilde{k} \|_{L^r}$: \cite{sriperumbudur2015optimal,sutherland2015error} \\
			$\Delta$-spectral approximation: \cite{avron2017random,choromanski2018geometry} \\
			$(\Delta_1, \Delta_2)$-spectral approximation: \cite{zhang2019f}
		}\\
	empirical risk: \cite{avron2017random,zhang2019f} \\
		\AB{expected risk\!\!}
		{
			\AB{squared loss\!\!}
			{
				$\bm \omega \! \sim \! p(\bm \omega)$: \cite{Rudi2017Generalization,li2019towards} \\
				$\bm \omega \! \sim \! q(\bm \omega)$: \cite{li2019towards}
			} \\
			\AB{Lipschitz continuous\!\!}
			{
				$\bm \omega \! \sim \! p(\bm \omega)$: \cite{rahimi2008uniform,li2019towards} \\
				$\bm \omega \! \sim \! q(\bm \omega)$: \cite{bach2017equivalence,sun2018but,li2019towards}
			}
		}
	}
	\caption{Taxonomy of theoretical results on random features.}\label{figtheo}\vspace{-0.3cm}
\end{figure}

In this section, we review a range of theoretical results that center around the two questions mentioned in the introduction and restated below:
\begin{enumerate}
	\item \textbf{Approximation:} how many random features are needed to ensure a high quality estimator in kernel approximation?
	\item \textbf{Generalization:} how many random features are needed to incur no loss of empirical risk and expected risk in a learning estimator?
\end{enumerate}
Figure~\ref{figtheo} provides a taxonomy of representative work on these two questions.

For the approximation error, existing work  focuses on $\| k - \tilde{k} \|_{\infty}$ \cite{rahimi2007random,sutherland2015error,sriperumbudur2015optimal}, $\| k - \tilde{k} \|_{L^r}$ with $1 \leq r < \infty$ \cite{sriperumbudur2015optimal}, $\Delta$-spectral approximation \cite{avron2017random,choromanski2018geometry}, and $(\Delta_1, \Delta_2)$-spectral approximation \cite{zhang2019f}.
For the empirical risk under the fixed design setting, existing work provides guarantees on the expected in-sample predication error of the KRR estimator based on $\Delta$-spectral approximation bounds \cite{avron2017random} and $(\Delta_1, \Delta_2)$-spectral approximation bounds \cite{zhang2019f}.
For the expected risk, a series of  works investigate the generalization properties of methods based on \emph{$p(\bm \omega)$-sampling} or \emph{$q(\bm \omega)$-sampling}. These results cover loss functions with/without  Lipschitz continuity and apply to e.g. KRR \cite{Rudi2017Generalization,li2019towards} and SVM \cite{rahimi2009weighted,bach2017equivalence,sun2018but} under different assumptions.

More specifically, Rahimi and Recht \cite{rahimi2009weighted} provide the earliest result on learning with RFF with Lipschitz continuous loss functions. Their results imply that  $\Omega(n)$ random features are sufficient to incur no loss of learning accuracy.
This result is improved in \cite{li2019towards}, which shows that  $\Omega(\sqrt{n} \log n)$ random features or even less suffice for the Gaussian kernel.
When using the data-dependent sampling $\{ {\bm \omega}_i \}_{i=1}^s \sim q({\bm \omega})$, the above results are further improved in \cite{bach2017equivalence,sun2018but,li2019towards} under various settings.
Note that some results above do not directly apply to the squared loss in KRR, whose Lipschitz parameter is unbounded. For squared losses, Rudi et al. \cite{Rudi2017Generalization} show that $\Omega(\sqrt{n}\log n)$ random features by RFF suffice to achieve a minimax optimal learning rate $\mathcal{O}(1/\sqrt{n})$.
A more refined analysis is given in \cite{li2019towards} under the $p(\bm \omega)$-sampling and $q(\bm \omega)$-sampling settings.

Below we discuss the above theoretical work in more details. 

\subsection{Approximation error}

Table~\ref{tabapp} summarizes representative theoretical results on the convergence rates, the upper bound of the growing diameter, and the resulting sample complexity under different metrics.
Here sample complexity means the number of random features sufficient for achieving a maximum approximation error at most $\epsilon$.

The first result of this kind is given by Rahimi and Recht \cite{rahimi2007random}, who use a covering number argument to derive a uniform convergence guarantee as follows. 
For a compact subset $\mathcal{S}$ of $\mathbb{R}^d$, let $|\mathcal{S}| := \sup_{\bm x, \bm x' \in \mathcal{S}}\| \bm x - \bm x' \|_2$ be its diameter and consider the $L^\infty$ error $\| k - \tilde{k} \|_{\infty} := \sup_{\bm x, \bm x' \in \mathcal{S}} | k(\bm x, \bm x') - \tilde{k}(\bm x, \bm x') |$.
\begin{theorem}\label{theoapp}
[Uniform convergence of RFF \cite{rahimi2007random,sutherland2015error}]\label{approtheo2}
Let $\mathcal{S}$ be a compact subset of $\mathbb{R}^d$ with diameter $|\mathcal{S}|$. Then, for a stationary kernel $k$ and its approximated kernel $\tilde{k}$ obtained by RFF, we have
\begin{equation*}
\operatorname{Pr}\left[\| k - \tilde{k} \|_{\infty} \geq \epsilon\right]  \leq C_d \left(\frac{\varsigma_{p} |\mathcal{S}|}{\epsilon}\right)^{\frac{2d}{d+2}} \exp \left(-\frac{s\epsilon^2}{4(d+2)} \right) \, ,
\end{equation*}
where $\varsigma_p^2 = \mathbb{E}_p[{\bm \omega}^{\!\top} {\bm \omega}] = \operatorname{tr} \nabla^2 k(0) \in \mathcal{O}(d)$, and $C_d := 2^{\frac{6 d+2}{d+2}}\left(\left(\frac{2}{d}\right)^{\frac{d}{d+2}} +\left(\frac{d}{2}\right)^{\frac{2}{d+2}}\right)$ satisfies $C_d \leq 256$ in \cite{rahimi2007random} and is further improved to $C_d \leq 66$ in \cite{sutherland2015error} by optimization balls of radius in covering number.
\end{theorem}
According to the above theorem by covering number, with $s:=\Omega(\epsilon^{-2} d \log ({1}/{\epsilon \delta}))$ random features, one can ensure an $\epsilon$ uniform approximation error with probability greater than $1-\delta$.
This result also applies to dot-product kernels by random Maclaurin feature maps (see \cite[Theorem~8]{kar2012random}).
The quadrature based algorithm \cite{munkhoeva2018quadrature} follows this proof framework, and achieves the same error bound with a smaller constant than RFF in Theorem~\ref{theoapp} by an extra boundedness assumption.
Instead, Fastfood \cite{le2013fastfood} on Gaussian kernels achieves $\mathcal{O}(\sqrt{\log(d/\delta)})$ times approximation error than RFF due to estimates for $\bm \Gamma \bm H \bm B_2$ in Eq.~\eqref{wfast}, which is based on concentration inequalities for Lipschitz continuous functions under the Gaussian distribution.

Different from the above results using Hoeffding's inequality for the covering number bound in their proof, Sriperumbudur and Szab\'{o} \cite{sriperumbudur2015optimal} revisit the above bound by refined technique of McDiarmid's inequality, symmetrization and bound the expectation of Rademacher average by Dudley entropy bound.
\begin{theorem}[Theorem 1 in \cite{sriperumbudur2015optimal}]
	\label{approtheo3}
	Under the same assumption of Theorem~\ref{approtheo2}, we have
	\begin{equation*}
		\operatorname{Pr}\left[\| k - \tilde{k} \|_{\infty} \geq \frac{h(d, |\mathcal{S}|, \sigma_p)+ \sqrt{2 \epsilon}}{\sqrt{s}} \right]  \leq e^{-\epsilon}\,,
	\end{equation*}
	where $h(d, |\mathcal{S}|, \sigma_p)$ is an appropriately defined function of $d$, $|\mathcal{S}|$, and $\sigma_p$.
	For better comparison, the above inequality can be rewritten as \cite{honorio2017error}
	\begin{equation*}
		\begin{split}
			\operatorname{Pr}\left[\| k - \tilde{k} \|_{\infty} \geq \epsilon\right]  & \leq [(\sigma_p+1)(2|\mathcal{S}|+1)]^{1024d}  \\&
			\exp \left( - \frac{s \epsilon^2}{2} + \frac{256d}{\log(2|\mathcal{S}|+1)} \right).
		\end{split}
	\end{equation*}
\end{theorem}
Theorem~\ref{approtheo3} shows that $\tilde{k}$ is a consistent estimator of $k$ in the topology of compact convergence as $s \rightarrow \infty$ with the convergence rate $\mathcal{O}_p(\sqrt{s^{-1} \log |\mathcal{S}| })$.
Consequently, $\mathcal{O}(\epsilon^{-2}\log |\mathcal{S}|)$ random features suffice to achieve an $\epsilon$ approximation accuracy. This sample complexity bound scales logarithmically with $|\mathcal{S}|$, which improves upon the $\mathcal{O}(\epsilon^{-2} |\mathcal{S}|^2 \log (|\mathcal{S}|/\epsilon))$ bound that follows from \cite{rahimi2007random,sutherland2015error} (cf.\ Theorem~\ref{approtheo2}).
Apart from the $L^\infty$ error bound, the authors of \cite{sriperumbudur2015optimal} further derive bounds on the $L^r$ error $\| k - \tilde{k} \|_{L^r} := \left( \int_{\mathcal{S}} \int_{\mathcal{S}} | k(\bm x, \bm x') - \tilde{k}(\bm x,  \bm x') |^r \mathrm{d} \bm x \mathrm{d} \bm x' \right)^{1/r}$ for  $1 \leq r < \infty$; see Table~\ref{tabapp} for a summary.
We remark that the ${L^2_{\rho_{\mathcal{X}}}}$ error bound is also given in \cite{sutherland2015error}, though the rate in \cite{sriperumbudur2015optimal} is sometimes better in terms of the diameter.

For the Gaussian kernel, the approximation guarantee  can be further improved. In particular, the following theorem gives a probability bound independent of $d$.
\begin{theorem}[Theorem~1 in \cite{honorio2017error}]
	\label{sharpuniform}
	Under the same assumption of Theorem~\ref{approtheo2}, for the Gaussian kernel $k$ and its approximation $\tilde{k}$ by RFF, we have
	\begin{equation*}
		\operatorname{Pr}\left[\| k - \tilde{k} \|_{\infty} \geq \epsilon\right]  \leq \frac{3}{s^{1/3}} \left(\frac{|\mathcal{S}|}{\epsilon} \right)^{2/3} \exp \left(-\frac{s \epsilon^2}{12} \right) \,.
	\end{equation*}
\end{theorem}


\begin{table*}[!htb]
        \centering
        \fontsize{7}{8}\selectfont
        \begin{threeparttable}
                \caption{Comparison of convergence rates and required random features for kernel approximation error.}
                \label{tabapp}
                \begin{tabular}{cccccccccccccccccccc}
                        \toprule
                        Metric &Results &Convergence rate & Upper bound of $|\mathcal{S}|$ & Required random features $s$   \cr
                        \midrule
                        \multirow{6}{1.2cm}{$\| k - \tilde{k} \|_{\infty}$} &Theorem~\ref{approtheo2} (\cite{rahimi2007random,sutherland2015error}) &  $\mathcal{O}_{p}\left( |\mathcal{S}| \sqrt{\frac{\log s}{s}}\right) $ & $|\mathcal{S}| \leq \Omega \left(\sqrt{\frac{s}{\log s}}\right)$ & $s \geq \Omega \left({d}\epsilon^{-2}\log \frac{|\mathcal{S}|}{\epsilon} \right)$   \cr
                        \cmidrule(lr){2-5}
                        &Theorem~1 in \cite{sriperumbudur2015optimal} & $\mathcal{O}_{p}\left( \sqrt{\frac{\log |\mathcal{S}|}{s}}\right) $  & $|\mathcal{S}| \leq \Omega (s^{c})$\tnote{1}  & $s \geq \Omega \left({d}\epsilon^{-2}\log {|\mathcal{S}|} \right)$ \cr
                        \cmidrule(lr){2-5}
                        &Theorem~1 in \cite{honorio2017error} (Gaussian kernels) & $\mathcal{O}_{p}\left( \sqrt{\frac{\log |\mathcal{S}|}{s}}\right) $  & $|\mathcal{S}| \leq \Omega (s^{c})$  &  $s \geq \Omega \left(\epsilon^{-2}\log {|\mathcal{S}|} \right)$ \cr
                        \midrule
                        $\| k - \tilde{k} \|_{L^r}~(1 \leq r < \infty)$   & Corollary 2 in \cite{sriperumbudur2015optimal} & $\mathcal{O}_{p}\left( |\mathcal{S}|^{\frac{2d}{r}} \sqrt{\frac{\log |\mathcal{S}|}{s}}\right) $  & $|\mathcal{S}| \leq \Omega \left((\frac{s}{\log s})^{\frac{r}{4d}} \right) $  & $s \geq \Omega \left({d}\epsilon^{-2}\log {|\mathcal{S}|} \right)$ \cr
                        \midrule
                        $\| k - \tilde{k} \|_{L^r}~(2 \leq r<\infty)$   &Theorem 3 in \cite{sriperumbudur2015optimal} & $\mathcal{O}_{p}\left( |\mathcal{S}|^{\frac{2d}{r}} \sqrt{\frac{1}{s}}\right) $  & $|\mathcal{S}| \leq \Omega \left(s^{\frac{r}{4d}}\right) $  & $s \geq \Omega \left({d}\epsilon^{-2}\log {|\mathcal{S}|} \right)$ \cr
                        \midrule
                        \multirow{4}{3cm}{$\Delta$-spectral approximation} &Theorem 7 in \cite{avron2017random} & $\mathcal{O}_{p}\left( \sqrt{\frac{n_{\lambda}}{s}}\right) $  & -  & $s \geq \Omega(n_{\lambda} \log d_{\bm{K}}^{\lambda})$  \cr
                        \cmidrule(lr){2-5}
                        &Theorem 5.4 in \cite{choromanski2018geometry} (Gaussian kernels) & $\mathcal{O}_{\text{RFF}/\text{ORF}}\left( \frac{1}{s \lambda^2} \right) $  & -  & $s \geq \Omega(n^{2 \alpha})$ \cr
                        \cmidrule(lr){2-5}
                        &Lemma~6 in \cite{avron2017random} & $\mathcal{O}_{q}\left( \sqrt{\frac{d_{\bm{K}}^{\lambda}}{s}}\right) $  & -  & $s \geq \Omega(d_{\bm{K}}^{\lambda} \log d_{\bm{K}}^{\lambda})$ \cr
                        \midrule
                        $(\Delta_1, \Delta_2)$-spectral approximation   &Theorem 2 in \cite{zhang2019f} & $\mathcal{O}_{\operatorname{LP}}\left( \sqrt{\frac{n_{\lambda}}{s}}\right) \tnote{2}$  & -  &$s \geq \Omega(n_{\lambda} \log d_{\bm{K}}^{\lambda})$  \cr
                        \bottomrule
                \end{tabular}
                \begin{tablenotes}
                        \footnotesize
                        \item[1] $c$ is some constant satisfying $0 < c < 1$.\\
                        \item[2] $\operatorname{LP}$ denotes that $\{ {\bm \omega}_i \}_{i=1}^s$ are obtained by RFF and then are quantized to a Low-Precision $b$-bit representation; see \cite{zhang2019f}.
                \end{tablenotes}
        \end{threeparttable}
\end{table*}

Avron et al. \cite{avron2017random} argue that the above point-wise distances $\| k - \tilde{k} \|_\infty$ or $\| k - \tilde{k}\|_{L^r}$ are not sufficient to accurately measure the approximation quality. Instead,  they focus on the following spectral approximation criterion.
\vspace{-0.1cm}
\begin{definition}\label{defdeltaspe}
[$\Delta$-spectral approximation \cite{avron2017random}]
  For $0 \leq \Delta < 1$, a symmetric matrix $\bm A$ is a $\Delta$-spectral approximation of another symmetric matrix $\bm B$, if $(1-\Delta)\bm B \preceq \bm A \preceq (1+\Delta)\bm B$, where $A \preceq B$ indicates that $B-A$ is a semi-positive definite matrix.
\end{definition}
According to this definition, $\bm Z \bm Z^{\!\top} + \lambda \bm I_n$ is $\Delta$-spectral approximation of $\bm K + \lambda \bm I_n$ if 
\begin{equation*}
	(1-\Delta)\left(\bm{K}+\lambda \bm{I}_{n}\right) \preceq \bm{Z} \bm{Z}^{\!\top}+\lambda \bm{I}_{n} \preceq(1+\Delta)\left(\bm{K}+\lambda \bm{I}_{n}\right)\,.
\end{equation*}

The follow theorem gives the number of random features $s$ that are sufficient to guarantee $\Delta$-spectral approximation.
\begin{theorem}[Theorem 7 in \cite{avron2017random}]
	\label{deltap}
	Let $k$ be a shift-invariant kernel and its associated probability distribution $p({\bm \omega})$ (i.e., the Fourier transform of $k$),  $\Delta \leq 1/2$, $\delta \in (0,1)$, and $n_{\lambda}:= n/\lambda$. Assume that $\| \bm K \|_2 \geq \lambda$ and $\{ {\bm \omega}_i \}_{i=1}^s \sim p(\bm \omega)$.
	If the total number of random features satisfies
	\begin{equation*}
		s \geq \frac{8}{3} \Delta^{-2} n_{\lambda} \log \left(16 d_{\bm{K}}^{\lambda} / \delta \right) \,,
	\end{equation*}
	then 
	\begin{equation*}
		\begin{split}
			& \operatorname{Pr}\left[\!(1\!-\!\Delta)\left(\bm{K}\!+\!\lambda \bm{I}_{n}\right) \preceq \bm{Z} \bm{Z}^{\!\top}\!+\!\lambda \bm{I}_{n} \preceq(1+\Delta)\left(\bm{K}+\lambda \bm{I}_{n}\right) \!\right] \\
			& \geq 1- 16 d_{\bm{K}}^{\lambda} \exp \left( \frac{-3s \Delta^2}{8n_{\lambda}} \right) \geq 1 - \delta\,.
		\end{split}
	\end{equation*}
\end{theorem}
Theorem~\ref{deltap} states that $\Omega( n_{\lambda} \log  d_{\bm{K}}^{\lambda})$ random features are sufficient to guarantee $\Delta$-spectral approximation by the matrix Bernstein concentration inequality and effective degree of freedom, where $n_{\lambda}:= n/\lambda$.
Under this framework, Choromanski et al. \cite[Theorem 5.4]{choromanski2018geometry} present a non-asymptotic comparison result between RFF and ORF for spectral approximation by virtue of the smallest singular value of $\bm K + n\lambda \bm I$. 
\begin{theorem}
	[Theorem 5.4 in \cite{choromanski2018geometry}]
	\label{BORFthem}
	For the Gaussian kernel, let $\widetilde{\Delta}$ be the smallest positive number such that $\widetilde{\bm K} + \lambda n \bm I_n $ is a $\widetilde{\Delta}$-spectral approximation of $\bm K + \lambda n \bm I_n$, where $\widetilde{\bm K}$ is an approximate kernel matrix obtained by RFF or ORF. Then, for any $\epsilon > 0$ we have
	\begin{equation*}
		\mathrm{Pr}[\widetilde{\Delta} > \epsilon] \leq \frac{B}{\epsilon^2 \sigma_{\min}^2}\,,
	\end{equation*}
	where $B:=\mathbb{E}[\| \widetilde{\bm K} - \bm K \|_{\F}^2]$ and $\sigma_{\min}^2$ is the smallest singular value of  $\bm K + \lambda n \bm I_n$. In particular, letting $B^{\textup{ORF}}$ denotes the value of $B$ for the estimator ORF and $B^{\textup{RFF}}$  for RFF, we have
	\begin{equation*}
		B^{\textup{RFF}} - B^{\textup{ORF}} \!=\! \frac{s-1}{s}\! \left(\! \frac{1}{2d} \sum_{i,j=1}^{n}\! \frac{\| \bm x_i \!-\! \bm x_j \|_2^4}{\varsigma^2} e^{-\frac{\| \bm x_i - \bm x_j\|_2^2}{\varsigma^2}} \!+\! \mathcal{O}\Big(\frac{1}{d}\Big) \!\!\right) \!.
	\end{equation*}
\end{theorem}
Theorem~\ref{BORFthem} shows that $B^{\text{RFF}} > B^{\text{ORF}}$ always holds  for the Gaussian kernel.
To better understand the above upper bound on $\mathrm{Pr}[\widetilde{\Delta} > \epsilon]$, we note that both $\mathbb{V}[\text{RFF}]$ and  $\mathbb{V}[\text{ORF}]$ are $\mathcal{O}(1/s)$, hence $B = \mathcal{O}(n^2/s)$.
Moreover, since the Gaussian kernel has exponentially decaying eigenvalues (see Assumption~\ref{eigenassum}), we have $\sigma_{\min}^2 = \Omega(n^2 \lambda^2)$.
Therefore, the upper bound of  $\mathrm{Pr}[\widetilde{\Delta} > \epsilon]$ is on the order of $\mathcal{O}(\frac{1}{s \lambda^2})$.
With the standard scaling  of the regularization parameter $\lambda = n^{-\alpha}$, $\alpha \in (0,1]$, we need $s:=\Omega(n^{2 \alpha})$ to get a non-trivial upper bound on the probability.
When $\alpha = 1/2$, these results for RFF and ORF require $\Omega(n)$ random Fourier features, which is somewhat unsatisfactory~\cite{li2019towards}. 

The results in Theorem~\ref{deltap} can be improved if we consider data-dependent sampling, i.e., $\{ {\bm \omega}_i \}_{i=1}^s$ are sampled from the empirical ridge leverage score distribution $q({\bm \omega}) = {l_{\lambda}({\bm \omega})}/{d^{\lambda}_{\bm{K}}}$ in Eq.~\eqref{qwdef} instead of the standard $p({\bm \omega})$.
\begin{theorem}[Lemma~6 in \cite{avron2017random}]
	\label{deltaq}
	Let $k$ be a shift-invariant kernel associated with the empirical ridge leverage score distribution $q(\bm \omega)$ in Eq.~\eqref{qwdef},  $\Delta \leq 1/2$ and $\delta \in (0,1)$. Assume that $\| \bm K \|_2 \geq \lambda$ and $\{ {\bm \omega}_i \}_{i=1}^s \sim q(\bm \omega)$.
	If the total number of random features satisfies
	\begin{equation*}
		s \geq \frac{8}{3} \Delta^{-2} d_{\bm{K}}^{\lambda} \log \left(16 d_{\bm{K}}^{\lambda} / \delta \right) \,,
	\end{equation*}
	then 
	\begin{equation*}
		\begin{split}
			& \operatorname{Pr}\left[(1-\Delta)\left(\bm{K}+\lambda \bm{I}_{n}\right) \preceq \bm{Z} \bm{Z}^{\!\top}+\lambda \bm{I}_{n} \preceq(1+\Delta)\left(\bm{K}+\lambda \bm{I}_{n}\right) \right] \\
			& \geq 1 - 16 d_{\bm{K}}^{\lambda} \exp \left( \frac{-3s \Delta^2}{8d_{\bm{K}}^{\lambda}} \right) \geq 1 - \delta\,.
		\end{split}
	\end{equation*}
\end{theorem}
Theorem~\ref{deltaq} shows that if we sample using the ridge leverage function, then $\Omega( d_{\bm{K}}^{\lambda} \log  d_{\bm{K}}^{\lambda})$ random features, which is less than $\Omega( n_{\lambda} \log  d_{\bm{K}}^{\lambda})$, suffice for spectral approximation of $\bm K$.

The authors of \cite{zhang2019f} generalize the notion of $\Delta$-spectral approximation to ($\Delta_1, \Delta_2)$-spectral approximation. 
\begin{definition}
	[($\Delta_1, \Delta_2)$-spectral approximation \cite{zhang2019f}]
	For $\Delta_1$,$\Delta_2 \geq 0$, a symmetric matrix $\bm A$ is a $(\Delta_1, \Delta_2)$-spectral approximation of another symmetric matrix $\bm B$, if $(1-\Delta_1)\bm B \preceq \bm A \preceq (1+\Delta_2)\bm B$.
\end{definition}
This definition is motivation by the argument that the quantities $\Delta_1$ and $\Delta_2$ in the upper and lower bounds may have different impact on the generalization performance. Using this definition, Zhang et al. \cite{zhang2019f} derive the following approximation guarantees when one  quantizes each random Fourier feature $\bm \omega_i$ to a low-precision $b$-bit representation, which allows more features to be stored in the same amount of space. 
\begin{theorem}
	[Theorem 2 in \cite{zhang2019f}]
	\label{deltalp}
	
	Let $\widetilde{\bm K}$ be an $s$-features $b$-bit LP-RFF approximation of a kernel matrix $\bm K$ and $\delta \in (0,1)$. Assume that $\| \bm K \|_2 \geq \lambda \geq \delta^{2}_b = 2/(2^b - 1)^2$ and define $a:=8 \operatorname{Tr}(\bm K + \lambda \bm I_n)^{-1} (\bm K + \delta^2_b \bm I_n)$. For $\Delta_1 \leq 3/2$ and $\Delta_2 \in [\frac{\delta_b^2}{\lambda}, \frac{3}{2}]$, 
	if the total number of random features satisfies
	\begin{equation*}
		s \geq \frac{8}{3} n_{\lambda} \max\left\{ \frac{2}{\Delta_1}, \frac{2}{\Delta_2 - \delta_b^2 / \lambda} \right\} \log \left( \frac{a}{\delta} \right) \,,
	\end{equation*}
	then 
	\begin{equation*}
		\begin{split}
			& \operatorname{Pr}\left[(1\!-\!\Delta_1)\left(\bm{K}\!+\!\lambda \bm{I}_{n}\right) \!\preceq\! \tilde{\bm K}\!+\!\lambda \bm{I}_{n} \preceq(1+\Delta_2)\left(\bm{K}\!+\!\lambda \bm{I}_{n}\right) \right]  \\
			& \geq  1- a \left[ \exp \left(\! \frac{-3s \Delta_1^2}{4n_{\lambda}(1\!+\!2/3\Delta_1)} \!\right) \right.\\
			& \quad \left. +  \exp \! \left(\! \frac{-3s(\Delta_2 \!-\! \delta_b^2 / \lambda)^2}{4n_{\lambda}(1+2/3(\Delta_2 \!-\! \delta_b^2 / \lambda))} \!\right) \right].
		\end{split}
	\end{equation*}
\end{theorem}
Theorem~\ref{deltalp} shows that when the quantization noise is small relative to the
regularization parameter, using low precision has minimal impact on the number of features required for the ($\Delta_1, \Delta_2)$-spectral approximation.
In particular, as $s \rightarrow \infty$, $\Delta_1$ converges to zero for any precision $b$, whereas $\Delta_2$ converges to a value upper bounded by $\delta_b^2/\lambda$.
If $\delta_b^2/\lambda \ll \Delta_2$, using $b$-bit precision has negligible effect on the number of
features required to attain this $\Delta_2$ see Table~\ref{tabapp} for a summary.

\vspace{-0.2cm}
\subsection{Risk and generalization property}

The above results on approximation error are  a means to an end.
More directly related to the learning performance is understanding generalization properties of random features based algorithms.
To this end, a series of work study the generalization properties of algorithms based on $p(\bm \omega)$-sampling and $q(\bm \omega)$-sampling. Under different assumptions, theoretical results have been obtained for loss functions with/without Lipschitz continuity and for learning tasks including KRR \cite{Rudi2017Generalization,li2019towards} and SVM \cite{rahimi2009weighted,bach2017equivalence,sun2018but}.
Apart from supervised learning with random features, results on randomized nonlinear component analysis refer to \cite{lopez2014randomized}, random features with matrix sketching \cite{ghashami2016streaming}, doubly stochastic gradients scheme \cite{xie2015scale}, statistical consistency \cite{sriperumbudur2017statistical,ullah2018streaming}.

\begin{figure*}[htb]
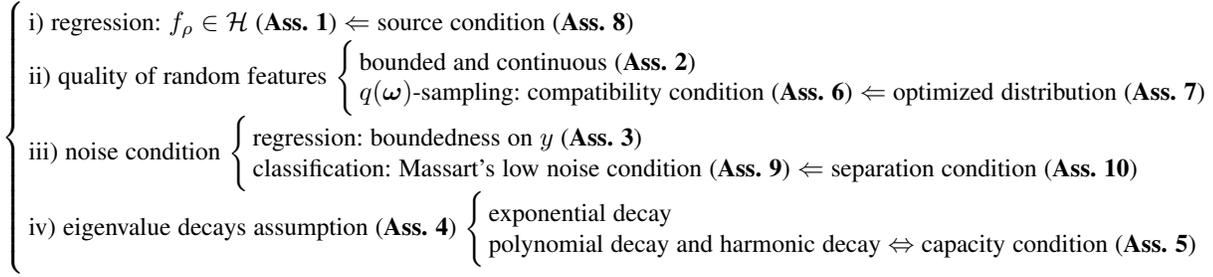

        \centering
        \AB{}
        {
                i) regression: $f_{\rho} \in \mathcal{H}$ ({\bf Ass.~\ref{existass}}) $\Leftarrow$ source condition ({\bf Ass.~\ref{regularity}})\\
                \AB{ii) quality of random features}
                {
                        bounded and continuous ({\bf Ass.~\ref{rfbc}}) \\
                        $q(\bm \omega)$-sampling: compatibility condition ({\bf Ass.~\ref{optimalrff}}) $\Leftarrow$ optimized distribution ({\bf Ass.~\ref{optdist}})
                }\\
                \AB{iii) noise condition}
                {
                        regression: boundedness on $y$ ({\bf Ass.~\ref{outputs}}) \\
                        classification: Massart’s low noise condition ({\bf Ass.~\ref{noiseass}}) $\Leftarrow$ separation condition ({\bf Ass.~\ref{sepacon}})
                } \\
                \AB{iv) eigenvalue decays assumption ({\bf Ass.~\ref{eigenassum}})}
                {
                        exponential decay \\
                        polynomial decay and harmonic decay $\Leftrightarrow$ capacity condition ({\bf Ass.~\ref{effectass}})
                }
        }
        \caption{Relationship between the needed assumptions. The notation $A \Leftarrow B$ means that B is a stronger assumption than A.}\label{figassum}
        \vspace{-0.3cm}
\end{figure*}

\vspace{-0.1cm}
\subsubsection{Assumptions}
Before we detail these theoretical results, we summarize the standard assumptions imposed in existing work.
Some assumptions are technical, and thus familiarity with statistical learning theory (see Section~\ref{sec:setting}) would be helpful.
We organize these assumptions in four categories as shown in Figure~\ref{figassum}, including i) the existence of $f_{\rho}$ (Assumption~\ref{existass}) and its stronger version (Assumption~\ref{regularity}); ii) quality of random features (Assumptions~\ref{rfbc},~\ref{optimalrff},~\ref{optdist}); iii) noise conditions (Assumptions~\ref{outputs},~\ref{noiseass},~\ref{sepacon}); iv) eigenvalue decay (Assumptions~\ref{eigenassum},~\ref{effectass}).

We first state three basic assumptions, which are needed in all of the (regression) results to be presented.
\begin{assumption}[Existence \cite{cucker2007learning,Rudi2017Generalization}]
\label{existass}
In regression task, we assume $f_{\rho} \in \mathcal{H}$.
\end{assumption}
Note that since we consider a potentially infinite dimensional RKHS $\mathcal{H}$, possibly universal \cite{Steinwart2008SVM}, the existence of the target function $f_{\rho}$ is not automatic.
However, if we restrict to a bounded subspace of $\mathcal{H}$, i.e., $\mathcal{H}_{R}=\{f \in \mathcal{H} :\|f\| \leq R\}$ with $R<\infty$ fixed a prior, then  a minimizer of the risk $\mathcal{E}(f)$ always exists as long as  $\mathcal{H}_R$ is not universal. If $f_{\rho}$ exists, then it must lie in a ball of some radius $R_{\rho, \mathcal{H}}$.
The results in this section do not require prior knowledge of $R_{\rho, \mathcal{H}}$ and they hold for any finite radius.

\begin{assumption}
[Random features are bounded and continuous \cite{Rudi2017Generalization}]
\label{rfbc}
For the shift-invariant kernel $k$, we assume that $\varphi(\bm \omega^{\!\top} \bm x)$ in Eq.~\eqref{kernelfor} is continuous in both variables and bounded, i.e., there exists $\kappa \geq 1$ such that $|\varphi(\bm \omega^{\!\top} \bm x)|<\kappa$ for all $\bm x \in \mathcal{X}$ and ${\bm \omega} \in \mathbb{R}^d$.
\end{assumption}

\begin{assumption}
[Bernstein's condition \cite{Steinwart2008SVM,blanchard2010optimal}]
\label{outputs}
  For any $\bm x \in \mathcal{X}$, we assume
$
\mathbb{E}\left[|y|^{b} \mid \bm x\right] \leq \frac{1}{2}b ! \varsigma^{2} B^{b-2} $ when $b \geq 2$\,. 
\end{assumption}
This noise condition is weaker than the boundedness on $y$.
It is satisfied when $y$ is bounded, sub-Gaussian, or sub-exponential.
In particular, if $y \in [-\frac{b}{2}, \frac{b}{2}]$ almost surely with $b > 0$, then Assumption~\ref{outputs} is satisfied with $\varsigma = B = b$.

The above three assumptions are needed in all theoretical results for regression presented in this section, so we omit them when stating these results.
We next introduce several additional assumptions, which are needed in some of the theoretical results.

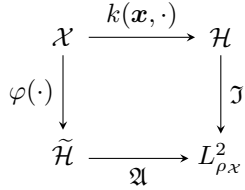
\begin{figure}[t]
	\centering
	\begin{tikzpicture}
		\matrix (m) [matrix of math nodes,row sep=3em,column sep=4em,minimum width=2em]
		{
			\mathcal{X} & \mathcal{H} \\
			\widetilde{\mathcal{H}} & L^2_{\rho_{\mathcal{X}}} \\
		};
		\path[-stealth]
		(m-1-1) edge node [left] {$\varphi(\cdot)$} (m-2-1)
		edge node [above] {$k(\bm x, \cdot)$} (m-1-2)
		(m-2-1.east|-m-2-2) edge node [below] {$\mathfrak{A}$}
		node [above] {} (m-2-2)
		(m-1-2) edge node [right] {$\mathfrak{I}$} (m-2-2);
	\end{tikzpicture}
	\caption{Maps between various spaces.}
	\label{mapspace}
\end{figure}

\underline{\bf Eigenvalue Decay Assumptions:} The following assumption, which characterizes the ``size” of the RKHS $\mathcal{H}$ of interest, is often discussed in learning theory.
\begin{assumption}
[Eigenvalue decays \cite{bach2013sharp}]
\label{eigenassum}
        A kernel matrix $\bm K$ admit the following three types of eigenvalue decays: 1) \emph{Geometric/exponential decay}: $\lambda_i(\bm K) \propto n\exp({-i^{1/c}})$, which leads to $d^{\lambda}_{\bm K} \lesssim \log (R_0/\lambda)$; 2) \emph{Polynomial decay}: $\lambda_i(\bm K) \propto ni^{-2a}$, which implies $d^{\lambda}_{\bm K} \lesssim (1/\lambda)^{1/2a}$; 3) \emph{Harmonic decay}: $\lambda_i(\bm K) \propto n/i$, which results in $d^{\lambda}_{\bm K} \lesssim (1/\lambda)$.
\end{assumption}
We give some remarks on the above assumption. 
For shift-invariant kernels, if the RKHS is small, the eigenvalues of the kernel matrix $\bm K$ often admit a fast decay.
Consequently, functions in the RKHS are smooth enough that a good prediction performance can be achieved.
On the other hand, if the RKHS is large and the eigenvalues decay slowly, then functions in the RKHS are not smooth, which would lead to a sub-optimal error rate for prediction.
It can be linked to the integral operator \cite{cucker2007learning,Steinwart2008SVM} characterizing the hypothesis space, defined as $\Sigma: L^2_{\rho_{\mathcal{X}}} \rightarrow L^2_{\rho_{\mathcal{X}}}$ such that
\begin{equation*}
(\Sigma g)(\bm x)=\int_{\mathcal{X}} k(\bm x,\bm x') g(\bm x') \mathrm{d} \rho_{\mathcal{X}}(\bm x'), \quad \forall g \in L^2_{\rho_{\mathcal{X}}} \,.
\end{equation*}
It is clear that the operator $\Sigma$ is self-adjoint, positive definite, and trace-class when $k(\cdot,\cdot)$ is continuous.
This operator can be represented as $\Sigma = \mathfrak{I} \mathfrak{I}^*$ in terms of the inclusion operator $\mathfrak{I}: \mathcal{H} \rightarrow L^2_{\rho_{\mathcal{X}}},~(\mathfrak{I}f) = f$.
Here $\mathfrak{I}^*$ is the adjoint of $\mathfrak{I}$ and is  given by 
\begin{equation*}
	\mathfrak{I}^*: L^2_{\rho_{\mathcal{X}}} \rightarrow \mathcal{H}, ~(\mathfrak{I}^* f) (\cdot) = \int_{\mathcal{X}} k(\bm x, \cdot) f(\bm x) \mathrm{d} \rho_{\mathcal{X}}\,,
\end{equation*}
due to the self-adjoint property of the Hilbert spaces $L^2_{\rho_{\mathcal{X}}}$ and $\mathcal{H}$  \cite{ullah2018streaming}.
With $s$ random features, the inclusion operator $\mathfrak{I}$ can be approximated by the operator $\mathfrak{A}: \widetilde{\mathcal{H}} \rightarrow L^2_{\rho_{\mathcal{X}}},~(\mathfrak{A}\bm \beta) =\langle \varphi(\cdot), \bm \beta \rangle_{\widetilde{\mathcal{H}}},~\forall \beta \in \mathbb{R}^s$.
Figure~\ref{mapspace} presents the relationship between various spaces under different operators.

The integral operator $\Sigma$ plays a significant role in characterizing the hypothesis space.
In particular, the decay rate of the spectrum of $\Sigma$ quantifies the capacity of the hypothesis space in which we search for the solution. This capacity in turn determines the number of random features required for accurate learning.
Rudi and Rosasco \cite{Rudi2017Generalization} consider the following assumption on $\Sigma$.
\begin{assumption}
[Capacity condition \cite{cucker2007learning,caponnetto2007optimal}]
\label{effectass}
       There exist $Q > 0$ and $\gamma \in [0,1]$ such that for any $\lambda > 0$, we have 
        \begin{equation}\label{nlambda}
        \begin{split}
        \mathcal{N}(\lambda) &:= \operatorname{tr}\left((\Sigma+\lambda I)^{-1} \Sigma \right) \leq Q^2 \lambda^{-\gamma}\,.
        \end{split}
        \end{equation}
\end{assumption}
The effective dimension $\mathcal{N}(\lambda)$ \cite{zhang2005learning} measures the ``size" of the RKHS, and is in fact the operator form of $d^{\lambda}_{\bm K} $ in Eq.~\eqref{dklambda}.
Assumption~\ref{effectass} holds if the eigenvalues $\lambda_i$ of $\Sigma$ decay as $i^{-1/\gamma}$, which corresponds to the eigenvalue decay of $\bm K$ in Assumption~\ref{eigenassum} with $\gamma := 1/(2a)$ \cite{shawe2002eigenspectrum}.
The case $\gamma = 0$ is the more benign situation, whereas $\gamma=1$ is the worst case.

\underline{\bf Quality of Random Features:} Here we introduce several technical assumptions on the quality of random features.
The leverage score in Eq.~\eqref{llambda} admits the operator form
\begin{equation*}
\mathcal{F}_{\infty}(\lambda):= \sup _{\bm \omega}\left\|(\Sigma+\lambda I)^{-1 / 2} \varphi(\bm x)\right\|^2_{L^2_{\rho_{\mathcal{X}}}},~ \forall \lambda > 0\,,
\end{equation*}
which is also called as the \emph{maximum random features dimension} \cite{Rudi2017Generalization}.
By defintion we always have  $\mathcal{N}(\lambda) \leq \mathcal{F}_{\infty}(\lambda)$.
Roughly speaking, when the random features are ``good", it is easy to control their leverage scores in terms of the decay of the spectrum of $\Sigma$.
Further, fast learning rates using fewer random features can be achieved if the features are  \emph{compatible} with the data distribution in the following sense.
\begin{assumption}
[Compatibility condition \cite{Rudi2017Generalization}]
\label{optimalrff}
 With the above definition of $\mathcal{F}_{\infty}(\lambda)$, assume that there exist $\varrho \in [0,1]$, and $F > 0$ such that $\mathcal{F}_{\infty}(\lambda) \leq F \lambda^{-\varrho}, \forall \lambda >0$.
\end{assumption}
It always holds that $\mathcal{F}_{\infty}(\lambda) \leq \kappa^2 \lambda^{-1}$ when $z$ is uniformly bounded by $\kappa$.
So the worst case is $\varrho=1$, which means that the random features are sampled in a problem independent way.
The favorable case is $\varrho = \gamma$, which means that $\mathcal{N}(\lambda) \leq \mathcal{F}_{\infty}(\lambda) \leq \mathcal{O}(n^{-\alpha \gamma})$.
In \cite{sun2018but}, the authors consider the following assumption.
\begin{assumption}
[Optimized distribution \cite{sun2018but}]
\label{optdist}
        The feature mapping $z(\bm \omega, \bm x)$ is called \emph{optimized} if there is a small constant $\lambda_0$ such that for any $\lambda \leq \lambda_0$, 
$
                \mathcal{F}_{\infty}(\lambda) \leq \mathcal{N}(\lambda) = \sum_{i=1}^{\infty} \frac{\lambda_i(\Sigma)}{\lambda_i(\Sigma) + \lambda}$.
\end{assumption}
Under the previous definitions, Assumption~\ref{optdist} holds only when $\mathcal{F}_{\infty}(\lambda) = \mathcal{N}(\lambda)$. This assumption  is stronger than the compatibility condition in Assumption~\ref{optimalrff}. Note that Assumption~\ref{optdist} is satisfied when sampling from $q(\bm \omega)$.

\begin{table*}[!htb]
        \centering
        \fontsize{7}{8}\selectfont
        \begin{threeparttable}
                \caption{Comparison of learning rates and required random features for expected risk with the squared loss function.}
                \label{tabkrr}
                \begin{tabular}{cccccccccccccccccccc}
                        \toprule
                        sampling scheme &Results & key assumptions &eigenvalue decays &$\lambda$ &learning rates  & required $s$   \cr
                        \midrule
                        \multirow{13}{*}{$\{ {\bm \omega}_i \}_{i=1}^s \sim p({\bm \omega})$} &
                         \cite[Theorem~1]{Rudi2017Generalization} &- &  - &  $n^{-\frac{1}{2}}$ &  $\mathcal{O}_{p}\left( n^{-\frac{1}{2}} \right) $ & $s \geq \Omega(\sqrt{n} \log n)$   \cr
                        \cmidrule(lr){2-7}
                        &\multirow{5}{2.5cm}{\centering{
                                         \cite[Theorem~2]{Rudi2017Generalization} }} &\multirow{4}{2cm}{source condition} &\multirow{1}{*}{$i^{-2t}$} & $n^{-\frac{2t}{1+4r t}}$ &  $\mathcal{O}_{p}\left( n^{-\frac{4rt}{1+4rt}} \right) $ & $s \geq \Omega(\frac{2t+2r-1}{1+4rt}\log n)$    \cr
                        \cmidrule(lr){4-7}
                        &&&\multirow{1}{*}{$1/i$} &  $n^{-\frac{1}{2r+1}}$ &   $\mathcal{O}_{p}\left( n^{-\frac{2r}{2r+1}} \right) $ & $s \geq \Omega(n^{\frac{2r}{2r+1}} \log n)$   \cr
                        \cmidrule(lr){2-7}
                        &\multirow{5}{2.5cm}{\centering{
                                         \cite[Corollary 2]{li2019towards} }} &\multirow{5}{*}{-} &  \multirow{1}{*}{$e^{-\frac{1}{c}i}$} &  $n^{-\frac{1}{2}}$ &  $\mathcal{O}_{p}\left( n^{-\frac{1}{2}} \right) $ & $s \geq \Omega(\sqrt{n} \log\log n)$   \cr
                        \cmidrule(lr){4-7}
                        &&&\multirow{1}{*}{$i^{-2t}$} & $n^{-\frac{1}{2}}$ &  $\mathcal{O}_{p}\left( n^{-\frac{1}{2}} \right) $ & $s \geq \Omega(\sqrt{n} \log n)$    \cr
                        \cmidrule(lr){4-7}
                        &&&\multirow{1}{*}{$1/i$} &  $n^{-\frac{1}{2}}$ &   $\mathcal{O}_{p}\left( n^{-\frac{1}{2}} \right) $ & $s \geq \Omega(\sqrt{n} \log n)$   \cr
                        \midrule
                        \multirow{11}{*}{$\{ {\bm \omega}_i \}_{i=1}^s \sim q({\bm \omega})$}
                        &\multirow{5}{2.5cm}{\centering{
                                         \cite[Theorem~3]{Rudi2017Generalization} }} &\multirow{4}{2.5cm}{source condition; compatibility condition} &\multirow{1}{*}{$i^{-2t}$} & $n^{-\frac{2t}{1+4r t}}$ &  $\mathcal{O}_{\!q\!}\!\left( n^{-\frac{4rt}{1+4rt}} \right) $ & $s \geq \Omega(\frac{\varrho +(2r-1)(2t+1-2t\varrho)}{1+4rt}\log n)$    \cr
                        \cmidrule(lr){4-7}
                        &&&\multirow{1}{*}{$1/i$} &  $n^{-\frac{1}{2r+1}}$ &   $\mathcal{O}_{\!q\!}\!\left( n^{-\frac{2r}{2r+1}} \right) $ & $s \geq \Omega(n^{\frac{2r}{2r+1}} \log n)$   \cr
                        \cmidrule(lr){2-7}
                        &\multirow{5}{2.5cm}{\centering{
                                         \cite[Corollary~1]{li2019towards}}} &\multirow{5}{*}{optimized distribution} &  \multirow{1}{*}{$e^{-\frac{1}{c}i}$} &  $n^{-\frac{1}{2}}$ &  $\mathcal{O}_{\!q\!}\!\left( n^{-\frac{1}{2}} \right)$ & $s \geq \Omega(\log^2 n )$   \cr
                        \cmidrule(lr){4-7}
                        &&&\multirow{1}{*}{$i^{-2t}$} & $n^{-\frac{1}{2}}$ &  $\mathcal{O}_{\!q\!}\!\left( n^{-\frac{1}{2}} \right)$ & $s \geq \Omega(n^{1/(2t)} \log n)$   \cr
                        \cmidrule(lr){4-7}
                        &&&\multirow{1}{*}{$1/i$} &  $n^{-\frac{1}{2}}$ &  $\mathcal{O}_{\!q\!}\!\left( n^{-\frac{1}{2}} \right)$ & $s \geq \Omega(\sqrt{n} \log n)$   \cr
                        \bottomrule
                \end{tabular}
        \end{threeparttable}
    \vspace{-0.0cm}
\end{table*}

\underline{\bf Source condition on $f_{\rho}$:} The following assumption states that  $f_{\rho}$ has some desirable regularity properties. 
\begin{assumption}
[Source condition  \cite{smale2007learning,Rudi2017Generalization}]
\label{regularity}
         There exist $1/2 \leq r \leq 1$ and $g \in {L_{\rho_{\mathcal{X}}}^{2}}$ such that
$
        f_{\rho}(\bm x) = (\Sigma^r g) (\bm x)$ almost surely.
\end{assumption}
Since $\Sigma$ is a compact positive operator on $L_{\rho_{\mathcal{X}}}^2$, its $r$-th power $\Sigma^r$ is well defined for any $r > 0$.\footnote{A more general condition ($r>0$) is often considered in approximation theory; see \cite{Guo2017Optimal,lin2017distributed}.}
Assumption~\ref{regularity} imposes a form of regularity/sparsity of $f_{\rho}$, which requires the expansion of $f_{\rho}$ on the basis given by the integral operator $\Sigma$.
Note that this assumption is more stringent  than the existence of $f_{\rho}$ in $\mathcal{H}$. The latter is equivalent to Assumption~\ref{regularity} with $r=\frac{1}{2}$ (the worst case), in which case $f_{\rho} \in \mathcal{H}$ need not have much regularity/sparsity.

{\bf \underline{Noise Condition:}} The following two assumptions on noise are considered in random features for classification.
\begin{assumption}
	[Massart's low noise condition \cite{sun2018but}]
	\label{noiseass}
	There exists $V \geq 2$ such that
	\begin{equation*}
		\left| \mathbb{E}_{(\bm x, y) \sim \rho} [y| \bm x] \right| \geq 2/V\,.
	\end{equation*}
\end{assumption}

\begin{assumption}
	[Separation condition \cite{sun2018but}]
	\label{sepacon}
	The points in $\mathcal{X}$ can be collected into two sets according to their labels as follows
	\begin{equation*}
		\begin{aligned} X_{1} & :=\{\bm x \in \mathcal{X}: \mathbb{E}[y |\bm x]>0\} \, ,\\
			X_{-1} & :=\{\bm x \in \mathcal{X}: \mathbb{E}[y | \bm x]<0\}\,.\end{aligned}
	\end{equation*}
	For $i\in\{\pm1\}$, the distance of a point $\bm x \in X_{i}$ to the set $X_{-i}$ is denoted by $\Delta(\bm x)$.
	We say that the data distribution satisfies a separation condition if there exists $\Delta > 0$ such that $\rho_X (\Delta(\bm x) < c) = 0$.
\end{assumption}
The above two assumptions, both controlling the noise level in the labels, can be cast under  into a unified framework \cite{koltchinskii2011oracle} as follows.
Define the \emph{regression function}  $\eta(\bm x) = \mathbb{E}[y|X=\bm x]$ in binary classification problems. The Massart's low noise condition means that  there exists $h \in (0,1]$ such that for  $|\eta(\bm x )| \geq h$ for all $\bm x \in \mathcal{X}$. Here $h$ characterizes the level of noise in classification problems. If small $h$ is small, then $\eta(\bm x)$ is close to zero, in which case correct classification is difficult.
Massart's condition can be extended to the following more flexible condition known as Tsybakov's low noise assumption \cite{koltchinskii2011oracle}. This assumption stipulates that there exists a constant $C > 0$ such that for all sufficiently small $t > 0$, we have
\begin{equation*}
	\operatorname{Pr} \big(\{\bm x \in \mathcal{X}:|2 \eta(\bm x)-1| \leq t\} \big) \leq C \cdot t^{q} \,,
\end{equation*}
for some $q > 0$.
The separation condition in Assumption~\ref{sepacon} is an extreme case of the Tsybakov’s noise assumption with $q = \infty$.
It is clear that noise-free distributions satisfy this separation assumption, since the conditional probability $\eta$ is bounded away from $1/2$.

\subsubsection{Squared loss in KRR}

In this section, we review theoretical results on the generalization properties of KRR with squared loss and random features, for both the $p(\bm \omega)$-sampling (data-independent) and $q(\bm \omega)$-sampling (data-dependent) settings.
Table~\ref{tabkrr} summarizes these results for the excess risk in terms of the key assumptions imposed, the learning rates, and the required number of random features.

We begin with the remarkable result by Rudi and Rosasco \cite{Rudi2017Generalization}. They are among the first to show that under some mild assumptions and appropriately chosen parameters, $\Omega(\sqrt{n} \log n)$ random features suffice for KRR to achieve minimax optimal rates. 
\begin{theorem}
[Generalization bound; Theorem~3 in \cite{Rudi2017Generalization}]
\label{krror}
        
        Suppose that  Assumption~\ref{regularity} (source condition) holds with $r \in [\frac{1}{2},1]$,  Assumption~\ref{optimalrff} (compatibility) holds with $\varrho \in [0,1]$, and Assumption~\ref{effectass} (capacity) holds with $\gamma \in [0,1]$.
        Assume that $n \geq n_0$ and choose $\lambda := n^{\frac{1}{2r+\gamma}}$. If the number of random features satisfies
        \begin{equation*}
s \geq c_{0} n^{\frac{\alpha+(2 r-1)(1+\gamma-\alpha)}{2 r+\gamma}} \log \frac{108 \kappa^{2}}{\lambda \delta} \,,
        \end{equation*}
        then the excess risk of $\tilde{f}_{\bm{z},\lambda}$ can be upper bounded as
        \begin{equation*}
        \mathcal{E}\left(\tilde{f}_{\bm{z},\lambda}\right)-\mathcal{E}\left(f_{\rho}\right) = \left\| \tilde{f}_{\bm{z},\lambda} - f_{\rho} \right\|^2_{L_{\rho_{\mathcal{X}}}^{2}}  \leq c_{1} \log ^{2} \frac{18}{\delta} n^{-\frac{2 r}{2 r+\gamma}} \,,
        \end{equation*}
        where $c_0$, $c_1$ are constants independent of $(n$, $\lambda$, $\delta)$, and $n_0$ does not depends on $n$, $\lambda$, $f_{\rho}$, or $\rho$.
\end{theorem}
Theorem~\ref{krror} unifies several results in \cite{Rudi2017Generalization} that impose different assumptions. 
The simplest result is Theorem~1 in \cite{Rudi2017Generalization}, which only requires the three basic Assumptions~\ref{existass}--\ref{outputs} on existence, boundedness and continuity, corresponding to the the worst case of Theorem~\ref{krror} with $\varrho = \gamma = 1$ and $r=1/2$.
In this case, by choosing $\lambda=n^{-1/2}$, we require $\Omega(\sqrt{n} \log n)$ random features to achieve the minimax convergence rate $\mathcal{O}(n^{-1/2})$; also see Table~\ref{tabkrr}.

A more refined result is given in Theorem~2 in \cite{Rudi2017Generalization}, which accounts for the capacity of the RKHS and the regularity of $f_{\rho}$, as quantified by the parameters $\gamma \in [0,1]$ (Assumption~\ref{effectass}) and $r \in [\frac{1}{2},1]$ (Assumption~\ref{regularity}), respectively.
Under these conditions and choosing $\lambda:=n^{-\frac{1}{2r+\gamma}}$, we require $\Omega\big(n^{\frac{1+\gamma (2r-1)}{2r+\gamma}} \log n \big)$ random features to achieve the convergence rate $\mathcal{O}\big(n^{-\frac{2r}{2r+\gamma}}\big)$.
Note that $\gamma = 1$ is the worst case, where the eigenvalues of $\bm K$ have the slowest decay, and $\gamma = 1/(2a) \in (0,1)$ means that the eigenvalues follow a polynomial decay $\lambda_i \propto ni^{-2a}$.
Table~\ref{tabkrr} presents this result with $\gamma := 1/(2a)$ for better comparison with the other results.

The above two results apply to the standard RFF setting with data-independent sampling.
When $\{ \bm \omega_{i} \}_{i=1}^s$ are sampled from a data-dependent distribution satisfying the compatibility condition in Assumption~\ref{optimalrff} with $\varrho \in [0,1]$, then Theorem~3 in \cite{Rudi2017Generalization} provide an improved result. In this case, by choosing $\lambda:=n^{-\frac{1}{2r+\gamma}}$, we require $\Omega\big(n ^{\frac{\varrho+(1+\gamma-\varrho)(2 r-1)}{2 r+\gamma}} \log n\big)$ random features to achieve the convergence rate $\mathcal{O}\big(n^{-\frac{2r}{2r+\gamma}}\big)$.

If the compatibility condition is replaced by the stronger Assumption~\ref{optdist} (optimized distribution), satisfied by $q(\bm \omega)$-sampling, the work \cite{li2019towards} derives an improved bound that is the sharpest to date. Below we state a general result from~\cite{li2019towards} that covers both $p(\bm \omega)$- and $q(\bm \omega)$-sampling.
\begin{theorem}
[Theorem~1 in \cite{li2019towards}]
\label{maintheo}
        Suppose that the regularization parameter $\lambda$ satisfies $0 \leq n \lambda \leq \lambda_1$.
       We consider two sampling schemes.
       \begin{itemize}
       \item $\{ {\bm \omega}_i \}_{i=1}^s \sim p(\bm \omega)$: if $s \geq (5 z_0^2/\lambda)  \log ({16 d_{\bm K}^{\lambda}}/{\delta})$ and $|z(\bm \omega, \bm x)| \leq z_0$,
            \item $\{ {\bm \omega}_i \}_{i=1}^s \sim q(\bm \omega)$: if $s \geq 5 d^{\lambda}_{\bm{K}} \log \left(16 d_{\bm{K}}^{\lambda} / \delta \right) $,
        \end{itemize}
        then for $0 < \delta <1$, with probability $1-\delta$, the excess risk of $\tilde{f}_{\bm{z},\lambda}$ can be upper bounded as
        \begin{equation}\label{lrli}
        \left\| \widetilde{f}_{\bm{z},\lambda} - f_{\rho} \right\|^2_{L_{\rho_{\mathcal{X}}}^{2}} \leq 2 \lambda+\mathcal{O}(1 / \sqrt{n})+\mathcal{E}\big(f_{\bm{z},\lambda} \big)-\mathcal{E}\left(f_{\rho}\right) \,,
        \end{equation}
        where we recall that $\mathcal{E}\big(f_{\bm{z},\lambda} \big)-\mathcal{E}\left(f_{\rho}\right) $ is the excess risk of standard KRR with an exact kernel (see Section~\ref{sec:overviewRFF}).
\end{theorem}
{\bf Remark:} A sharper convergence rate can be achieved if the Rademacher complexity used in \cite{li2019towards} is substituted by the local Rademacher complexity \cite{bartlett2005local}, see \cite{li2021jmlr} for details.

For $p(\bm \omega)$-sampling, Theorem~\ref{maintheo} improves on the results of \cite{Rudi2017Generalization} under the exponential and polynomial decays.
Specifically, if $\{ {\bm \omega}_i \}_{i=1}^s \sim p(\bm \omega)$, Theorem~\ref{maintheo} requires $s \propto 1/\lambda \log d^{\lambda}_{\bm{K}}$.
Specialized to the exponential decay case, this result requires $\Omega(\sqrt{n} \log \log n)$ random features to achieve an $\mathcal{O}(n^{-1/2})$ learning rate, which is an improvement compared to \cite{Rudi2017Generalization} with $\Omega(\sqrt{n} \log n)$ random features.

For $q(\bm \omega)$-sampling, Theorem~\ref{maintheo} shows that if $\lambda= n^{-1/2}$,  then $s \propto d^{\lambda}_{\bm{K}} \log d^{\lambda}_{\bm{K}}$ random features is sufficient to incurs no loss in the expected risk if KRR, with a minimax learning rate $\mathcal{O}(n^{-1/2})$. Corollaries of this result under three different regimes of eigenvalue decay are summarized in Table~\ref{tabkrr}.

Carratino et al. \cite{carratino2018learning} extend the result of \cite{Rudi2017Generalization} to the setting where KRR is solved by stochastic gradient descent (SGD).
They show that under the basic Assumptions~\ref{existass}--\ref{outputs} and some mild conditions for SGD, $\Omega(\sqrt{n})$ random features suffice to achieve the minimax learning rate $\mathcal{O}(n^{-1/2})$.
This result matches those for standard KRR with an exact kernel \cite{steinwart2007fast}. 
The above results can be improved if in addition the source condition in Assumption \ref{regularity} holds, in which case  $\Omega(n^{\frac{1+\alpha(2r-1)}{2r+ \alpha}})$ random features suffice to achieve an  $\mathcal{O}(n^{-\frac{2r}{2r+\alpha}})$ learning rate.

The work in \cite{wang2019simple} shows that if the randomized feature map is bounded (which is weaker than Assumption~\ref{rfbc}), then we have the following out-of-sample bound
\begin{equation*}
	\mathcal{E} ( \widetilde{f_{\bm{z},\lambda}} )  - \mathcal{E} ( f_{\bm{z},\lambda} ) \leq \mathcal{O} \left(\frac{1}{s \lambda} \right) \,.
\end{equation*}
If we choose $\lambda:= n^{-1/2}$, then $\Omega(n)$ random features are sufficient to ensure an $\mathcal{O}(n^{-1/2})$ rate in the out-of-sample bound.

\begin{table*}[tp]
	\centering
	\fontsize{7}{8}\selectfont
	\begin{threeparttable}
		\caption{Comparison of learning rates and required random features for expected risk with a Lipschitz continuous loss function.}
		\label{tablisex}
		\begin{tabular}{cccccccccccccccccccc}
			\toprule
			sampling scheme &Results & key assumptions &eigenvalue decay &$\lambda$ &learning rates  & required $s$   \cr
			\midrule
			\multirow{7}{*}{$\{ {\bm \omega}_i \}_{i=1}^s \sim p({\bm \omega})$} &
			\cite[Theorem 1]{rahimi2009weighted} &- &  - &  - &  $\mathcal{O}_{p}\left( n^{-\frac{1}{2}} \right) $ & $s \geq \Omega(n \log n)$   \cr
			\cmidrule(lr){2-7}
			&\multirow{5}{2.5cm}{\centering{
					\cite[Corollary 4]{li2019towards} }} &\multirow{5}{*}{-} &  \multirow{1}{*}{$e^{-\frac{1}{c}i}$} &  $\frac{1}{n}$ & $\mathcal{O}_{p}\left( n^{-\frac{1}{2}} \right) $ & $s \geq \Omega(n \log \log n)$ \cr
			\cmidrule(lr){4-7}
			&&&\multirow{1}{*}{$i^{-2t}$} &  $\frac{1}{n}$ & $\mathcal{O}_{p}\left( n^{-\frac{1}{2}} \right) $ & $s \geq \Omega(n \log n)$ \cr
			\cmidrule(lr){4-7}
			&&&\multirow{1}{*}{$1/i$} &  $\frac{1}{n}$ & $\mathcal{O}_{p}\left( n^{-\frac{1}{2}} \right) $ & $s \geq \Omega(n \log n)$ \cr
			\midrule
			\multirow{15}{*}{$\{ {\bm \omega}_i \}_{i=1}^s \sim q({\bm \omega})$}
			&\multirow{7}{2.5cm}{\centering{
					 \cite[Theorem~1]{sun2018but} }} &\multirow{9}{*}{optimized distribution}  &  \multirow{1}{*}{$e^{-\frac{1}{c}i}$} &  $\frac{1}{n}$ &  $\mathcal{O}_{\!q\!}\!\left( \frac{1}{n} \log^{c+2} n \right)$ & $s \geq \Omega(\log^c n \log\log^c n)$\\
			&&low noise condition&&&& \cr
			\cmidrule(lr){4-7}
			&&&\multirow{1}{*}{$i^{-2t}$} & $n^{-\frac{t}{1+t}}$ &  $\mathcal{O}_{\!q\!}\!\left( n^{-\frac{t}{1+t}} \log n \right) $ & $s \geq \Omega(n^{\frac{1}{1+t}} \log n)$   \cr
			\cmidrule(lr){4-7}
			&&&\multirow{1}{*}{$1/i$} &  $\frac{1}{n}$ &   $\mathcal{O}_{\!q\!}\!\left( n^{-\frac{1}{2}} \right) $ & $s \geq \Omega(n \log n)$   \cr
			\cmidrule(lr){2-7}
			&
			\multirow{2}{*}{\cite[Theorem~2]{sun2018but}} &separation condition &  \multirow{2}{*}{$e^{-\frac{1}{c}i}$} &  \multirow{2}{*}{$n^{-2c^2}$} &  \multirow{2}{*}{$\mathcal{O}_{\!q\!}\!\left( \frac{1}{n} \log^{2c+1} n \log \log n \right)$} & \multirow{2}{*}{$s \geq \Omega(\log^{2c} n \log\log n)$} \\
			&&optimized distribution&&&&  \cr
			\cmidrule(lr){2-7}
			&\multirow{5}{2.5cm}{\centering{
					\cite[Section 4.5]{bach2017equivalence}\\
					\cite[Corollary~3]{li2019towards} }} &\multirow{5}{*}{optimized distribution} &  \multirow{1}{*}{$e^{-\frac{1}{c}i}$} &  $\frac{1}{n}$ &  $\mathcal{O}_{\!q\!}\!\left( n^{-\frac{1}{2}} \right) $ & $s \geq \Omega(\log^2 n)$   \cr
			\cmidrule(lr){4-7}
			&&&\multirow{1}{*}{$i^{-2t}$} & $\frac{1}{n}$ &  $\mathcal{O}_{\!q\!}\!\left( n^{-\frac{1}{2}} \right) $ & $s \geq \Omega(n^{1/(2t)} \log n)$   \cr
			\cmidrule(lr){4-7}
			&&&\multirow{1}{*}{$1/i$} &  $\frac{1}{n}$ &  $\mathcal{O}_{\!q\!}\!\left( n^{-\frac{1}{2}} \right) $ & $s \geq \Omega(n \log n)$   \cr
			\bottomrule
		\end{tabular}
	\end{threeparttable}
\vspace{-0.0cm}
\end{table*}

\vspace{-0.0cm}
\subsubsection{Lipschitz continuous loss function}
In this section, we consider loss functions $\ell$ that are Lipschitz continuous. Examples include the hinge loss in SVM and the cross-entropy loss in kernel logistic regression.
Table~\ref{tablisex} summarizes several existing results for such loss functions in terms of the learning rate and the required number of random features.
We briefly discuss these results below and refer the readers to the cited work for the precise theorem statements.

If $\{ \bm \omega_i \}_{i=1}^s \sim p(\bm \omega)$, i.e., under the standard RFF setting with data-independent sampling, we have the following results.
\begin{itemize}
        \item Theorem~1 in  \cite{rahimi2009weighted} shows that the excess risk converges at a certain $\mathcal{O}(n^{-1/2})$ rate with $\Omega(n \log n)$ random features.
        \item Corollary~4 in \cite{li2019towards} shows that with $\lambda \in \mathcal{O}(1/n)$ and  $\Omega\big((1/\lambda) \log d_{\bm K}^{\lambda}\big)$ random features, the excess risk of $\tilde{f}_{\bm{z},\lambda}$ can be upper bounded by
        \begin{equation*}
        \mathcal{E}(\tilde{f}_{\bm{z},\lambda})-\mathcal{E}\left(f_{\rho}\right) \leq \mathcal{O}\left({1}/{\sqrt{n}} \right) + \mathcal{O}(\sqrt{\lambda})\,.
        \end{equation*}
        The above bound scales with $\sqrt{\lambda}$, which is different from the bound in Eq.~\eqref{lrli} for the squared loss. Therefore, for  Lipschitiz continuous loss functions, we need to choose a smaller regularization parameter  $\lambda \in \mathcal{O}(1/n)$ to achieve the same  $\mathcal{O}(n^{-1/2})$ convergence rate. Also note that as before we can bound $d_{\bm K}^{\lambda}$ under the three types of eigenvalue decay.
\end{itemize}

If $\{ \bm \omega_i \}_{i=1}^s \sim q(\bm \omega)$, i.e., under the data-dependent sampling setting, we have the following results.
\begin{itemize}
        \item For SVM with random features, under the optimized distribution in Assumption~\ref{optdist} and the low noise condition in Assumption~\ref{noiseass}, Theorem~1 in \cite{sun2018but} provides bounds on the learning rates and the required number of random features. This result is improved in \cite[Theorem~2]{sun2018but} if we consider the stronger separation condition in Assumption~\ref{sepacon}. Details can be found in Table~~\ref{tablisex}.
        \item In Section~4.5 in \cite{bach2017equivalence} and Corollary~3 in \cite{li2019towards}, it is shown that if Assumption~\ref{optdist} holds, then the excess risk of $\tilde{f}_{\bm{z},\lambda}$ converges at an $\mathcal{O}(n^{-1/2})$  rate with $\Omega(d_{\bm K}^{\lambda} \log d_{\bm K}^{\lambda})$ random features, if we choose $\lambda \in \mathcal{O}(1/n)$.
\end{itemize}

There is an abnormal but common experiment phenomenon on kernel approximation and risk generalization, that is, a higher kernel approximation quality does not always translate to better generalization performance, see the discussion in \cite{avron2017random,munkhoeva2018quadrature,zhang2019f}. 
Understanding this inconsistency between approximation quality and generalization performance is an important open problem in this topic. 
Here we present a preliminary result for KRR: a better approximation quality cannot guarantee a lower generalization risk, see Proposition~\ref{proincon} as below,  with proof deferred to Appendix~\ref{app:proincon}.
\begin{proposition}\label{proincon}
	Given the target function $f_{\rho}$ and the original kernel matrix $\bm K$, consider two random features based algorithms $\operatorname{A1}$ and $\operatorname{A2}$ yielding two approximated kernel matrices $\widetilde{\bm K}_1$ and $\widetilde{\bm K}_2$, and their respective KRR estimators $\tilde{f}^{(\operatorname{A1})}_{\bm z, \lambda}$ and $\tilde{f}^{(\operatorname{A2})}_{\bm z, \lambda}$. 
	Then for a new sample $\bm x$, even if $\| \bm K - \widetilde{\bm K}_1 \| \leq  \| \bm K - \widetilde{\bm K}_2 \|$ holds in some norm metric, there exists one case for the excess risk such that 
	\begin{equation*}
	\mathcal{E} [ \tilde{f}^{(\operatorname{A1})}_{\bm z, \lambda} (\bm x) ]  - \mathcal{E} [ f_{\rho} (\bm x) ] \geq 	\mathcal{E} [ \tilde{f}^{(\operatorname{A2})}_{\bm z, \lambda} (\bm x) ]  - \mathcal{E} [ f_{\rho} (\bm x)  ]\,.
	\end{equation*} 
\end{proposition}
\noindent{\bf Remark:} Our proof is geometric by constructing a counter-example. It requires that the kernel admits (at least) polynomial decay, which holds for the common-used Gaussian kernel and could be further relaxed for the existence of the proof.  

\subsection{Results for nonlinear component analysis}

In addition to supervised learning problems such as classification and regression,  random features can also be used in unsupervised learning, e.g., randomized nonlinear component analysis.
Here we give an overview of the results for this problem.

The authors of \cite{lopez2014randomized} propose to use random features to approximate the kernel matrix in kernel Principal Component
Analysis (KPCA) and kernel Canonical Correlation Analysis (KCCA). They show that the approximate kernel matrix converges to the true one in operator norm at a rate of $\mathcal{O}(n \sqrt{\log n / s})$.
More precisely, $s=\mathcal{O}((\log n)^2 / \epsilon^2)$ suffices to ensure that $\| \widetilde{\bm K} - \bm K \|_2 \leq \epsilon n$ with the probability $1-1/n$.
Their algorithm takes $\mathcal{O}(ns^2+nsd)$ time to construct feature functions and $\mathcal{O}(s^2+sd)$ space to store the feature functions and covariance matrix.
Ghashami et al. \cite{ghashami2016streaming} combine random features with matrix sketching for KPCA.  For finding the top-$\ell$ principal components, they improve the time and space complexities to $\mathcal{O}(nsd+n \ell s)$ and $\mathcal{O}(sd+\ell s)$, respectively.
Xie et al. \cite{xie2015scale} propose to use  the doubly stochastic gradients scheme to accelerate KPCA.
The authors of  \cite{sriperumbudur2017statistical} investigate the statistical consistency of KPCA with random features.
They show that the top-$\ell$ eigenspace of the empirical covariance matrix in $\widetilde{\mathcal{H}}$ converges to the covariance operator in $\mathcal{H}$ at the rate of $\mathcal{O}(1/\sqrt{n}+1/\sqrt{s})$.
Therefore,  $\Omega(n)$ random features are required to guarantee a $\mathcal{O}(1/\sqrt{n})$ rate.
Ullah et al. \cite{ullah2018streaming} instead pose KPCA as a stochastic optimization problem and show that the empirical risk minimizer (ERM) in the random feature space converges in objective value at an  $\mathcal{O}(1/\sqrt{n})$ with $\Omega(\ell \sqrt{n} \log n)$ random features.

\begin{table}[t]
	\centering
	\caption{\footnotesize Dataset statistics.}
	\label{tabstat}
	\begin{threeparttable}
		\begin{tabular}{cccccccccccccc}
			\toprule[1.5pt]
			datasets & $d$  & \#traing & \#test & random split & scaling \\
			\midrule[1pt]
			\emph{ijcnn1} & 22 & 49,990 & 91,701 & no & - \\
			\hline 
			\emph{EEG} &14 &7,490 &7,490 & yes &mapstd
			\\
			\hline
			\emph{cod-RNA} &8 &59,535 &157,413 & no &mapstd
			\\
			\hline
			\emph{covtype} &54 &290,506 &290,506 & yes &minmax
			\\
			\hline
			\emph{magic04} &10 & 9,510 &9,510 & yes &minmax \\
			\hline
			\emph{letter} &16 & 12,000 &6,000 & no & minmax \\
			\hline
			\emph{skin} &3 & 122,529 &122,529 & yes & minmax \\
			\hline
			\emph{a8a} &123 & 22,696 &9,865 & no & - \\
			\hline
			\hline
			\emph{MNIST} &784 & 60,000 &10,000 & no & minmax \\
			\hline
			\emph{CIFAR-10} &3072 & 50,000 &10,000 & no & - \\
			\hline
			\hline
			\emph{MNIST-8M} & 784 & 8,100,000 & 10,000 & no & -\\
			\bottomrule[1.5pt]
		\end{tabular}
	\end{threeparttable}
	\vspace{-0.3cm}
\end{table}

\section{Experiments}
\label{sec:experiments}

In this section, we empirically evaluate the kernel approximation and classification performance of representative random features algorithms on several benchmark datasets.
All experiments are implemented in MATLAB and carried out on a PC with Intel$^\circledR$ i7-8700K CPU (3.70 GHz) and 64 GB RAM. The source code of our implementation can be found in \url{http://www.lfhsgre.org}.

\subsection{Experimental settings}
\label{app:expexp}
\noindent{\bf Kernel:} We choose the popular Gaussian kernel, zero/first-order arc-cosine kernels, and polynomial kernels for experiments.

i) Gaussian kernel:
\begin{equation}\label{gaussnew}
	k(\bm x, \bm x') = \exp \left(-\frac{\| \bm x - \bm x' \|_2^2}{2 \varsigma^2}\right)\,,
\end{equation}
where the kernel width parameter $\varsigma$ is tuned via 5-fold inner cross validation over a grid of $\{ 0.01,0.1,1,10, 100 \}$.

To evaluate the Gaussian kernel, we conduct the following representative algorithms for comparison: RFF \cite{rahimi2007random}, ORF \cite{Yu2016Orthogonal}, SORF \cite{Yu2016Orthogonal}, ROM \cite{choromanski2017unreasonable}, Fastfood \cite{le2013fastfood}, QMC \cite{yang2014quasi}, SSF \cite{lyu2017spherical}, GQ \cite{dao2017gaussian}, and LS-RFF \cite{li2019towards}. These algorithms include both data-independent and data-dependent approaches and involve a variety of techniques including Monte Carlo and quasi-Monte Carlo sampling, quadrature rules, variance reduction, and computational speedup using structural/circulant matrices.  

ii) arc-cosine kernels: Different from Gaussian kernels and polynomial kernels, the designed arc-cosine kernels \cite{cho2009kernel} can be closely connected to neural networks, which include feature spaces that mimic the sparse, nonnegative, distributed representations of single-layer threshold networks. 
The used zeroth order  kernel is given explicitly by
\begin{equation*}
	k(\bm x, \bm x') = 1 - \frac{\theta}{\pi} \,,
\end{equation*}
which corresponds to the Heaviside step function $\sigma({\bm \omega}^{\top} \bm{x}) = \frac{1}{2}(1+\sign({\bm \omega}^{\top} \bm{x}))$ in Eq.~\eqref{kernelfor}.
The first order kernel is
\begin{equation*}
	k(\bm x, \bm x') =  \frac{1}{\pi} \| \bm x \|_2 \| \bm x' \|_2 \left(\sin \theta + (\pi - \theta) \cos \theta \right)\,,
\end{equation*}
which corresponds to the ReLU activation function $\sigma({\bm \omega}^{\top} \bm{x}) = \max \{ 0, {\bm \omega}^{\!\top}\! \bm{x} \}$ in Eq.~\eqref{kernelfor}. 

Here we consider the zero/first-order arc-cosine kernel and compare these ten algorithms (used for Gaussian kernel approximation) as well.
Note that, the theoretical foundation behind random features, Bocher's theorem, is invalid to arc-cosine kernels.
Thankfully, according to the formulation of arc-cosine kernels admitting in Eq.~\eqref{kernelfor}, the Monte Carlo sampling (e.g., RFF) is able to used for arc-cosine kernel approximation.
In this case, the remaining algorithms, e.g., ORF, QMC, and Fastfood, on various sampling strategies, can be still applicable to arc-cosine kernels, at least in the algorithmic aspect.

iii) Polynomial kernel: This is a widely used family of dot product kernels given by
\begin{equation*}
	k(\bm x, \bm x') = (1+ \langle \bm x, \bm x' \rangle)^b \,,
\end{equation*}
where $b$ is the order. In our experiments, the order is set to $b=2$.
Note that, different from Gaussian kernels and arc-cosine kernels, polynomial kernels admit neither the Bochner's theorem nor the sampling formulation in Eq.~\eqref{kernelfor}, so classical random features based algorithms are applicable to arc-cosine kernels but still invalid to polynomial kernels even though both of them are dot-product.
As a result, algorithms for polynomial kernel approximation are often totally different. In this survey, we include three representative approaches for evaluation, including Random Maclaurin (RM) \cite{kar2012random}, Tensor Sketch (TS) \cite{Pham2013Fast}, and Tensorized Random Projection (TRP) \cite{meister2019tight}.

\noindent{\bf Datasets:} 
We consider eight non-image benchmark datasets, two representative image datasets, and a ultra-large scale dataset for evaluation.
Table~\ref{tabstat} gives an overview of these datasets including the number of feature dimension, training samples, test data, training/test split, and the normalization scheme.
These eight non-image benchmark datasets can be downloaded from \url{https://www.csie.ntu.edu.tw/~cjlin/libsvmtools/datasets/} or the UCI Machine Learning Repository\footnote{\url{https://archive.ics.uci.edu/ml/datasets.html.}}.
Some datasets include a training/test partition, denoted as ``no" in the random split column. 
For the other datasets, we randomly pick half of the data for training and the rest for testing, denoted as ``yes" in the random split column.
There are two typical normalization schemes used in these datasets: ``mapstd" and ``minmax".
The ``mapstd" scheme sets each sample's mean to 0 and deviation to 1, while the ``minmax" scheme is a standard min-max scaling operation mapping the samples to the bounded set $[0,1]^d$.
Two representative image datasets are the \emph{MNIST} handwritten digits dataset \cite{L1998Gradient} and the \emph{CIFAR10} natural image classification dataset \cite{Krizhevsky2009Learning}, summarized in the last two rows in Table~\ref{tabstat}.
The MNIST dataset contains 60,000 training samples and 10,000 test samples, each of which is a $28 \times 28$ gray-scale image of a handwritten digit from 0 to 9.
Here the ``minmax" normalization scheme means that each pixel value is divided by 255. 
The CIFAR10 dataset consists of 60,000 color images of size  $32 \times 32 \times 3$  in 10 categories, with 50,000 for training and 10,000 for test.
Besides, apart from medium/large scale datasets in our experiments, we also evaluate the compared approaches on a ultra-large scale dataset \emph{MNIST 8M} \cite{loosli2007training}, which is derived from the \emph{MNIST} dataset by random deformations and translations.
It shares the same number of feature dimension and test data with the \emph{MNIST} dataset, but has 8,100,000 training data.

\noindent{\bf Evaluation metrics:} We evaluate the performance of all the compared algorithms in terms of approximation error, time cost, and test accuracy.
We use  $ \| \bm K - \widetilde{\bm K} \|_{\mathrm{F}}/\| \bm K \|_{\mathrm{F}}$ as the error metric for kernel approximation.
A small error indicates a high approximation quality.
To compute the approximation error, we randomly sample 1,000 data points to construct the sub-feature matrix and the sub-kernel matrix.
We record the time cost of each algorithm on generating feature mappings.
The kernel width $\varsigma$ in the Gaussian kernel is tuned by five-fold cross validation over the grid $\{ 0.01,0.1,1,10,100 \}$.
The regularization parameter $\lambda$ in ridge linear regression and the balance parameter in liblinear are tuned via 5-fold inner cross validation on a grid of $\{ 10^{-8}, 10^{-6}, 10^{-4}, 10^{-3}, 10^{-2}, 0.05, 0.1, 0.5, 1, 5, 10\}$ and $\{ 0.01,0.1,1,10,100 \}$, respectively.
For the sake of computational efficiency, we conduct a relatively coarse hyper-parameter tuning.
Nevertheless, a refined hyper-parameter search might result in better classification performance.
The random features dimension $s$ in our experiments takes value in $\{2d, 4d, 8d, 16d, 32d \}$.
All experiments are repeated 10 times and we report the average approximation error, average classification accuracy with their respective standard deviations as well as the time cost for generating random features.

\begin{table*}[!htb]
	\centering
	\small
	\caption{\footnotesize Results statistics on several classification datasets. The best algorithm on each dataset is given in two cases: low dimensional (i.e., $s=2d, 4d$) and high dimensional (i.e., $s=16d, 32d$) according to approximation quality, test accuracy in linear regression or liblinear. The notation ``-" means that there is no \emph{statistically significant} difference in the performance of most algorithms.}
	\label{tabresstat}
	\begin{threeparttable}
		\begin{tabular}{cccccccccccccc}
			\toprule[1.5pt]
			\multirow{2}{0.7cm}{datasets}& \multicolumn{2}{c}{approximation} &\multicolumn{2}{c}{lr} &\multicolumn{2}{c}{liblinear}  \cr
			\cmidrule(lr){2-7}
			&small $s$ &large $s$ &small $s$  &large $s$ &small $s$  &large $s$  \\
			\midrule[1pt]
			\emph{ijcnn1} &SSF &SORF, QMC, ORF &- &- &Fastfood & - \\
			\hline 
			\emph{EEG} &SSF &ORF &- &- &-&-
			\\
			\hline
			\emph{cod-RNA} &SSF &- &- & - &- &-
			\\
			\hline
			\emph{covtype} &ORF &- &- & - &- &-
			\\
			\hline
			\emph{magic04} &SSF &SSF, ORF, QMC, ROM &- &- &- &- \\
			\hline
			\emph{letter} &SSF &SSF, ORF &- &- &- &- \\
			\hline
			\emph{skin} &SSF, ROM &QMC &- &- &- &- \\
			\hline
			\emph{a8a} &- &- &- & - &SSF &- \\

			\bottomrule[1.5pt]
		\end{tabular}
	\end{threeparttable}
\end{table*}



\vspace{-0.2cm}
\subsection{Results for the Gaussian Kernel}


\begin{figure}[!htb]
	\centering
	\subfigure[MNIST]{\label{figminst}
		\includegraphics[width=0.233\textwidth]{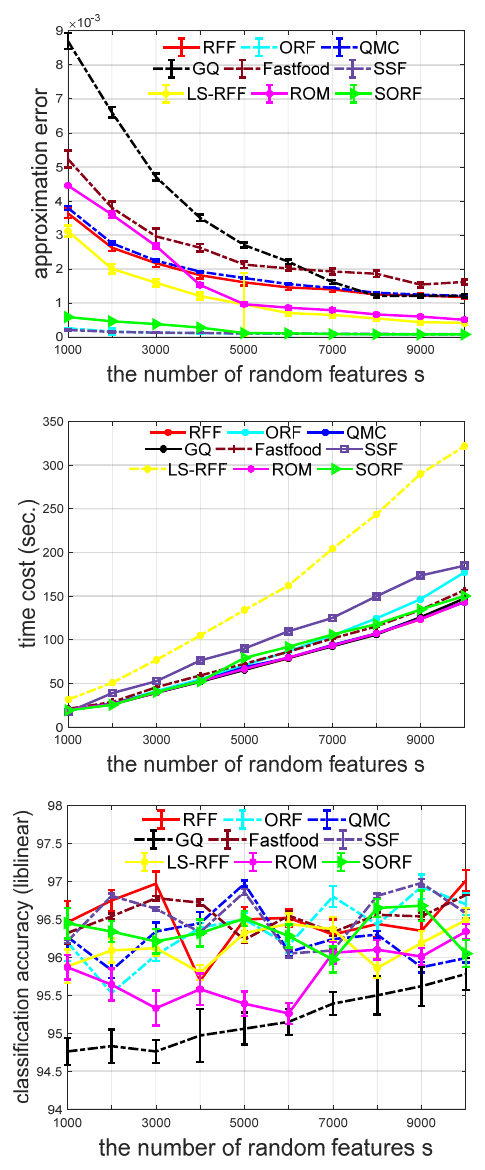}}
	\subfigure[CIFAR10]{\label{figcifar10}
		\includegraphics[width=0.238\textwidth]{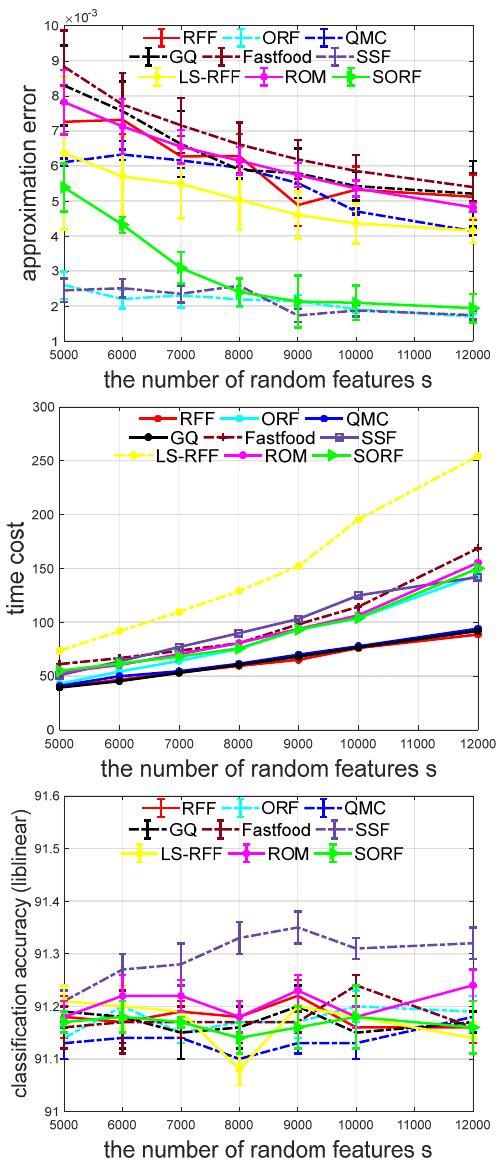}}
	\caption{Approximation error, time cost, and test accuracy of various algorithms with liblinear on two image classification datasets.}\label{figimg}
	\vspace{-0.4cm}
\end{figure}

\vspace{-0.05cm}

\subsubsection{Results on non-image benchmark datasets}
\label{sec:expgauss}
Here we test various random features based algorithms, including RFF \cite{rahimi2007random}, ORF \cite{Yu2016Orthogonal}, SORF \cite{Yu2016Orthogonal}, ROM \cite{choromanski2017unreasonable}, Fastfood \cite{le2013fastfood}, QMC \cite{yang2014quasi}, SSF \cite{lyu2017spherical}, GQ \cite{dao2017gaussian}, LS-RFF \cite{li2019towards} for kernel approximation and then combine these algorithms with lr/liblinear for classification on eight non-image benchmark datasets, refer to Appendix~\ref{app:expgauss} for details. Here we summarize the best performing algorithm on each dataset in terms of the approximation quality and classification accuracy in Table~\ref{tabresstat}, where we distinguish the small $s$ case  (i.e., $s=2d$ or $s=4d$) and the large $s$ case (i.e., $s=16d$ or $s=32d$). The notation ``-" therein means that there is no \emph{statistically significant} difference in the performance of most algorithms. 

In terms of approximation error, we find that SSF, ORF, and QMC achieve promising approximation performance in most cases.
Recall that the goal of using random features is to find a finite-dimensional (embedding) Hilbert space to approximate the original infinite-dimensional RKHS so as to preserve the inner product. 
To achieve this goal, SSF, QMC, and ORF are based on a similar principle, namely, generating random features that are as independent/complete as possible to reduce the randomness in sampling.
Regarding to SSF, we find that SSF performs well under the small $s$ case, but the significant improvement does not hold for the large $s$ case.
This might be because, a few points can be adequate in SSF, additional points (i.e., a larger $s$) may have a small marginal benefit in variance reduction under the large $s$ setting. Consequently, the approximation error of SSF sometimes stays almost the same with a larger number of random features.
QMC and ORF seek for variance reduction on random features. Nevertheless, they often work well in the large $s$ case.
As demonstrated by the expression for variance of ORF \cite{Yu2016Orthogonal} and convergence rate in QMC \cite{yang2014quasi}, this theoretical result is consistent with the numerical performance of ORF and QMC, which may explain the reason why they work better in a large $s$ setting than a small $s$ case.

Results on arc-cosine kernels and polynomial kernels can be in Appendix~\ref{app:experiments}.
Besides, apart from the above used medium/large scale datasets in our experiments, we also evaluate the compared approaches on a ultra-large scale dataset \emph{MNIST 8M} \cite{loosli2007training} with millions of data.
Due to the memory limit, following the doubly stochastic framework \cite{dai2014scalable}, we incorporate these random features based approaches under the data streaming setting for the reduction of time and space complexity.

\vspace{-0.1cm}
\subsubsection{Classification results on \emph{MNIST} and \emph{CIFAR10}}

Here we consider the MNIST and CIFAR10 datasets, on which we test these random features based algorithms for kernel approximation and then combine these algorithms with liblinear for image classification.
In our experiment, we use the Gaussian kernel\footnote{As indicated by \cite{arora2019exact,arora2020harnessing}, (convolutional) NTK generally performs better than Gaussian kernel but it is still non-trivial to obtain a efficient random features mapping for (convolutional) NTK without much loss on prediction.}, whose kernel width $\varsigma$ is tuned by 5-fold cross validation over the grid  $\varsigma=[0.01,0.1,1,10,100]$.
For the MNIST database, we directly use the original 784-dimensional feature as the data. For better performance on the CIFAR10 dataset, we use VGG16 with batch normalization \cite{ioffe2015batch} pre-trained on ImageNet \cite{Deng2009ImageNet} as a feature extractor.
We fine-tune this model on the CIFAR10 dataset with 240 epochs and a mini-batch size 64.
The learning rate is initialized at 0.1 and then divided by 10 at the 120-th, 160-th, and 200-th epochs.
For each color image, a 4096 dimensional feature vector is obtained from the output of the first fully-connected layer in this fine-tuned neural network.

Figure~\ref{figminst} shows the approximation error, the time cost (sec.), and the classification accuracy by liblinear across a range of $s=1000$ to $s=10,000$ random features on the MNIST database.
We find that ORF and SSF yield the best approximation quality. Despite that most algorithms achieve different approximation errors, there is no significant difference in the test accuracy, which corresponds to the results on non-image datasets.
Similar results are observed on the CIFAR10 dataset with $s=5000$ to $s=12,000$ random features; see Figure~\ref{figcifar10}.
Note that most algorithms take the similar time cost on generating random features except for the data-dependent algorithm LS-RFF. 
Several structured based approaches (e.g., Fastfood, SORF, ROM) do not achieve significant reduction on time cost due to the relatively inefficient Matlab built-in function to implement the Walsh-Hadamard transform.

\vspace{-0.1cm}
\section{Trends: High-dimensional Random Features in Over-parameterized settings}
\label{sec:DNNs}

\begin{figure}[t]
	\centering
	\subfigure[sonar (low-dimensioanl)]{
		\includegraphics[width=0.225\textwidth]{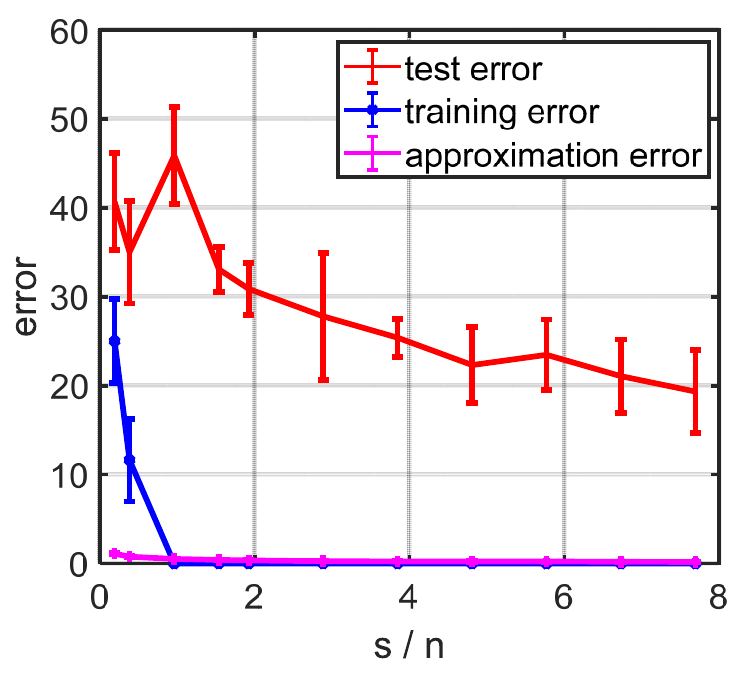}}
	\subfigure[MNIST (high-dimensional)]{
		\includegraphics[width=0.222\textwidth]{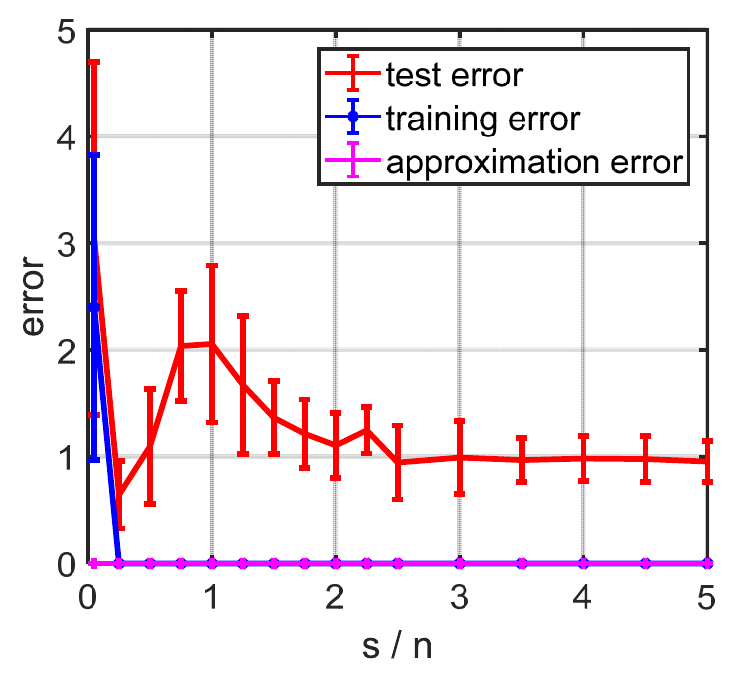}}
	\caption{Training error, test error, and approximation error of random features regression with $\lambda=10^{-8}$ on the \emph{sonar} dataset with $n=208, d=60$ and the sub-set of MNIST (class 1 versus class 2) with $n=200, d=784$.}\label{figdouble}
	\vspace{-0.2cm}
\end{figure}

In the previous sections, we review random features based algorithms and their theoretical results, that works under a fixed $d$ setting with $s \ll n$.
Random features based approaches are simple in formulation but enjoy nice empirical validations and theoretical guarantees in kernel approximation and generalization properties.
Recently, analysis of over-parameterized models \cite{hastie2019surprises,mei2019generalization,arora2019fine,liang2020multiple,liu2020kernelreg} has attracted a lot of attention in learning theory, partly due to the observation of several intriguing  phenomena, including capability of fitting random labels, strong generalization performance of overfitted classifiers \cite{zhang2016understanding} and double descent in the test error curve \cite{belkin2019reconciling,nakkiran2019deep}.
Moreover, Belkin et al. \cite{belkin2018understand,belkin2019reconciling} point out that the above phenomena are not unique to deep networks but also exist in random features and random forests.
In Figure~\ref{figdouble}, we report the empirical training error, the test error, and the kernel approximation error of random features regression as a function of $s/n$ on the \emph{sonar} dataset and the \emph{MNIST} dataset \cite{L1998Gradient}.
Even with $n$, $d$, $s$ only in the hundreds, we can still observe that as $s$ increases, the training error reduces to zero and the approximation error monotonously decreases. However, the test error exhibits double descent, i.e., a phase transition at the \emph{interpolation threshold}: moving away from this threshold on both sides trends to reduce the generalization error.
This is somewhat striking as it goes against the conventional wisdom on \emph{bias-variance trade-off}\cite{cucker2002mathematical}: predictors that generalize well should trade off the model complexity against training data fitting.

The above observations have motivated researchers to build on the elegant theory of random features to provide an analysis of neural networks in the over-parameterized regime.
To be specific, RFF can be regarded as a two-layer (large-width) neural network, where the weights in the first layer are chosen randomly/fixed and only the output layer is optimized. This is a typical over-parameterized model if we take $s \gg n$.
As such, two-layer neural networks in this regime are more amenable to theoretical analysis as compared to general arbitrary deep networks.
This is a potentially fruitful research direction, and one hand, the optimization and generalization of such model have been studied in \cite{arora2019fine,allen2019learning} in deep learning theory.
On the other hand, in order to explain the double descent curve of random features in over-parameterized regimes, we often work in a high dimensional setting, which is more subtle than classical results in standard settings, as indicated by recent random matrix theory (RMT) \cite{Tao2012Topics,pennington2017nonlinear,liao2018spectrum}. An intuitive example \cite{liao2020random} is, $\| \bm K - \bm Z \bm Z^{\!\top} \|_{\operatorname{F}} \rightarrow 0 $ always hold in low/high dimensions as $s \rightarrow \infty$ but $\| \bm K - \bm Z \bm Z^{\!\top} \|_{2} \rightarrow 0 $ does not hold for $n,d,s \rightarrow \infty$.
Accordingly, in this section, we provide an overview on analysis of (high dimensional) random features in over-parameterized setting, especially on double descent.
We remark upfront that the random features model on double descent is not the only way for analyzing DNNs.
Many other approaches, with different points of views, have been proposed for deep learning theory, but they are out of scope of this survey.

\subsection{Results on High Dimensional Random Features in Over-parameterized Setting}
\label{sec:highrff}

\begin{table*}[!htb]
	\fontsize{7}{8}\selectfont
	\begin{threeparttable}
		\caption{Comparison of problem settings on analysis of high dimensional random features on double descent.}
		\label{Tabsetting}
		\begin{tabular}{cccccccccccc}
			\toprule
			\multirow{2}{0.7cm}{studies}&\multirow{2}{0.7cm}{metric}  &\multicolumn{4}{c}{data generation} & \multirow{2}{0.9cm}{\centering{asymptotic?}} & \multirow{2}{2cm}{\centering{result}} \cr
			\cmidrule(lr){3-6}
			&&$\{ \bm x_i \}_{i=1}^n$ &$f_{\rho}$ &activation function &$\bm W$  \cr
			\midrule
			\cite[Theorem 7]{hastie2019surprises} &population risk &$\mathcal{N}(\bm 0, \bm I_d)$ &$\langle \bm x, \bm \zeta \rangle$  &normalized &$\mathcal{N}(0,1/d)$ &$\cmark$  & variance $\nearrow$ $\searrow$  \cr
			\midrule
			\cite[Theorem 4]{ba2020generalization} &population risk &$\mathcal{N}(\bm 0, \bm I_d)$ &$\langle \bm x, \bm \zeta \rangle$  &bounded &$\mathcal{N}(0,1/d)$ & $\cmark$  & variance $\nearrow$ $\searrow$  \cr
			\midrule
			\cite[Theorem 2]{mei2019generalization} &expected excess risk &$\mathbb{S}^{d-1}(\sqrt{d})$ &$\langle \bm x, \bm \zeta \rangle + \mbox{nonlinear}$~\tnote{1}  &bounded &$\mbox{Unif}(\mathbb{S}^{d-1}(\sqrt{d}))$ & $\cmark$ & variance, bias $\nearrow$ $\searrow$  \cr 
			\midrule
			\cite{d2020double} &expected excess risk &$\mathcal{N}(\bm 0, \bm I_d)$ &$\langle \bm x, \bm \zeta \rangle $  &ReLU &$\mathcal{N}(0,1)$ & $\cmark$ &  refined~\tnote{2}  \cr
			\midrule
			\cite{gerace2020generalisation} &generalization error &$\mathcal{N}(\bm 0, \bm I_d)$ &$ f(\langle \bm x, \bm \zeta \rangle) $  &general &general & $\cmark$ &  $\nearrow$ $\searrow$ \cr
			\midrule
			\cite[Theorem 1]{adlam2020understanding} &generalization error &$\mathcal{N}(\bm 0, \bm I_d)$ &$ \langle \bm x, \bm \zeta \rangle $  &normalized &$\mathcal{N}(0,1)$ & $\cmark$ &  refined~\tnote{2}  \cr
			\midrule
			\cite[Theorem 1]{dhifallah2020precise} &generalization error &$\mathcal{N}(\bm 0, \bm I_d)$ & $\langle \bm x, \bm \zeta \rangle$ & general &general & $\cmark$ &  $\nearrow$ $\searrow$ \cr
			\midrule
			\cite[Proposition 1]{hu2020universality} &generalization error &$\mathcal{N}(\bm 0, \bm I_d)$ & $\langle \bm x, \bm \zeta \rangle$ & odd, bounded &sub-Gaussian & $\cmark$ &  $\nearrow$ $\searrow$ \cr
			\midrule
			\cite[Theorem 5.1]{jacot2020implicit} &expected excess risk &Gaussian & general  &$[\cos(\cdot), \sin(\cdot)]$ &$\mathcal{N}(0,1)$ & $\xmark$ &  $\nearrow$ $\searrow$ \cr
			\midrule
			\cite[Theorem 3]{liao2020random} &generalization error &general & -~\tnote{3}  &$[\cos(\cdot), \sin(\cdot)]$ &$\mathcal{N}(0,1)$ & $\cmark$ &  $\nearrow$ $\searrow$ \cr
			\bottomrule
		\end{tabular}
	\begin{tablenotes}
		\footnotesize
		\item[1] The nonlinear component is a centered isotropic Gaussian process indexed by $\bm x \in \mathbb{S}^{d-1}(\sqrt{d})$.
		\item[2] A refined decomposition on variance is conducted by sources of randomness: ``noise variance", ``initialization variance", and ``sampling variance" to possess each term
		\cite{d2020double} or their interpretations \cite{adlam2020understanding}.
		\item[3] It makes no assumptions on $f_{\rho}$ but requires that test data ``behave" statistically like the training data by concentrated random vectors. 
	\end{tablenotes}
	\end{threeparttable}
\end{table*}

Here we briefly introduce the problem setting of high dimensional random features in over-parameterized regimes, and then discuss the techniques used in various studies.

In the basic setting, high dimensional random features often work with least squares regression setting in an asymptotic viewpoint, i.e., $n,d,s \rightarrow \infty$ with $d/n \rightarrow \psi_1 \in (0, \infty)$ and $s/n \rightarrow \psi_2 \in (0, \infty)$, in which overparameterization corresponds to $\psi_2 \geq 1$. 
The considered data generation model in the basic setting is quite simple. To be specific, the training data is collected in a matrix $\bm X \in \mathbb{R}^{n \times d}$, the rows of which are assumed to be drawn i.i.d from $\mathcal{N}(0,1)$ or $\mathbb{S}^{d-1}(\sqrt{d})$. The labels are given by a linear ground truth corrupted by some independent additive Gaussian noise: $y_i = f_{\rho}(\bm x_i) + \varepsilon_i$, where $f_{\rho}(\bm x)=\langle \bm x, \bm \zeta \rangle$ for a fixed but unknown $\bm \zeta$ and $\varepsilon_i \sim \mathcal{N}(0,1)$.
The transformation matrix under this setting is often taken as the random Gaussian matrix with the ReLU activitation function (recall Eq.~\eqref{kernelfor}).
Current approaches employ various data generation schemes and assumptions to obtain a refined analysis beyond double descent under the basic setting.
According to these criteria, we summarize the problem setting of various representative approaches in Table~\ref{Tabsetting}.
In the next, we briefly review the conceptual and technical contributions of underlying approaches on high dimensional random features.

Belkin et al. \cite{belkin2019two} begin with an one-dimensional (noise-free) version of the random features model, and provide an asymptotic analysis to explain the double descent phenomenon.
The subsequent work focuses on the standard random features model under different settings and assumptions.
It is clear that, the presence of the nonlinear activation function $\sigma(\cdot)$ makes the random features model intractable to study the related (limiting) spectral distribution.
Accordingly, the key issue in this topic mainly focuses on studying random matrices with nonlinear dependencies, e.g., how to disentangle the nonlinear function $\sigma(\cdot)$ by Gaussian equivalence conjecture.
Hastie et al.~\cite{hastie2019surprises} consider the basic setting endowed by a bounded activation function with a standardization condition, i.e., $\mathbb{E}[\sigma(t)]=0$ and $\mathbb{E}[\sigma(t)^2]=1$ for $t \sim N(0,1)$.
By establishing asymptotic results on resolvents of random block matrices from RMT, the limiting of the variance is theoretically demonstrated to be increasing for $\psi_2 \in (0,1)$, decreasing for $\psi_2 \in (1,\infty)$, and diverging as $\psi_2 \rightarrow 1$.

In a similar spirit, Mei and Montanari \cite{mei2019generalization} use RMT to study the spectral distribution of the Gram matrix $\bm Z = \sigma(\bm X \bm W^{\!\top}/\sqrt{d})/\sqrt{d}$ by considering the Stieltjes transform of a related random block matrix, and show that, under least squares regression setting in an asymptotic viewpoint, both the bias and variance have a peak at the interpolation threshold $\psi_2 = 1$ and diverge there when $\lambda \rightarrow 0$.
Under this framework, according to the randomness stemming from label noise, initialization, and training features, 
a refined bias-variance decomposition is conducted by \cite{d2020double,rocks2020memorizing} and further improved by \cite{adlam2020understanding,lin2020causes} using the \emph{analysis of variance}.
Apart from refined error decomposition schemes, the authors of \cite{ba2020generalization,gerace2020generalisation,dhifallah2020precise} consider a general setting on convex loss functions, transformation matrix, and activation functions for regression and classification.
Here the techniques used for analysis are not limited to RMT. Instead, replica method \cite{mezard1987spin} (a non-rigorous heuristic method from statistical physics) used in  \cite{d2020double,rocks2020memorizing,gerace2020generalisation} and the convex Gaussian min-max (CGMM) theorem \cite{thrampoulidis2015regularized} used in \cite{dhifallah2020precise} are two alternative way to derive the desired results.
Note that, CGMM requires the data to be Gaussian, which might restrict the application scope of their results but is still a common-used technical tool for max-margin linear classifier \cite{montanari2019generalization}, boosting classifiers \cite{liang2020precise}, and adversarial training for linear regression \cite{javanmard2020precise} in over-parameterized regimes.
Admittedly, the applied replica method in statistical physics is quite different from \cite{mei2019generalization} for tackling inverse random matrices in RMT. However, most of the above methods admit the equivalence between the considered data model and the Gaussian covariate model. That means, problem~\eqref{ermrff} with Gaussian data can be asymptotically equivalent to
\begin{equation*}
\min _{\bm \beta \in \mathbb{R}^{s}} \frac{1}{n} \sum_{i=1}^{n} \ell\left(y_{i}, \bm \beta^{\top}\left(\mu_{0} \bm{1}_{k}+\mu_{1} \bm W \bm x_{i}+\mu_{\star} \bm{t}_{i}\right)\right)+ \lambda \|\bm \beta \|_2^{2}\,,
\end{equation*}
where $\{ \bm t_i \}_{i=1}^n \sim \mathcal{N}(\bm 0, \bm I_d)$, $\mu_0 = \mathbb{E}[\sigma(t)]$, $\mu_1 = \mathbb{E}[t \sigma(t)]$ and $\mu_{\star} = \mathbb{E}[\sigma(t)^2] - \mu_0^2 - \mu_1^2$ for a standard Gaussian variable $t$.
This equivalence on generalization error in an asymptotic viewpoint is proved in \cite{hu2020universality}.

Different from the above results in an asymptotic view,  Jacot~et~al.~\cite{jacot2020implicit} present a non-asymptotic result by taking finite-size Stieltjes transform of generalized Wishart matrix, and further argue that random feature models can be close to KRR with an additional regularization. The used technical tool is related to the ``calculus of deterministic equivalents" for random matrices \cite{louart2018random}.
This technique is also used in \cite{liao2020random} to derive the exact asymptotic deterministic equivalent of $\mathbb{E}_{\bm W}[(\bm Z \bm Z^{\!\top} + n \lambda \bm I)^{-1}]$, which captures the asymptotic behavior on double descent.
Note that, this work makes no data assumption to match real-world data, which is different from previous work relying on specific data distribution.

\subsection{Discussion on Random Features and DNNs}
\label{sec:gaprfdnn}

As mentioned, random features models have been fruitfully used to analyze the double descent phenomenon.
However, it is non-trivial to transfer results for these models to practical neural networks, which are typically deep but not too wide.
There is still a substantial gap between existing theory based on random features and the modern practice of DNNs in approximation ability. 
For example, under the spherical data setting, Ghorbani et al. \cite{ghorbani2019linearized} (a more general version in \cite{mei2021generalization} on data distribution) point out that as $n \rightarrow \infty$, a random features regression model can only fit the projection of the target function onto the space of degree-$\ell$ polynomials when $s = \Omega(d^{\ell+1-\delta})$ random features are used for some $\delta > 0$. 
More importantly, if $s,d$ are taken as large with $s=\Omega(d)$, then the function space by random features can only capture linear functions.
Even if we consider the NTK model, it can just capture quadratic functions.
That means, both random features and NTK have limited approximation power in the lazy training scheme \cite{chizat2019lazy}.
In addition, Yehudai and Shamir \cite{yehudai2019power} show that the random features model cannot efficiently approximate a single ReLU neuron as it requires the number of random features to be  exponentially large in the feature dimension $d$.
This is consistent with the classical result for kernel approximation in the under-parameterized regime: the random features model, QMC, and quadrature based methods require $s=\Omega(\exp(d))$ to achieve an $\epsilon$ approximation error \cite{dao2017gaussian}.

Admittedly, the above results may appear pessimistic due to the simple architecture.
Nevertheless, random features is still an effective tool, at least the first step, for analyzing and understanding DNNs in certain regimes, and we believe its potential has yet to be fully exploited. Note that the random features model is still a strong and universal approximator \cite{sun2018approximation} in the sense that the RKHSs induced by a broad class of random features are dense in the space of continuous functions.
While the aforementioned results show that the number of required features may be exponential in the worst case, a more refined analysis can still provide useful insights for DNNs.
One potential way forward in deep learning theory is to use the random features model to analyze DNNs \emph{with pruning}.
For example, the best paper \cite{frankle2018lottery} in the \emph{Seventh International Conference on Learning Representations} (ICLR2019) put forward the following \emph{Lottery Ticket Hypothesis}: a deep neural network with random initialization contains a small sub-network which, when trained in isolation, can compete with the performance of the original one.
Malach et al. \cite{malach2020proving} provide a stronger claim that a randomly-initialized and sufficiently over-parameterized neural network contains a sub-network with nearly the same accuracy as the original one, without any further training.
Their analysis points to the equivalence between random features and the sub-network model.
As such, the random features model is potentially useful for network pruning \cite{hu2016network} in terms of, e.g., guiding the design of neurons pruning for accelerating computations, and understanding network pruning and the full DNNs. 

\section{Conclusion}
\label{sec:conclusion}

In this survey, we systematically review random features based algorithms and their associated theoretical results.
We also give an overview on generalization properties of high dimensional random features in over-parameterized regimes on double descent, and discuss the limitations and potential of random features in the theory development for neural networks.
Below we provide additional remarks and discuss several open problems that are of interest for future research.
\begin{itemize}
	\item As a typical data independent method, random features are simpler to implement, easy to parallelize, and naturally apply to streaming or dynamic data. Current efforts on Nystr\"{o}m approximation by a preconditioned gradient solver parallelized with multiple GPUs \cite{meanti2020kernel} and quantum algorithms \cite{yamasaki2020fast} can guide us to design powerful implementation for random features to handling millions/billions data.
	\item Experimental comparisons show that better kernel approximation does not directly translate to lower generalization errors. We partly answer this question in the current survey but it may be not sufficient to explain this phenomenon. We believe this issue deserves further in-depth study.
	\item Kernel learning via the spectral density is a popular direction \cite{Wilson2013Gaussian,shen2019harmonizable}, which can be naturally combined with Generative Adversarial Networks (GANs); see \cite{li2019implicit} for details. In this setting, one may associate the learned model with an implicit probability density that is flexible to characterize the relationships and similarities in the data.
	This is an interesting area for further research.
	\item 
	The double descent phenomenon has been observed and studied in random features model by various technical tools under different settings. Current theoretical results, such as those in  \cite{mei2019generalization,liao2020random}, may be extended to a more general setting with less restricted assumptions on data generation, model formulation, and the target function.
	Besides,
	more refined analysis and delicate phenomena beyond double descent have been investigated on the linear model, e.g., multiple descent phenomena \cite{chen2020multiple} and optimal (negative) regularization \cite{wu2020optimal,kobak2020optimal}. Understanding these more delicate phenomena for random features requires further investigation and refined analysis.
	\item There exist significant gaps between the random features model and practical neural networks, both in theory and empirically. Even for fitting simple quadratic or mixture models, the random features model cannot achieve a zero error with a finite number of neurons in general, while NTK and fully trained networks can \cite{ghorbani2019limitations}.
	Numerical experiments indicate that the prediction performance of NTK and CNTK may significantly degrade if the random features are generated from practically sized nets \cite{arora2019exact}.  
	\item Despite the limitations of existing theory, random features models are still useful for understanding and improving DNNs. For example, understanding the equivalence between the random features model and weight pruning in the Lottery Ticket Hypothesis \cite{malach2020proving}, may be promising future directions.
\end{itemize}
We hope that this survey will stimulate further research on the above open problems.

\vspace{-0.3cm}
\section*{Acknowledgements}
The research leading to these results has received funding from the European Research Council under the European Union's Horizon 2020 research and innovation program / ERC Advanced Grant E-DUALITY (787960). This paper reflects only the authors' views and the Union is not liable for any use that may be made of the contained information.
This work was supported in part by Research Council KU Leuven: Optimization frameworks for deep kernel machines C14/18/068; Flemish Government: FWO projects: GOA4917N (Deep Restricted Kernel Machines: Methods and Foundations), PhD/Postdoc grant. This research received funding from the Flemish Government (AI Research Program). 
This work was supported in part by Ford KU Leuven Research Alliance Project KUL0076 (Stability analysis and performance improvement of deep reinforcement learning algorithms), EU H2020 ICT-48 Network TAILOR (Foundations of Trustworthy AI - Integrating Reasoning, Learning and Optimization), Leuven.AI Institute; and in part by the National Natural Science Foundation of China 61977046, in part by National Science Foundation grants CCF-1657420 and CCF-1704828, and in part by SJTU Global Strategic Partnership Fund (2020 SJTU-CORNELL) and Shanghai Municipal Science and Technology Major Project (2021SHZDZX0102).

\clearpage
\onecolumn
\appendices

\section{Proof of Proposition~\ref{proincon}}
\label{app:proincon}

\begin{proof}
It is clear that an exact KRR estimator is $f_{\bm z, \lambda} (\bm x) = k(\bm x, \bm X) (\bm K + n \lambda \bm I)^{-1} \bm y$ and its random features based version is $ \tilde{f}_{\bm{z},\lambda}(\bm x) = \tilde{k}(\bm x, \bm X) (\widetilde{\bm K} + n \lambda \bm I)^{-1} \bm y$, where $\widetilde{\bm K} = \bm Z \bm Z^{\!\top}$ with $\bm Z \in \mathbb{R}^{n \times s}$.
The definition of the excess risk for least squares implies
\begin{equation*}
\begin{split}
\mathcal{E} ( \tilde{f}_{\bm{z},\lambda} )  - \mathcal{E} ( f_{\rho} ) & = \left[ \mathcal{E} ( \tilde{f}_{\bm{z},\lambda} ) -  \mathcal{E} ( f_{\bm{z},\lambda} ) \right] + \left[ \mathcal{E} ( f_{\bm{z},\lambda} ) - \mathcal{E} ( f_{\rho} ) \right]  = \| \tilde{f}_{\bm{z},\lambda} - f_{\bm{z},\lambda}\|^2 + \| f_{\bm{z},\lambda} - f_{\rho} \|^2 \,,
\end{split}
\end{equation*}
where the first term in the right hand is the expected error difference between the original KRR and its random features approximation version. The second term in the right hand is the excess risk of KRR, which is independent of the quality of kernel approximation.
Specifically, the first term can be further expressed by the representer theorem
\begin{equation}\label{ftilz}
	\| \tilde{f}_{\bm{z},\lambda} - f_{\bm{z},\lambda}\|^2 = \mathbb{E}_{\bm x}[ \tilde{f}_{\bm{z},\lambda}(\bm x) - f_{\bm{z},\lambda}(\bm x)]^2 = \mathbb{E}_{\bm x}\left( \sum_{i=1}^n \left[ \tilde{\alpha}_i \tilde{k}(\bm x_i, \bm x) - \alpha_i {k}(\bm x_i, \bm x) \right] \right)^2\,.
\end{equation}
Intuitively speaking, kernel approximation aims to preserve the inner product in two Hilbert spaces, i.e., $\langle k(\bm x, \cdot), k(\bm x', \cdot) \rangle_{\mathcal{H}} \approx \langle \tilde{k}(\bm x, \cdot), \tilde{k}(\bm x', \cdot) \rangle_{\widetilde{\mathcal{H}}}$.
Nevertheless, the preservation of the inner-product does not immediately guarantee a small value of $\tilde{\alpha}_i \langle \tilde{k}(\bm x, \cdot), \tilde{k}(\bm x', \cdot) \rangle_{\widetilde{\mathcal{H}}} - \alpha_i \langle k(\bm x, \cdot), k(\bm x', \cdot) \rangle_{\mathcal{H}}$ in Eq.~\eqref{ftilz}.

Informally, the proof idea is the following: define $\widetilde{\bm K} = \bm K + \bm E$ and $\tilde{k}(\bm x, \bm X) = k(\bm x, \bm X) + \tilde{\bm \epsilon}$ with the residual error matrix $\bm E \in \mathbb{R}^{n \times n}$ and the residual error vector $\tilde{\bm \epsilon} \in \mathbb{R}^{1 \times n}$ such that $\tilde{k}(\bm x, \bm X) \in \mathbb{R}^{1 \times n}$. Generally, the residual error $\bm E$ and $\tilde{\bm \epsilon} $ show the consistency, that is, a small kernel approximation error $\| \bm E \|$ implies a small $\|\tilde{\bm \epsilon}\| $. 
	Consider two random features based algorithms $\operatorname{A1}$ and $\operatorname{A2}$ yielding two approximated kernel matrices $\widetilde{\bm K}_1$ and $\widetilde{\bm K}_2$, and their respective KRR estimators $\tilde{f}^{(\operatorname{A1})}_{\bm z, \lambda}$ and $\tilde{f}^{(\operatorname{A2})}_{\bm z, \lambda}$. The corresponding residual error matrices/vectors are defined as $(\bm E_1,  \tilde{\bm \epsilon}_1)$ and $(\bm E_2,  \tilde{\bm \epsilon}_2)$ such that $\widetilde{\bm K}_1 := \bm K + \bm E_1$ and  $\widetilde{\bm K}_2 := \bm K + \bm E_2$. Without loss of generality, we assume $\| \bm E_1 \| \leq \| \bm E_2 \|$ and $\| \tilde{\bm \epsilon}_1 \| \leq \| \tilde{\bm \epsilon}_2 \|$. In this case, our target is to prove that, there exists one case such that $| \tilde{f}^{(\operatorname{A1})}_{\bm{z},\lambda}(\bm x) - f_{\bm{z},\lambda}(\bm x)| \geq | \tilde{f}^{(\operatorname{A2})}_{\bm{z},\lambda}(\bm x) - f_{\bm{z},\lambda}(\bm x)|$.
	For notational simplicity, denote $ T(\bm E, \tilde{\bm \epsilon}) := \langle \bm y^{\!\top}, k(\bm x, \bm X) (\bm K + n \lambda \bm I)^{-1} \bm E - \tilde{\bm \epsilon} \rangle $,  $ T_1(\bm E_1, \tilde{\bm \epsilon}_1) := \langle \bm y^{\!\top}, k(\bm x, \bm X) (\bm K + n \lambda \bm I)^{-1} \bm E_1 - \tilde{\bm \epsilon}_1 \rangle $, and  $ T_2(\bm E_2, \tilde{\bm \epsilon}_2) := \langle \bm y^{\!\top}, k(\bm x, \bm X) (\bm K + n \lambda \bm I)^{-1} \bm E_2 - \tilde{\bm \epsilon}_2 \rangle $.
To prove our result, we make the following three assumptions:
\begin{itemize}
	\item I. the residual matrix $\bm E$ is semi-positive definite and $\widetilde{\bm K}_1$, $\widetilde{\bm K}_2$ are non-singular.
	\item II. $n \lambda \leq \lambda_1(\widetilde{\bm K}_1) \leq \lambda_1(\widetilde{\bm K}_2)$, and $\widetilde{\bm K}_2$ admits (at least) polynomial decay.
	\item III. the inner product $\langle \bm y^{\!\top}, k(\bm x, \bm X) (\bm K + n \lambda \bm I)^{-1} \bm E - \tilde{\bm \epsilon} \rangle =: T(\bm E, \tilde{\bm \epsilon})$ is non-negative.
\end{itemize}
The above three assumptions are mild, common-used and attainable, see in \cite{li2019towards}.
Specifically, we only need to prove the existence of our claim:  there exists one case such that $| \tilde{f}^{(\operatorname{A1})}_{\bm{z},\lambda}(\bm x) - f_{\bm{z},\lambda}(\bm x)| \geq | \tilde{f}^{(\operatorname{A2})}_{\bm{z},\lambda}(\bm x) - f_{\bm{z},\lambda}(\bm x)|$ under $\| \bm E_1 \| \leq \| \bm E_2 \|$ and $\| \tilde{\bm \epsilon}_1 \| \leq \| \tilde{\bm \epsilon}_2 \|$.
Therefore, the above assumptions could be further relaxed.

According to Eq.~\eqref{ftilz}, for a new sample $\bm x$, we in turn focus on $| \tilde{f}_{\bm{z},\lambda}(\bm x) - f_{\bm{z},\lambda}(\bm x)| $, which can be upper bounded by
\begin{equation}\label{prooffzin}
\begin{split}
| \tilde{f}_{\bm{z},\lambda}(\bm x) - f_{\bm{z},\lambda}(\bm x)| & = |k(\bm x, \bm X) (\bm K + n \lambda \bm I)^{-1} \bm y - [k(\bm x, \bm X) + \tilde{\bm \epsilon}] (\bm K + \bm E + n \lambda \bm I)^{-1} \bm y | \\
& = |k(\bm x, \bm X) [(\bm K + n \lambda \bm I)^{-1} - (\bm K + n \lambda \bm I + \bm E)^{-1}]\bm y - \tilde{\bm \epsilon} (\bm K + n \lambda \bm I + \bm E )^{-1} \bm y | \\
& =  | [ k(\bm x, \bm X) (\bm K + n \lambda \bm I)^{-1} \bm E - \tilde{\bm \epsilon} ] (\bm K + n \lambda \bm I + \bm E )^{-1} \bm y | \\
& \leq [ k(\bm x, \bm X) (\bm K + n \lambda \bm I)^{-1} \bm E - \tilde{\bm \epsilon}] \bm y \sum_{i=1}^n{ \frac{1}{\lambda_i(\bm K + \bm E) + n \lambda}} \\
& = \langle \bm y^{\!\top}, ( k(\bm x, \bm X) (\bm K + n \lambda \bm I)^{-1} \bm E - \tilde{\bm \epsilon}\rangle  \sum_{i=1}^n{ \frac{1}{\lambda_i(\bm K + \bm E) + n \lambda}} =: \sum_{i=1}^n{ \frac{T(\bm E, \tilde{\bm \epsilon})}{\lambda_i(\bm K + \bm E) + n \lambda}}  \\
\end{split}
\end{equation}
where the third equality holds by $\bm A^{-1} - \bm B^{-1} = \bm A^{-1} (\bm B - \bm A) \bm B^{-1}$. The first inequality derives from $\bm a^{\!\top} \bm A \bm b \leq \bm a^{\!\top} \bm b \operatorname{tr}(\bm A) $ for two semi-positive definite matrix $\bm A$ and $\bm b^{\!\top} \bm a$ (which can be derived from the used assumptions).
Further, by virtue of $\bm a^{\!\top} \bm A \bm b \geq \lambda_n(\bm A) \operatorname{tr}(\bm a^{\!\top} \bm b) $, the error $| \tilde{f}_{\bm{z},\lambda}(\bm x) - f_{\bm{z},\lambda}(\bm x)|$ can be lower bounded by
\begin{equation}\label{prooffzinlow}
\begin{split}
| \tilde{f}_{\bm{z},\lambda}(\bm x) - f_{\bm{z},\lambda}(\bm x)| 
& =  | [ k(\bm x, \bm X) (\bm K + n \lambda \bm I)^{-1} \bm E - \tilde{\bm \epsilon} ] (\bm K + n \lambda \bm I + \bm E )^{-1} \bm y | \\
& \geq \frac{\langle \bm y^{\!\top}, [ k(\bm x, \bm X) (\bm K + n \lambda \bm I)^{-1} \bm E - \tilde{\bm \epsilon}] \rangle}{ \lambda_1(\bm K + \bm E) + n \lambda} =:   \frac{ T(\bm E, \tilde{\bm \epsilon}) }{ \lambda_1(\bm K + \bm E) + n \lambda}
\end{split}
\end{equation}
Combining Eqs.~\eqref{prooffzin} and ~\eqref{prooffzinlow}, we have
\begin{equation}\label{fzuplow}
0 \leq	\frac{T(\bm E, \tilde{\bm \epsilon})}{ \lambda_1(\bm K + \bm E) + n \lambda}  \leq | \tilde{f}_{\bm{z},\lambda}(\bm x) - f_{\bm{z},\lambda}(\bm x)| \leq  \sum_{i=1}^n{ \frac{T(\bm E, \tilde{\bm \epsilon})}{\lambda_i(\bm K + \bm E) + n \lambda}}\,.
\end{equation}

Considering such two algorithms  $\operatorname{A1}$ and $\operatorname{A2}$, under the condition of $\| \bm E_1 \| \leq \| \bm E_2 \|$ and $\| \tilde{\bm \epsilon}_1 \| \leq \| \tilde{\bm \epsilon}_2 \|$, there exists one case such that $T_1(\bm E_1, \tilde{\bm \epsilon}_1) \geq T_2(\bm E_2, \tilde{\bm \epsilon}_2)$, i.e.,
\begin{equation}\label{proofinner}
\langle \bm y^{\!\top}, ( k(\bm x, \bm X) (\bm K + n \lambda \bm I)^{-1} \bm E_1 - \tilde{\bm \epsilon}_1 \rangle \geq \langle \bm y^{\!\top}, ( k(\bm x, \bm X) (\bm K + n \lambda \bm I)^{-1} \bm E_2 - \tilde{\bm \epsilon}_2 \rangle\,,
\end{equation}
which can be achieved by a geometry explanation in Figure~\ref{figgeo}.
By virtue of Eq.~\eqref{proofinner} and Assumption II, we have
\begin{equation*}
\frac{T_1(\bm E_1, \tilde{\bm \epsilon}_1)}{\lambda_1(\bm K + \bm E_1) + n \lambda}  - \frac{T_2(\bm E_2, \tilde{\bm \epsilon}_2)}{\lambda_1(\bm K + \bm E_2) + n \lambda} =: \widetilde{C} \geq 0 \,.
\end{equation*}
The above inequality implies
\begin{equation*}
\widetilde{C}  -  \sum_{i=2}^n{ \frac{T_2(\bm E_2, \tilde{\bm \epsilon}_2)}{\lambda_i(\bm K + \bm E_2) + n \lambda}} 	\leq  \left| \tilde{f}^{\operatorname{(A1)}}_{\bm{z},\lambda}(\bm x) - f_{\bm{z},\lambda}(\bm x) \right| - \left| \tilde{f}^{\operatorname{(A2)}}_{\bm{z},\lambda}(\bm x) - f_{\bm{z},\lambda}(\bm x) \right|  \leq \widetilde{C}  +  \sum_{i=2}^n{ \frac{T_1(\bm E_1, \tilde{\bm \epsilon}_1)}{\lambda_i(\bm K + \bm E_1) + n \lambda}}\,.
\end{equation*}
The left-hand of the above inequality can be further improved as
\begin{equation*}
\begin{split}
\left| \tilde{f}^{\operatorname{(A1)}}_{\bm{z},\lambda}(\bm x) - f_{\bm{z},\lambda}(\bm x) \right| - \left| \tilde{f}^{\operatorname{(A2)}}_{\bm{z},\lambda}(\bm x) - f_{\bm{z},\lambda}(\bm x) \right| & \geq \widetilde{C}  -  \sum_{i=2}^n{ \frac{T_2(\bm E_2, \tilde{\bm \epsilon}_2)}{\lambda_i(\bm K + \bm E_2) + n \lambda}} \\
& \geq \frac{T_1(\bm E_1, \tilde{\bm \epsilon}_1)}{\lambda_1(\bm K + \bm E_2) + n \lambda} - \sum_{i=1}^n{ \frac{T_2(\bm E_2, \tilde{\bm \epsilon}_2)}{\lambda_i(\bm K + \bm E_2) + n \lambda}} \\
& \geq \frac{T_1(\bm E_1, \tilde{\bm \epsilon}_1)}{\lambda_1(\bm K + \bm E_2) + n \lambda} - \frac{T_2(\bm E_2, \tilde{\bm \epsilon}_2)}{\lambda_n(\bm K + \bm E_2)} \sum_{i=1}^n{ \frac{\lambda_i(\bm K + \bm E_2) }{\lambda_i(\bm K + \bm E_2) + n \lambda}} \\
& \geq \frac{T_1(\bm E_1, \tilde{\bm \epsilon}_1)}{2 \lambda_1(\widetilde{\bm K}_2) } - \frac{T_2(\bm E_2, \tilde{\bm \epsilon}_2)}{\lambda_n(\widetilde{\bm K}_2)}  d_{\widetilde{\bm{K}}_2}^{\lambda}\,,
\end{split}
\end{equation*}
where $d_{\widetilde{\bm{K}}_2}^{\lambda}$ is the ``effective dimension" of $\widetilde{\bm{K}}_2$ defined in Eq.~\eqref{dklambda} and the last inequality follows from Assumption II.

According to the above result, $\left| \tilde{f}^{\operatorname{(A1)}}_{\bm{z},\lambda}(\bm x) - f_{\bm{z},\lambda}(\bm x) \right| - \left| \tilde{f}^{\operatorname{(A2)}}_{\bm{z},\lambda}(\bm x) - f_{\bm{z},\lambda}(\bm x) \right| \geq 0$ holds by the following condition
\begin{equation*}
T_1(\bm E_1, \tilde{\bm \epsilon}_1) \geq 2 \underbrace{\left[ \frac{\lambda_1(\widetilde{\bm K}_2) }{\lambda_n(\widetilde{\bm K}_2)} \right] d_{\widetilde{\bm{K}}_2}^{\lambda}}_{=\mathcal{O}(1)} T_2(\bm E_2, \tilde{\bm \epsilon}_2)\,.
\end{equation*} 
We observe that, an invertible matrix $\widetilde{\bm K}_2$ admits a finite condition number $\lambda_1(\widetilde{\bm K}_2)/\lambda_n(\widetilde{\bm K}_2) < \infty$.
Besides, a fast polynomial eigenvalue decay of $\widetilde{\bm K}_2$ ensures the effective dimension $d_{\widetilde{\bm{K}}_2}^{\lambda}$ to be finite, which can be obtained by Assumption~\ref{effectass} with $\gamma = 0$.
Accordingly, in this case, when these condition are satisfied, $T_1(\bm E_1, \tilde{\bm \epsilon}_1) \geq c T_2(\bm E_2, \tilde{\bm \epsilon}_2)$ can be achieved for some constant $c$, which can be intuitively observed by Figure~\ref{figgeo}. 
Finally, we conclude the proof for existence.

	\begin{figure}[t]
	\centering
	\begin{tikzpicture}{scale=0.25}
		\coordinate (o)  at (0,0);   
		\coordinate (g1) at (-4.5,-5); 
		\coordinate (g2) at (5,5);   
		\coordinate (V1) at (2,0.7);
		\coordinate (V2) at (4,2.5);
		\coordinate (Vr) at ($(V1) + (V2)$);
		
		\coordinate (V3) at (4.5,0.4);
		\coordinate (V4) at (5.5,5);
		
		\coordinate (V1p) at (2,0);
		\coordinate (V2p) at (4,0);
		
		\coordinate (V3p) at (4.5,0);
		\coordinate (V4p) at (5.5,0);
		
		%
		\fill[poly] (V1) -- (V1p) -- (V2p) -- (V2) -- (V1);
		
		\fill[poly] (V3) -- (V3p) -- (V4p) -- (V4) -- (V3);
		%
		\draw[color=gray,thick,->] (0,0) -- (7.5,0) node[anchor=east, xshift=10]{$\bm y$};
		
		\draw[color=blue,vect] (o) -- (V1) node [anchor=east, xshift=18] {$\tilde{\bm \epsilon}_1$};
		%
		\draw[color=red,vect] (o) -- (V2) node [yshift=32,vnode] {$k(\bm x, \bm X) (\bm K + n \lambda \bm I)^{-1} \bm E_1$};
		%
		
		\draw[color=green,vect] (V1) -- (V2) node [anchor=north,xshift=2] {};
		
		\draw[color=cyan,vect] (o) -- (V3) node [anchor=east, xshift=18] {$\tilde{\bm \epsilon}_2$};
		
		\draw[color=magenta,vect] (o) -- (V4) node [anchor=north,yshift=12] {$k(\bm x, \bm X) (\bm K + n \lambda \bm I)^{-1} \bm E_2$};
		
		\draw[color=teal,vect] (V3) -- (V4) node [anchor=north,xshift=2] {};
		
		\draw[color=black,vect] (V1p) -- (V2p) node [yshift=-12,vnode] {$T_1(\bm E_1, \tilde{\bm \epsilon}_1)$};
		\draw[color=black,vect] (V3p) -- (V4p) node [yshift=-12,vnode] {$T_2(\bm E_2, \tilde{\bm \epsilon}_2)$};
	\end{tikzpicture}
	\caption{Illustration of the geometric proof for one case such that $T_1(\bm E_1, \tilde{\bm \epsilon}_1) \geq c T_2(\bm E_2, \tilde{\bm \epsilon}_2)$ under the condition of $\| \bm E_1 \| \leq \| \bm E_2 \|$ and $\| \tilde{\bm \epsilon}_1 \| \leq \| \tilde{\bm \epsilon}_2 \|$, where $c$ is some constant.}\label{figgeo}
\end{figure}
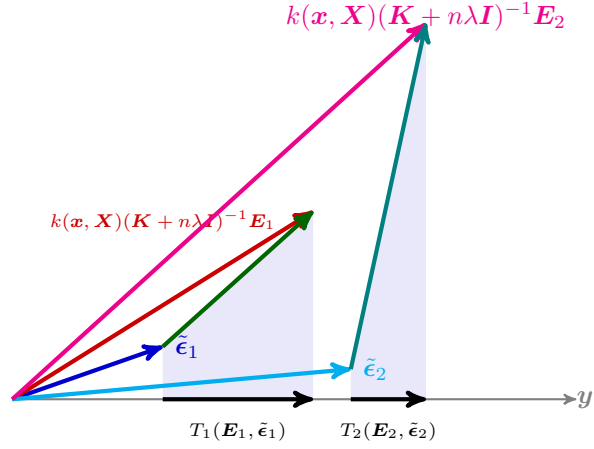

\end{proof}

\section{Experiments}
\label{app:experiments}

In this section, we detail the experimental settings and present the comparison results on the compared approaches on several benchmark datasets across various kernels. This part is organized as follows.
\begin{itemize}
	\item In Section~\ref{app:expgauss}, we present experimental results across the Gaussian kernel on eight non-image datasets in terms of approximation error, the time cost (sec.) for generating random features mappings, classification accuracy by linear regression and liblinear.
	\item Results on approximation error and test accuracy (by linear regression) across arc-cosine kernels and polynomial kernels are presented in Sections~\ref{app:exparc} and \ref{app:exppoly}, respectively.
	\item In Section~\ref{app:explarge},  a ultra-large scale dataset is applied to further validate the related algorithms.
\end{itemize}

\begin{figure*}[t]
	\centering
	\subfigure{
		\includegraphics[width=0.235\textwidth]{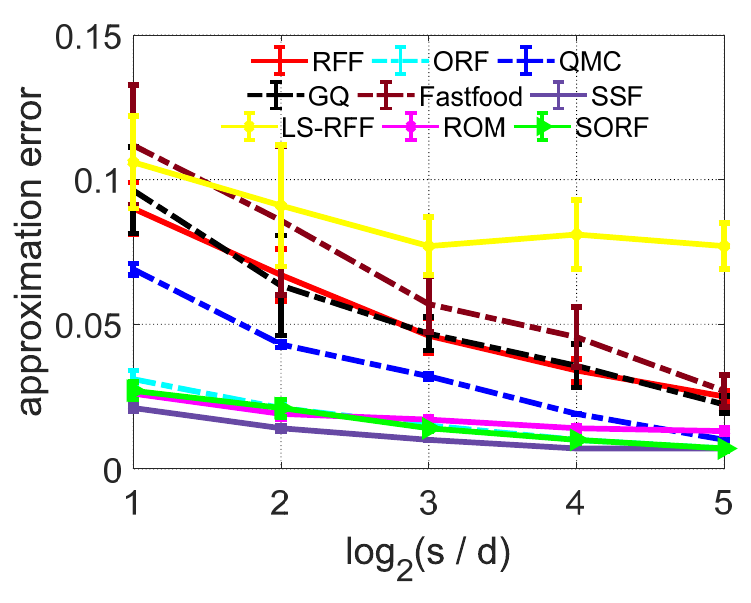}}
	\subfigure{
		\includegraphics[width=0.235\textwidth]{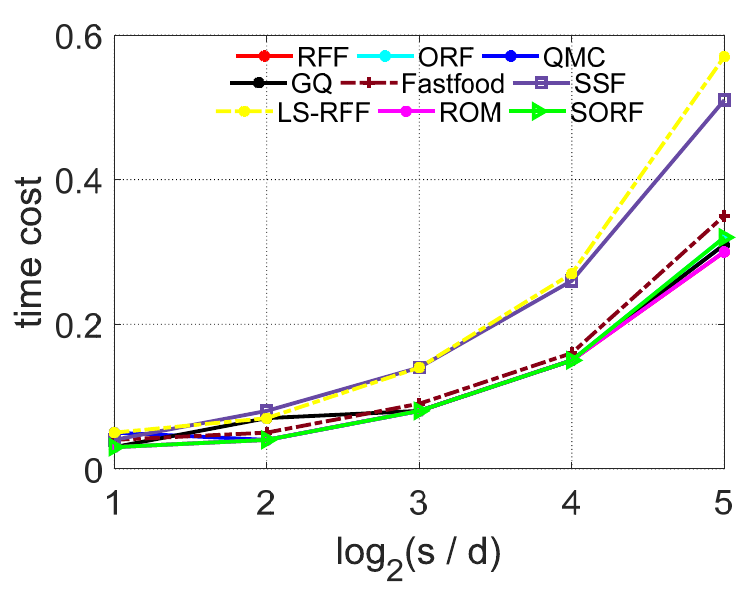}}
	\subfigure{
		\includegraphics[width=0.235\textwidth]{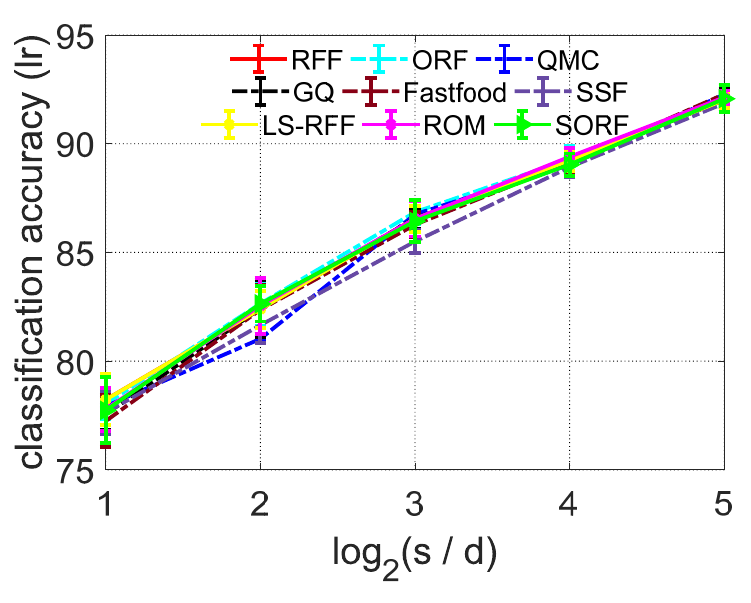}}
	\subfigure{
		\includegraphics[width=0.235\textwidth]{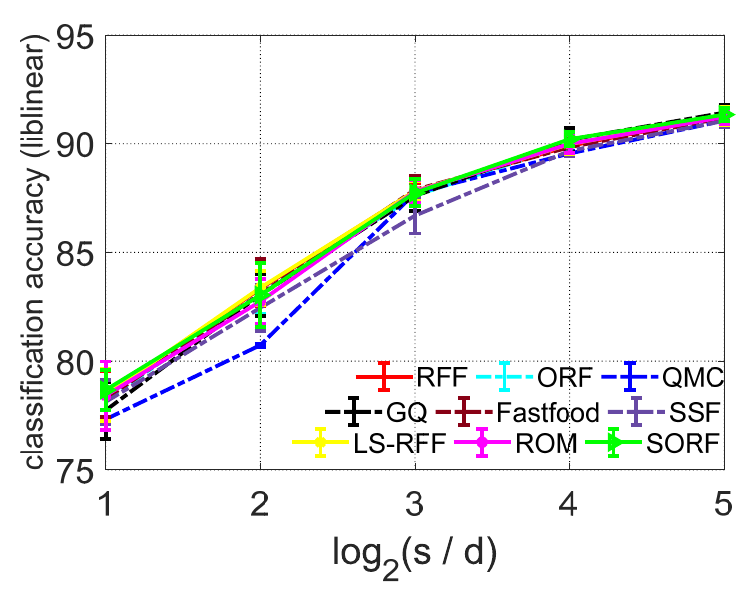}}
	
	\emph{(a) letter}
	
	\subfigure{
		\includegraphics[width=0.235\textwidth]{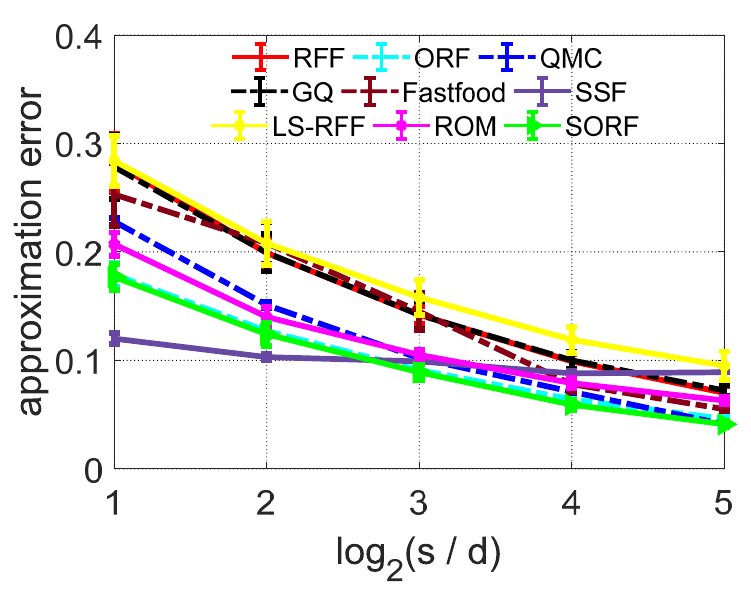}}
	\subfigure{
		\includegraphics[width=0.235\textwidth]{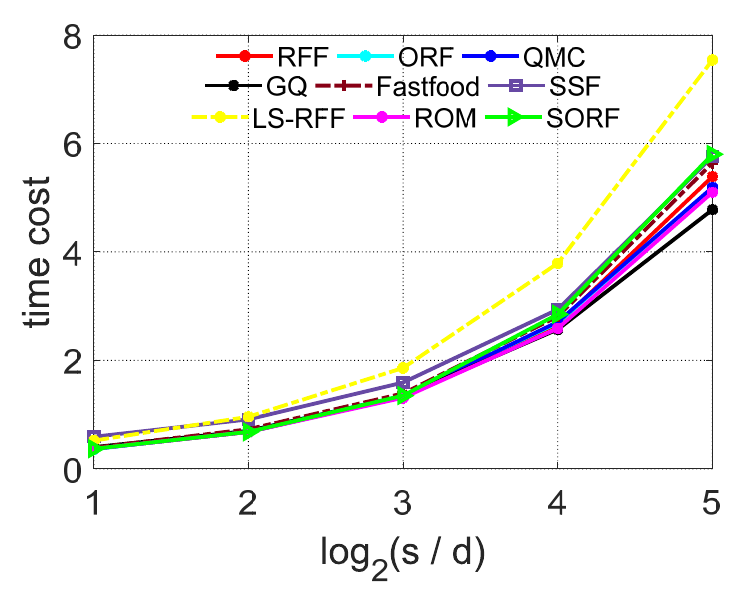}}
	\subfigure{
		\includegraphics[width=0.235\textwidth]{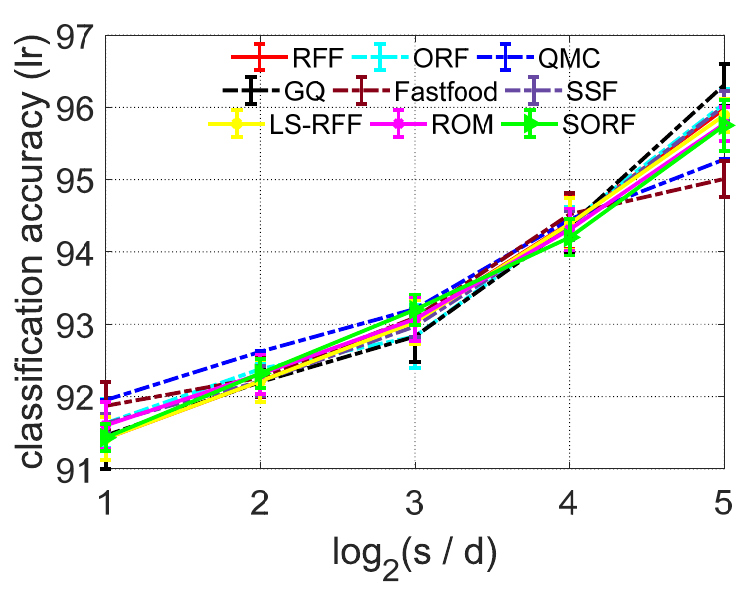}}
	\subfigure{
		\includegraphics[width=0.235\textwidth]{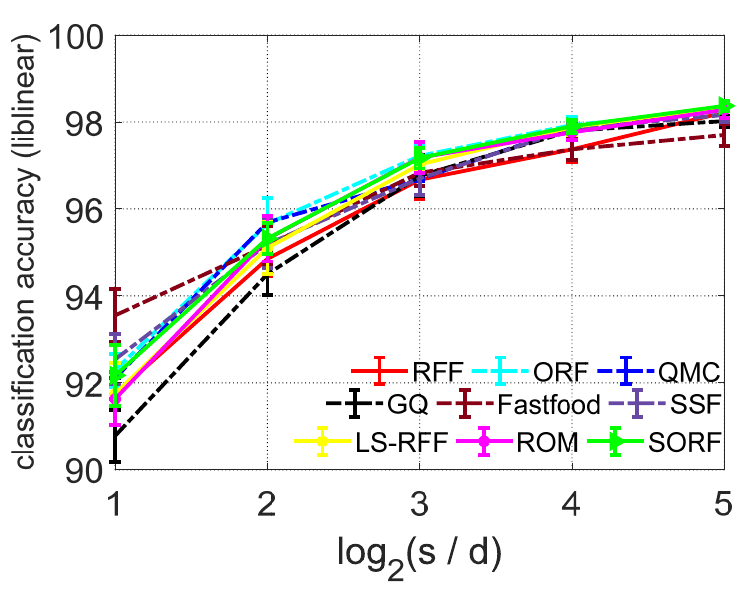}}
	
	\emph{(b) ijcnn1}
	
	\subfigure{
		\includegraphics[width=0.235\textwidth]{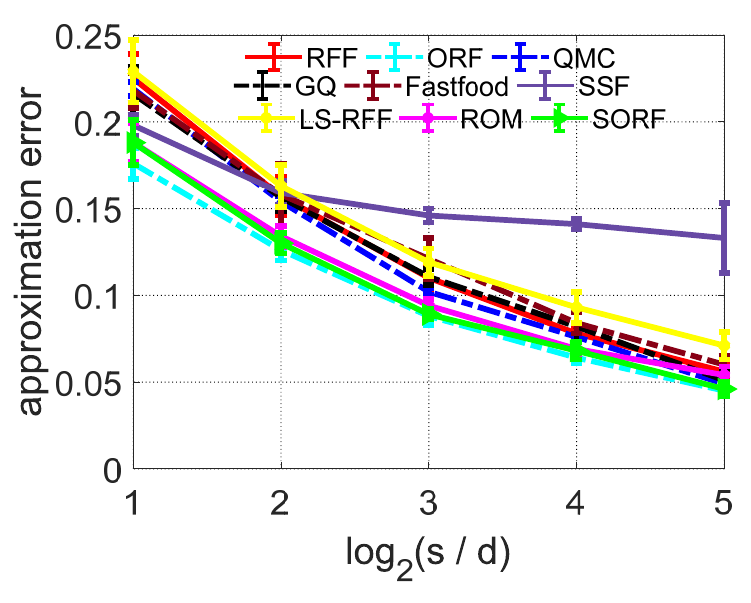}}
	\subfigure{
		\includegraphics[width=0.235\textwidth]{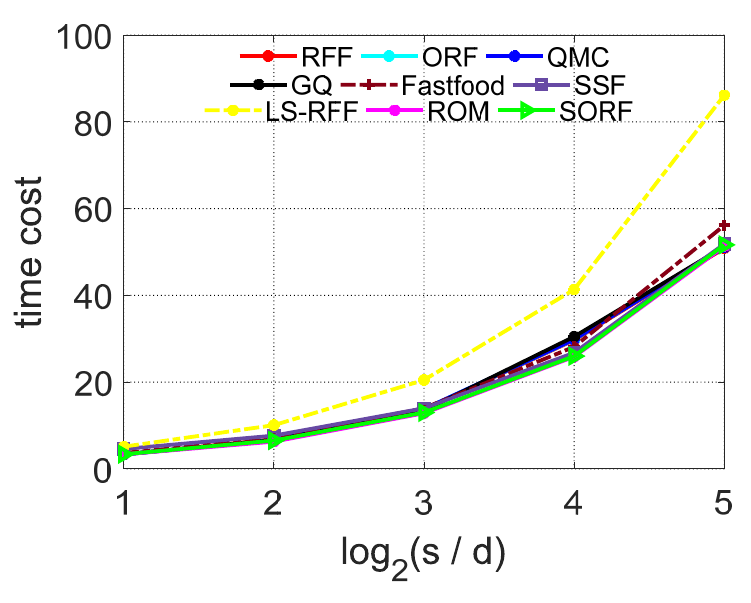}}
	\subfigure{
		\includegraphics[width=0.235\textwidth]{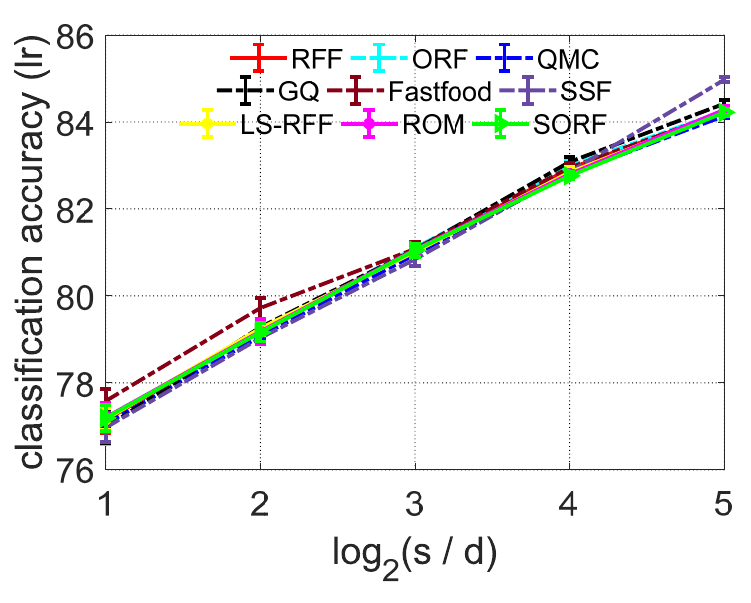}}
	\subfigure{
		\includegraphics[width=0.235\textwidth]{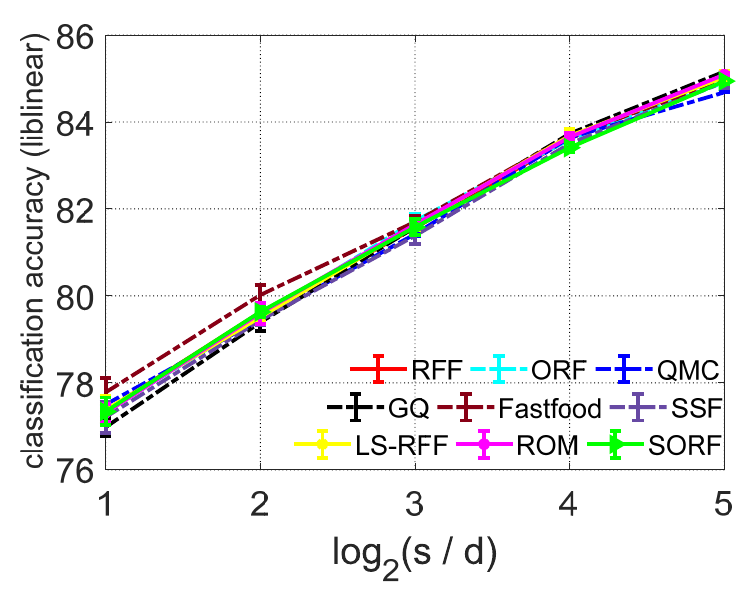}}
	
	\emph{(c) covtype}
	
	\subfigure{
		\includegraphics[width=0.235\textwidth]{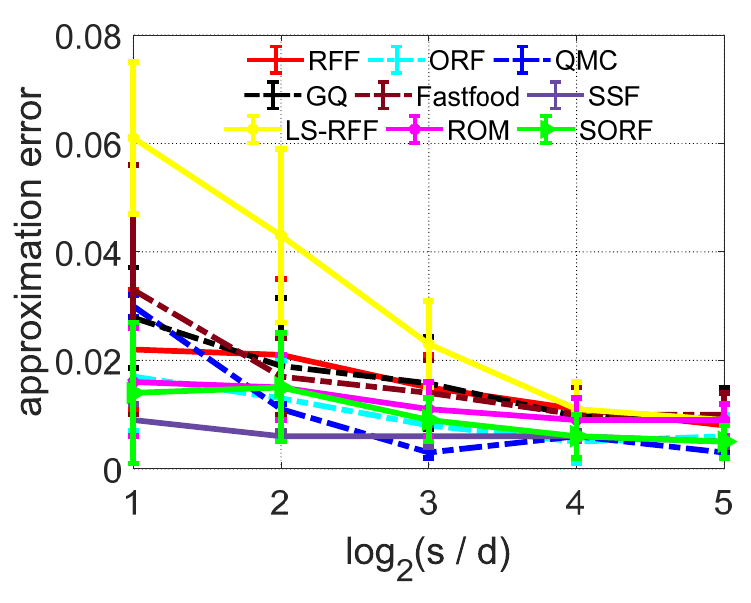}}
	\subfigure{
		\includegraphics[width=0.235\textwidth]{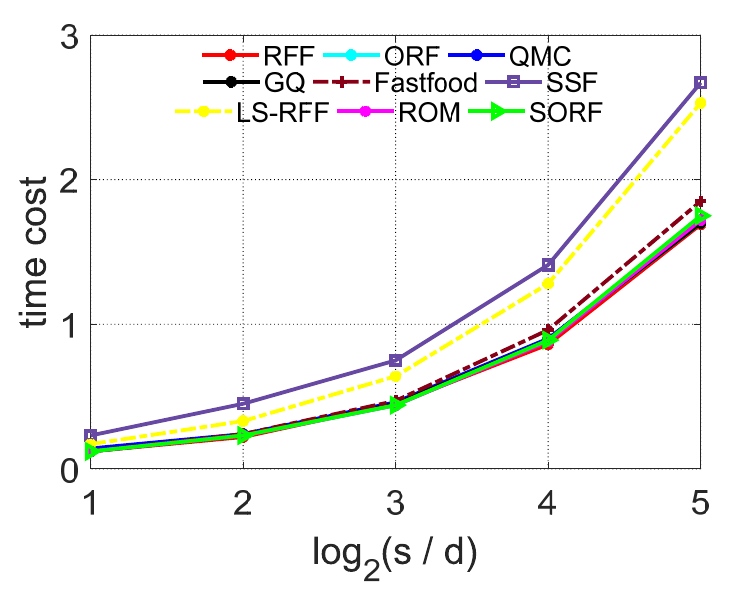}}
	\subfigure{
		\includegraphics[width=0.235\textwidth]{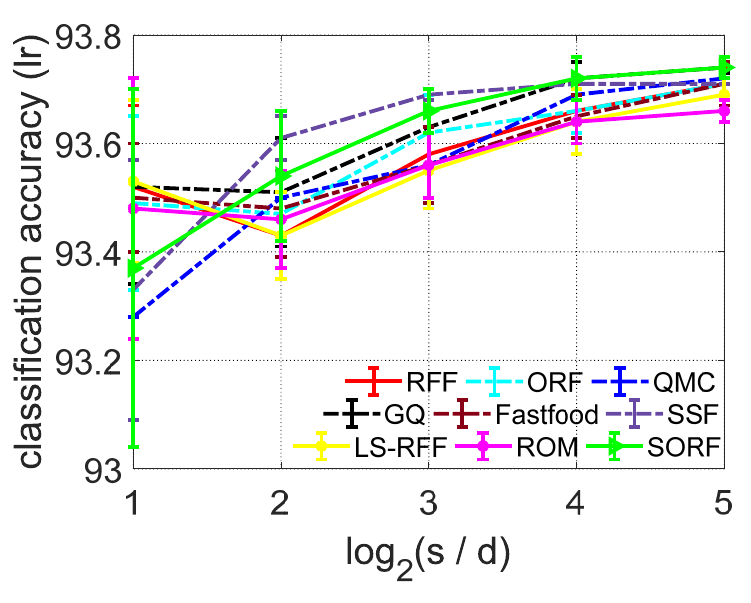}}
	\subfigure{
		\includegraphics[width=0.235\textwidth]{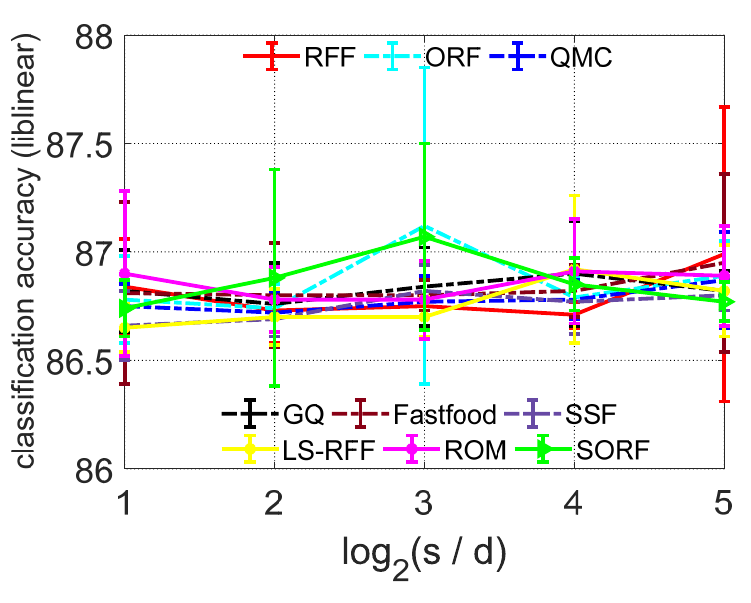}}
	
	\emph{(d) cod-RNA}

	\caption{Results of various algorithms across the Gaussian kernel on the \emph{letter}, \emph{ijcnn1}, \emph{covtype}, \emph{cod-RNA} datasets.}	\label{figgaussuci1}
	\vspace{-0.05cm}
\end{figure*}

\begin{figure*}[!htb]
	\centering
	\subfigure{
		\includegraphics[width=0.235\textwidth]{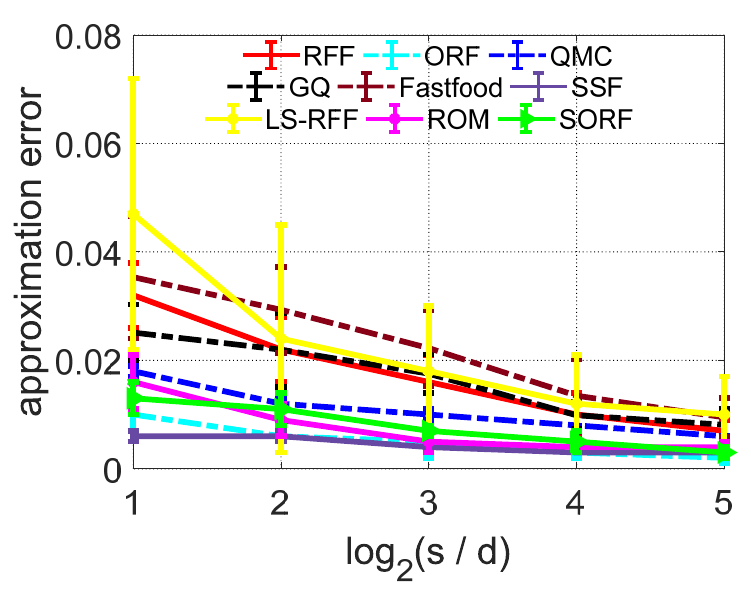}}
	\subfigure{
		\includegraphics[width=0.235\textwidth]{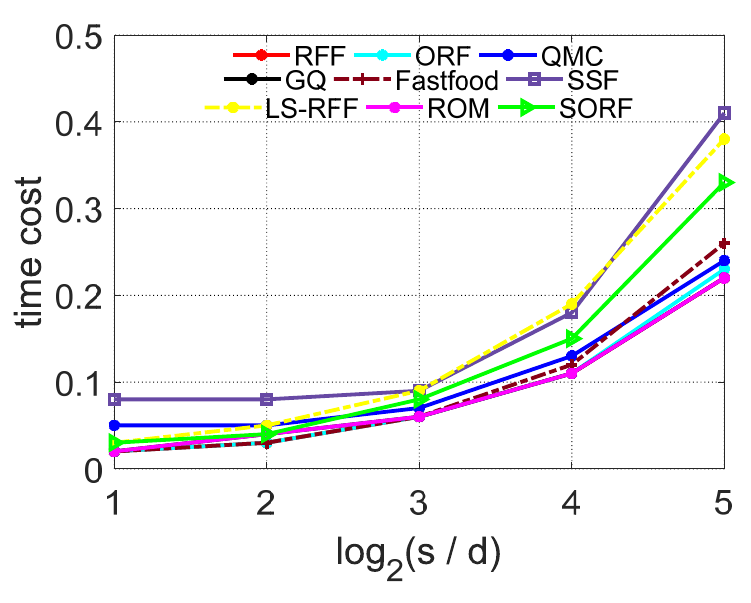}}
	\subfigure{
		\includegraphics[width=0.235\textwidth]{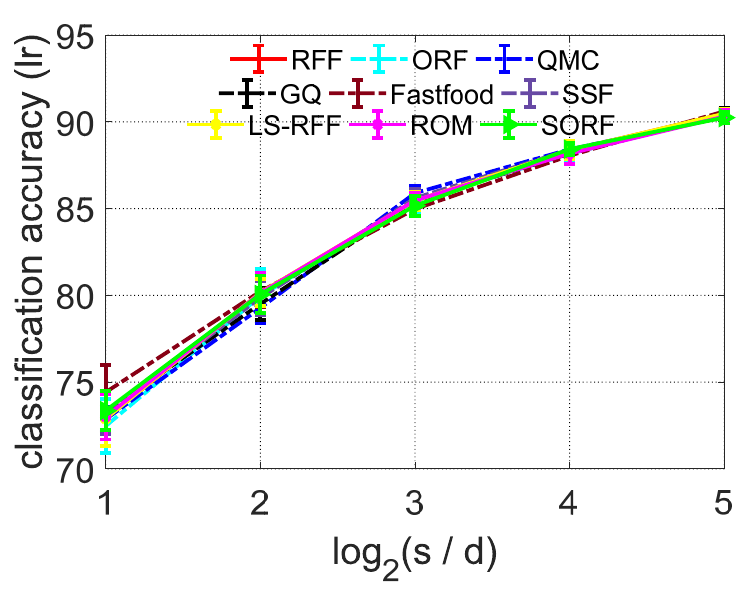}}
	\subfigure{
		\includegraphics[width=0.235\textwidth]{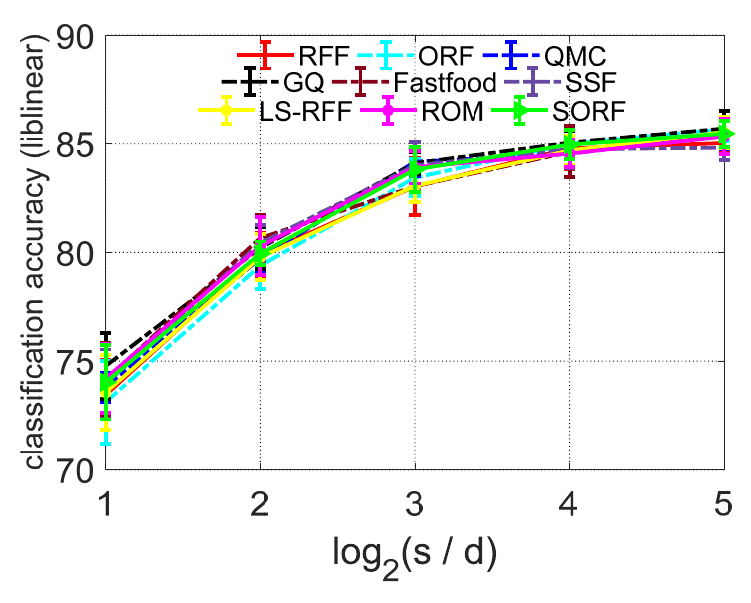}}
	
	\emph{(a) EEG}
	
	\subfigure{
		\includegraphics[width=0.235\textwidth]{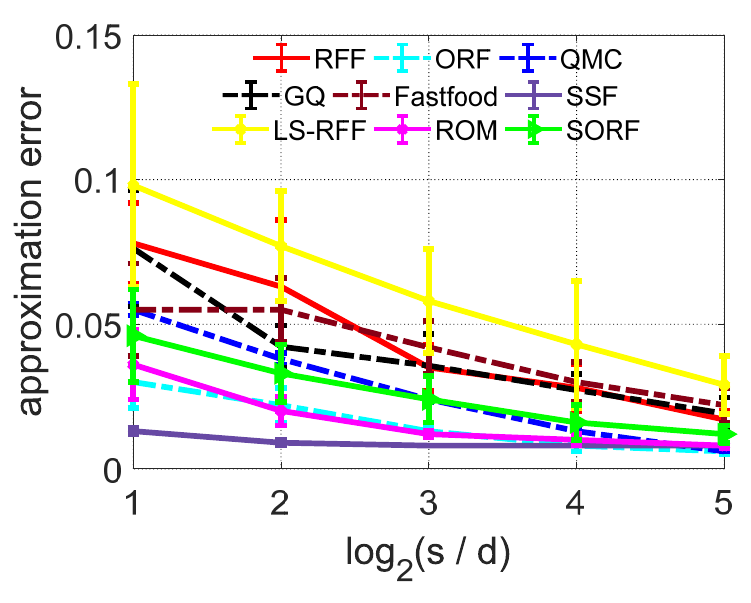}}
	\subfigure{
		\includegraphics[width=0.235\textwidth]{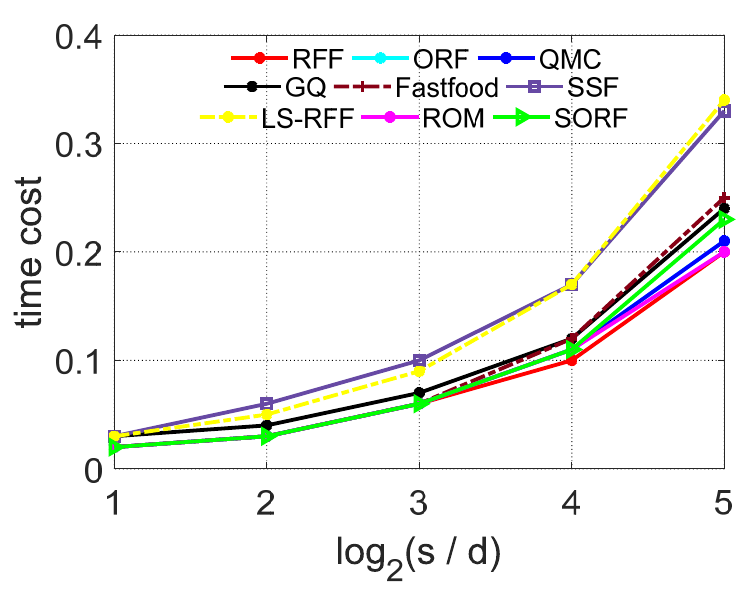}}
	\subfigure{
		\includegraphics[width=0.235\textwidth]{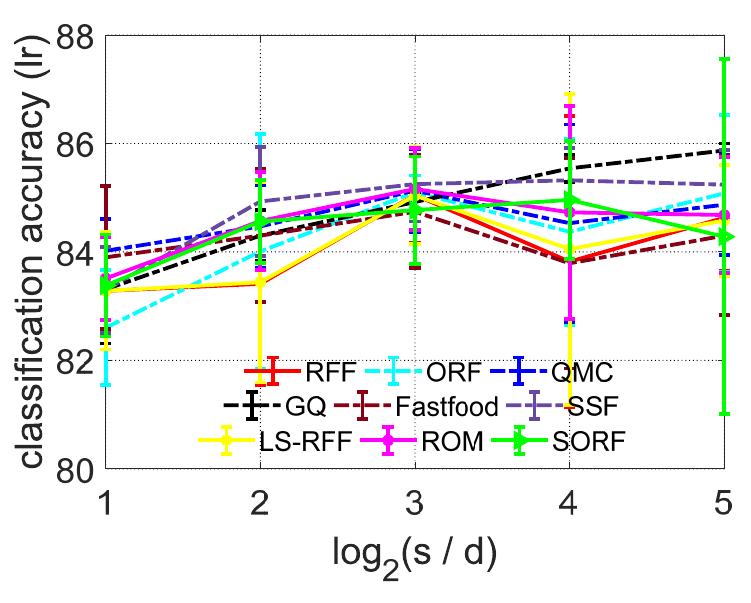}}
	\subfigure{
		\includegraphics[width=0.235\textwidth]{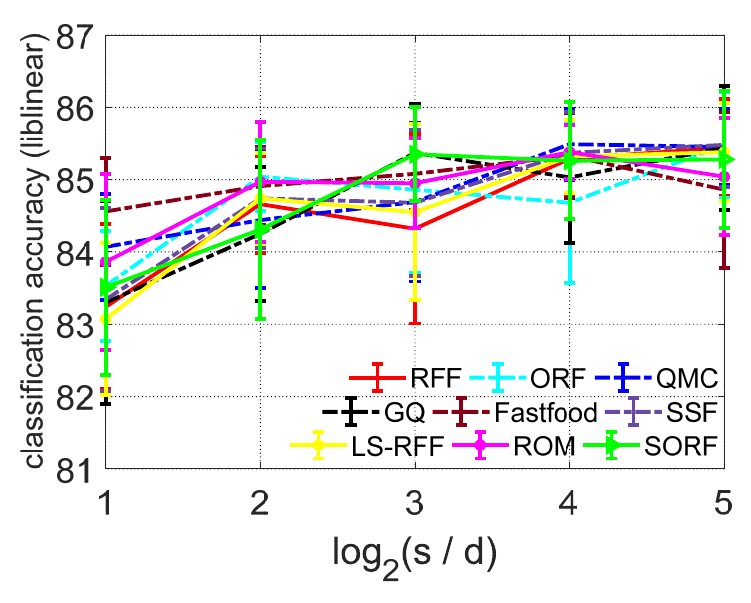}}
	
	\emph{(b) magic04}
	
	\subfigure{
		\includegraphics[width=0.235\textwidth]{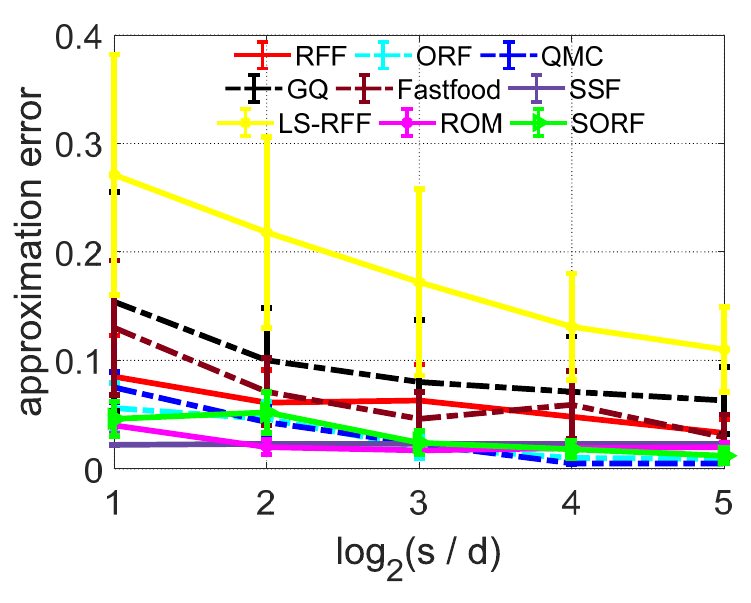}}
	\subfigure{
		\includegraphics[width=0.235\textwidth]{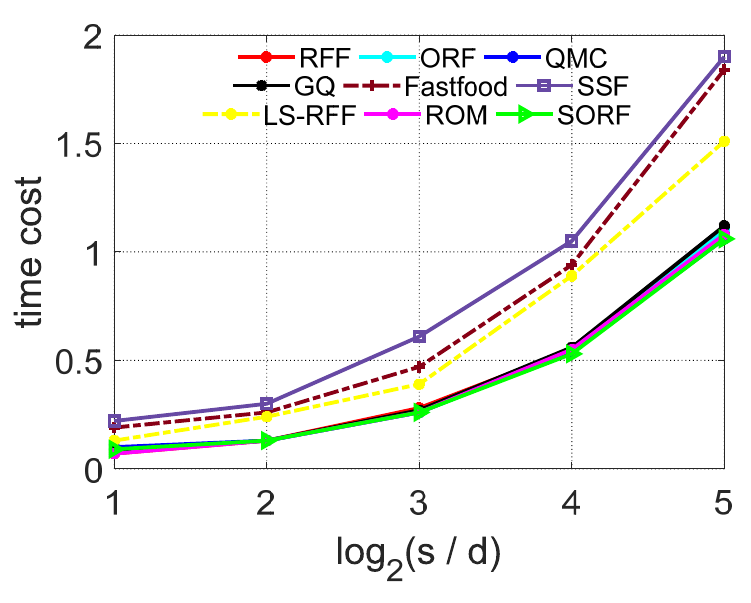}}
	\subfigure{
		\includegraphics[width=0.235\textwidth]{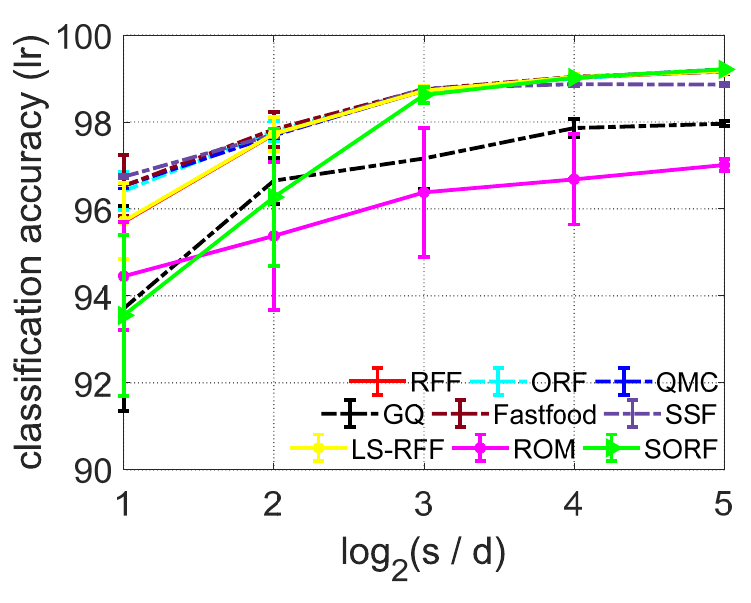}}
	\subfigure{
		\includegraphics[width=0.235\textwidth]{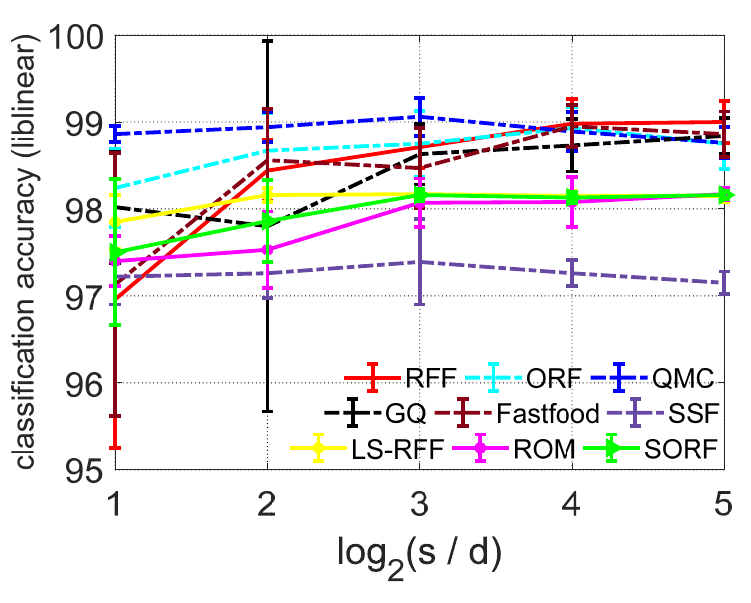}}
	
	\emph{(c) skin}
	
	\subfigure{
		\includegraphics[width=0.235\textwidth]{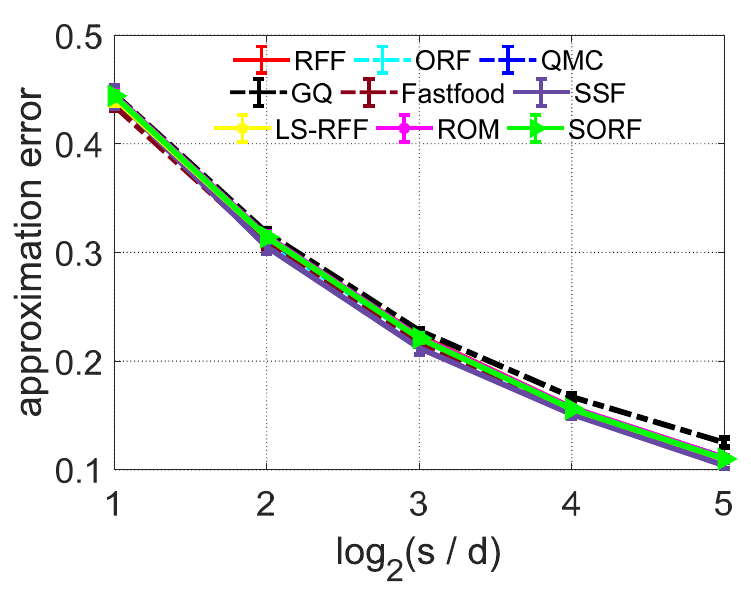}}
	\subfigure{
		\includegraphics[width=0.235\textwidth]{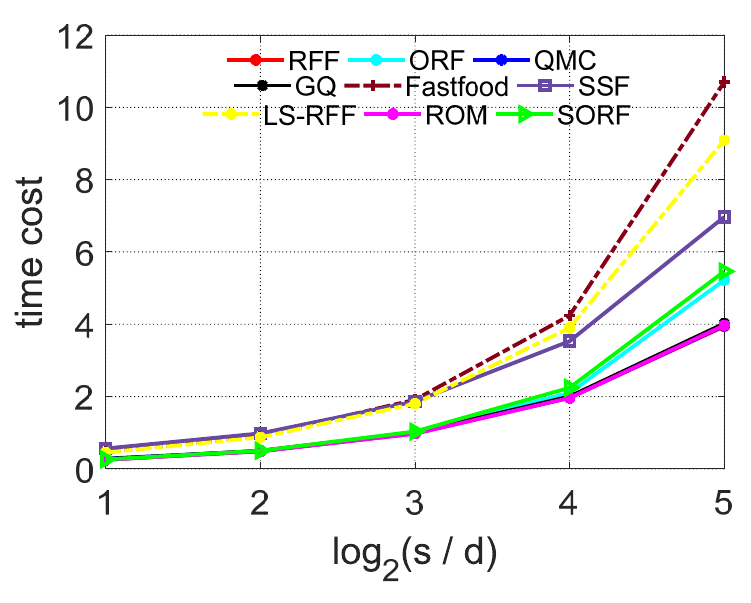}}
	\subfigure{
		\includegraphics[width=0.235\textwidth]{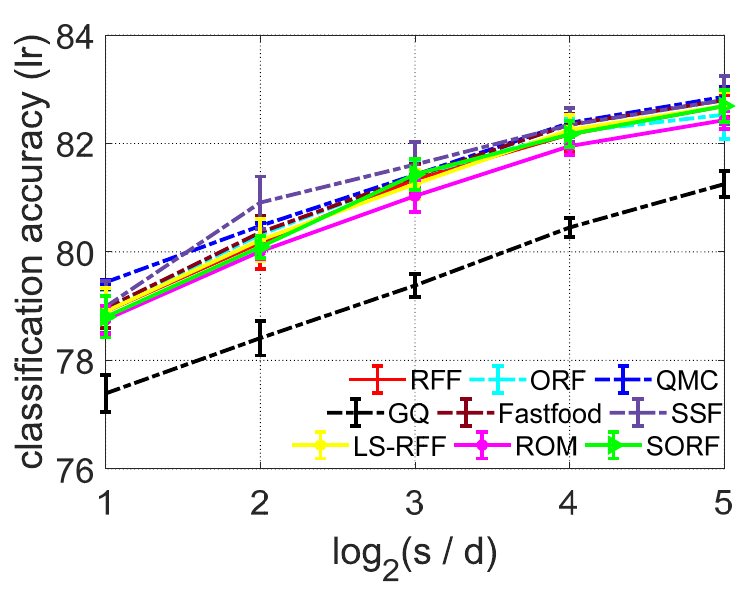}}
	\subfigure{
		\includegraphics[width=0.235\textwidth]{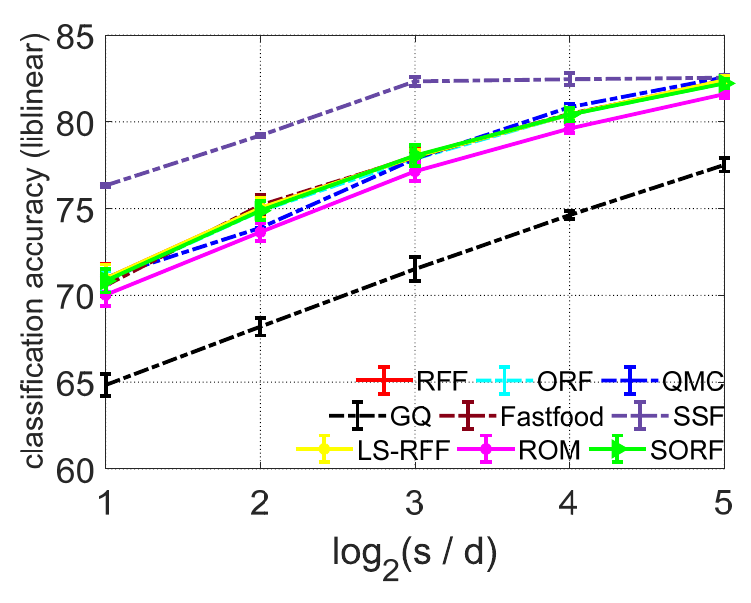}}
	
	\emph{(d) a8a}

	\caption{Results of various algorithms across the Gaussian kernel on the \emph{EEG}, \emph{magic04}, \emph{skin}, \emph{a8a} datasets.}	\label{figgaussuci2}
	\vspace{-0.05cm}
\end{figure*}

\subsection{Results on Gaussian kernels}
\label{app:expgauss}
Figures~\ref{figgaussuci1}, \ref{figgaussuci2} show the approximation error for the Gaussian kernel, the time cost (sec.) of generating randomized feature mappings, and the test accuracy yielded by linear regression and liblinear on the eight datasets, respectively.
We see that as the number of random features increases, these algorithms achieve a smaller approximation error and a higher classification accuracy for both classifiers.
We notice some interesting phenomena in terms of the relation between approximation quality and prediction performance, depending on whether the feature dimension is low  (i.e., $s=2d$ or $s=4d$) or high  (i.e., $s=16d$ or $s= 32d$).
In particular, the algorithms with the best kernel approximation performance are often different in the low-dimensional case and the high dimensional case. Therefore, no algorithm always dominate the others. 
On the other hand, while the approximation quality of these algorithms varies, their prediction performance are often similar. 
Further, to better understand the above observations, we summarize the best performing algorithm on each dataset in terms of the approximation quality and classification accuracy in Table~\ref{tabresstat}, as illustrated in our main text (refer to Section~\ref{sec:expgauss}). 

Regarding to computational efficiency, most algorithms achieve the similar time cost on generating random features except SSF and LS-RFF.
SSF requires constructing the transformation matrix by minimizing the discrete Riesz 0-energy in advance; LS-RFF is a data-dependent algorithm that needs to calculate the approximated ridge leverage score.
Nevertheless, Fastfood/SORF/ROM does not achieve the reduction on time cost, which appears contradictory to the underlying theoretical result on time complexity. 
This might be because, one hand, the feature dimension of the used datasets in our experiments often ranges from 10 to 100, except for the image datasets. In this case, it appears difficult to observe the computational saving from $\mathcal{O}(sd)$ to $\mathcal{O}(s\log d)$ or $\mathcal{O}(d \log d)$.
On the other hand, in our experiments, due to the relatively inefficient Matlab implementation of Fast Discrete Walsh-Hadamard Transform, typical algorithms (e.g., Fastfood/SORF/ROM) do not show a significant reduction on computational efficiency than RFF.

\subsection{Results on Arc-cosine kernels}
\label{app:exparc}
As mentioned before, according to Eq.~\eqref{kernelfor}, various algorithms based on different sampling strategies can be still applicable to arc-cosine kernels, e.g., ORF, QMC, and Fastfood.
Accordingly, eight representative algorithms are taken into comparison on arc-cosine kernels, including RFF, ORF, SORF, ROM, Fastfood, QMC, SSF, and GQ.

\begin{figure*}[t]
	\centering
	\subfigure{
		\includegraphics[width=0.21\textwidth]{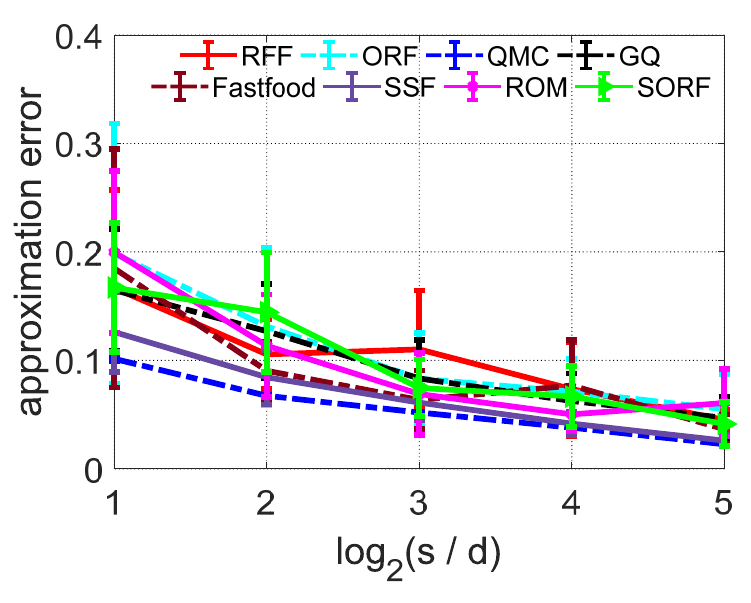}}
	\subfigure{
		\includegraphics[width=0.21\textwidth]{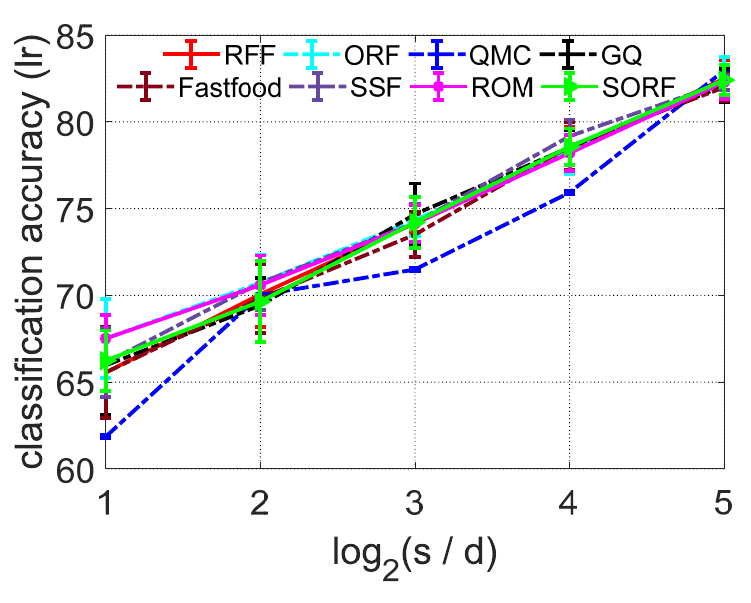}}
	\subfigure{
		\includegraphics[width=0.21\textwidth]{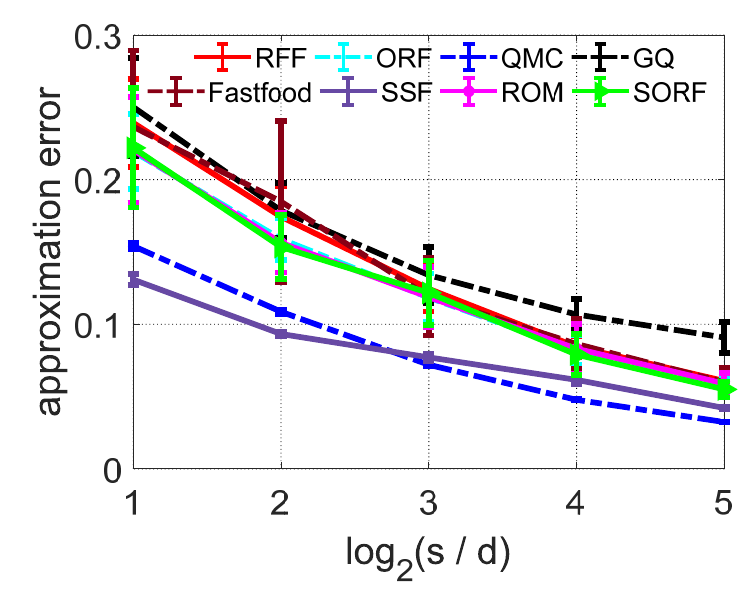}}
	\subfigure{
		\includegraphics[width=0.21\textwidth]{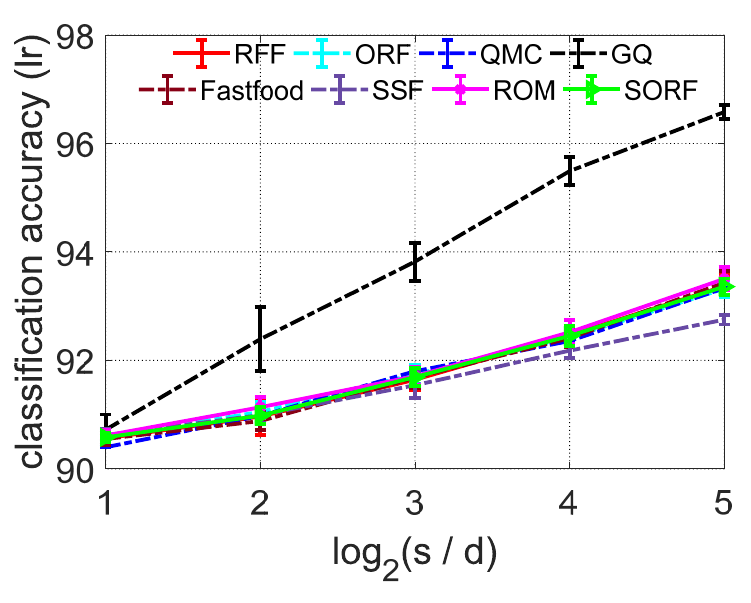}}\\
	\emph{(a) letter} \hspace{6cm} \emph{(b) ijcnn1}
	
	\subfigure{
		\includegraphics[width=0.21\textwidth]{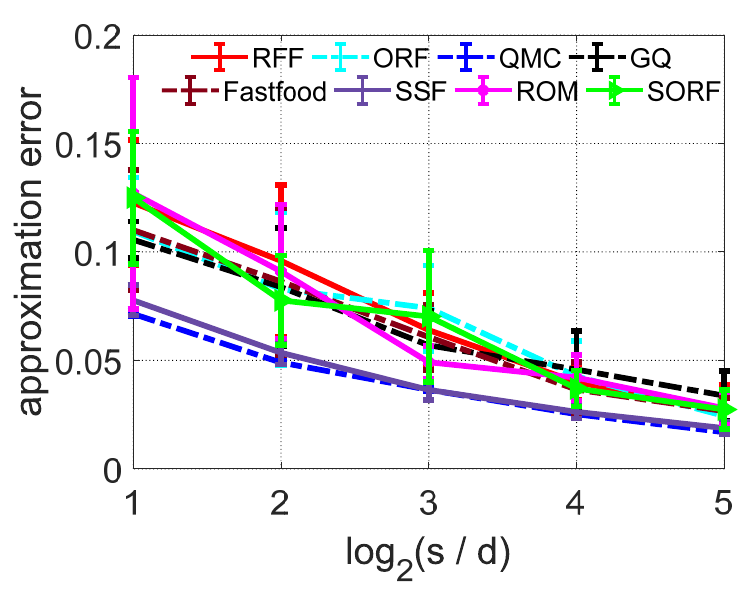}}
	\subfigure{
		\includegraphics[width=0.21\textwidth]{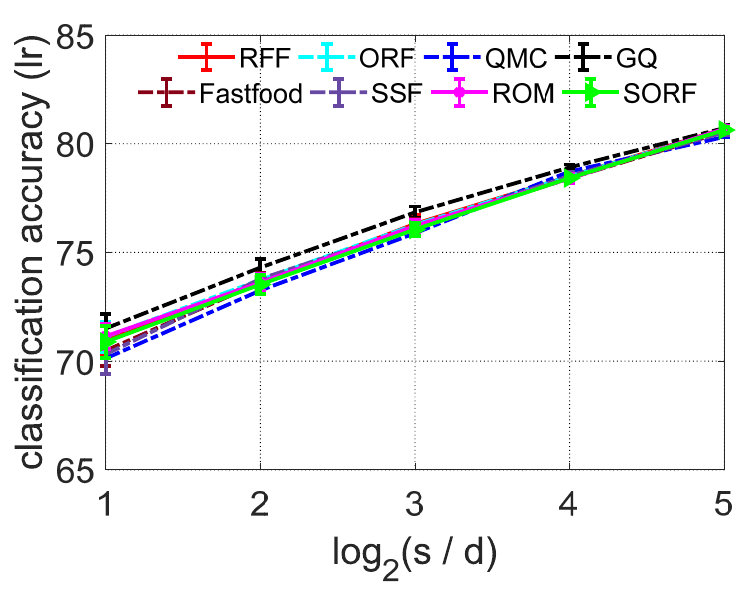}}
	\subfigure{
		\includegraphics[width=0.21\textwidth]{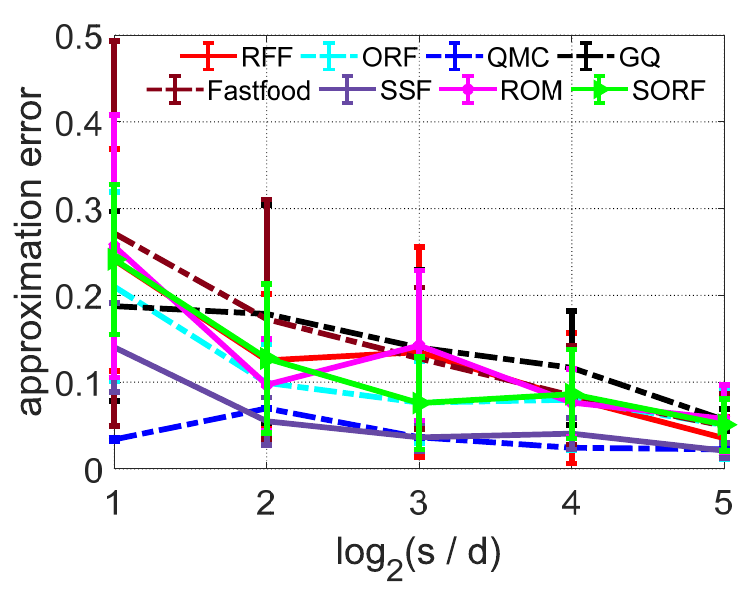}}
	\subfigure{
		\includegraphics[width=0.21\textwidth]{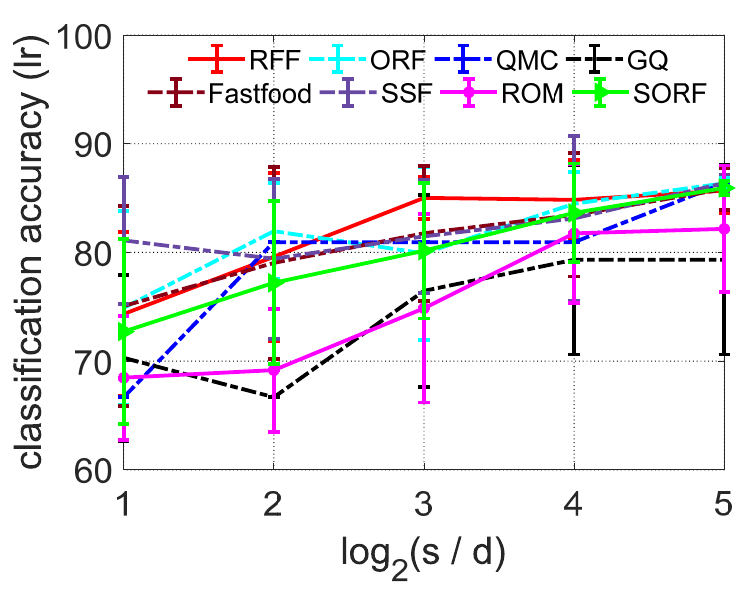}}\\
	\emph{(c) covtype} \hspace{6cm} \emph{(d) cod-RNA}
	
	\subfigure{
		\includegraphics[width=0.21\textwidth]{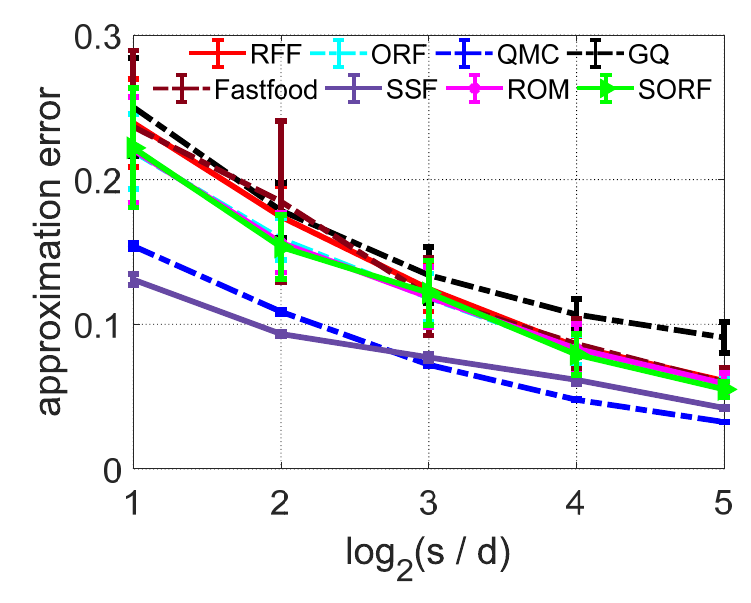}}
	\subfigure{
		\includegraphics[width=0.21\textwidth]{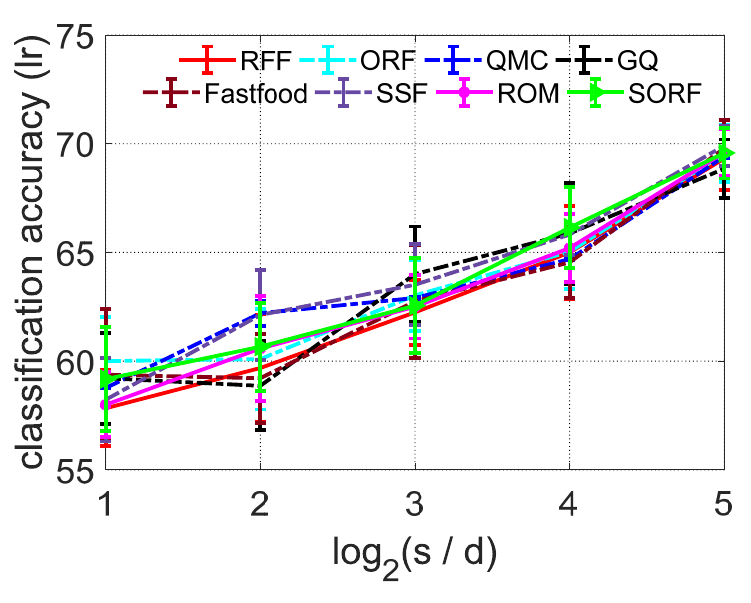}}
	\subfigure{
		\includegraphics[width=0.21\textwidth]{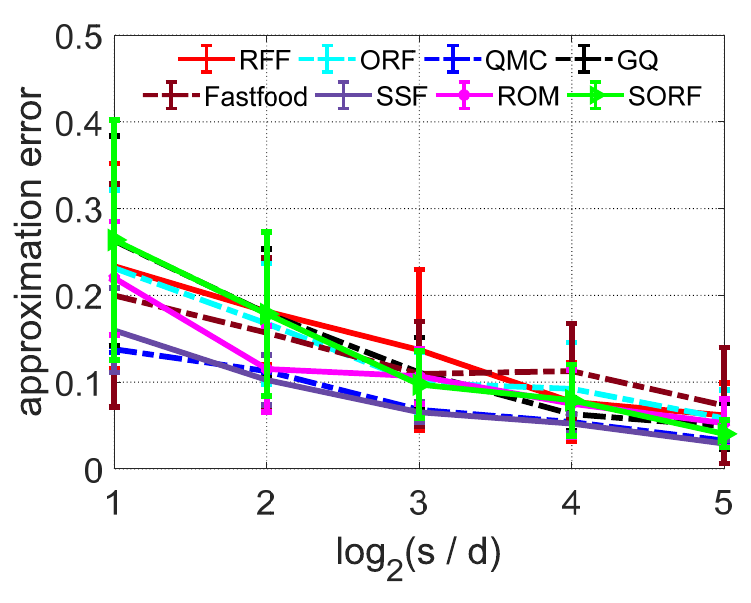}}
	\subfigure{
		\includegraphics[width=0.21\textwidth]{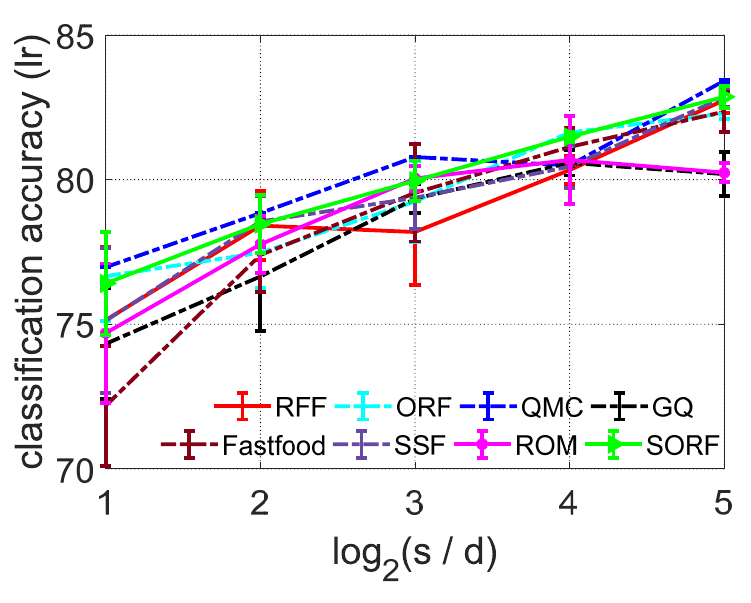}}\\
	\emph{(e) EEG} \hspace{6cm} \emph{(f) magic04}
	
	\subfigure{
		\includegraphics[width=0.21\textwidth]{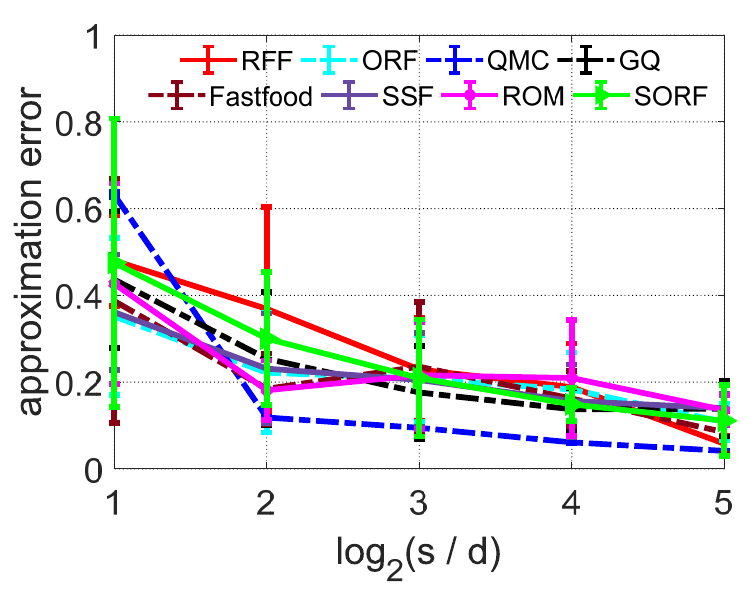}}
	\subfigure{
		\includegraphics[width=0.21\textwidth]{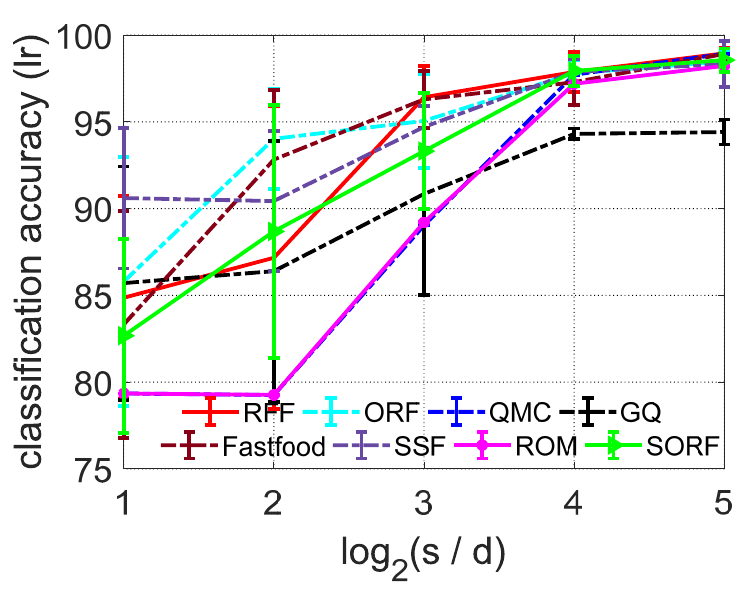}}
	\subfigure{
		\includegraphics[width=0.215\textwidth]{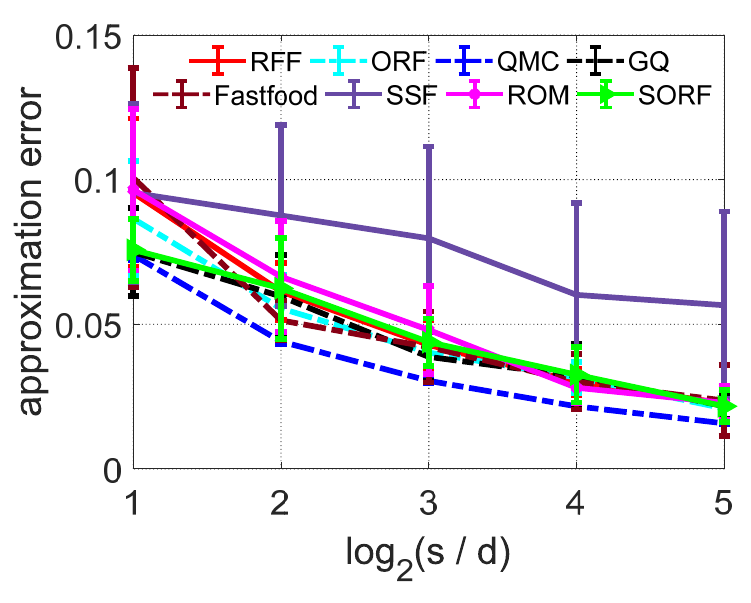}}
	\subfigure{
		\includegraphics[width=0.21\textwidth]{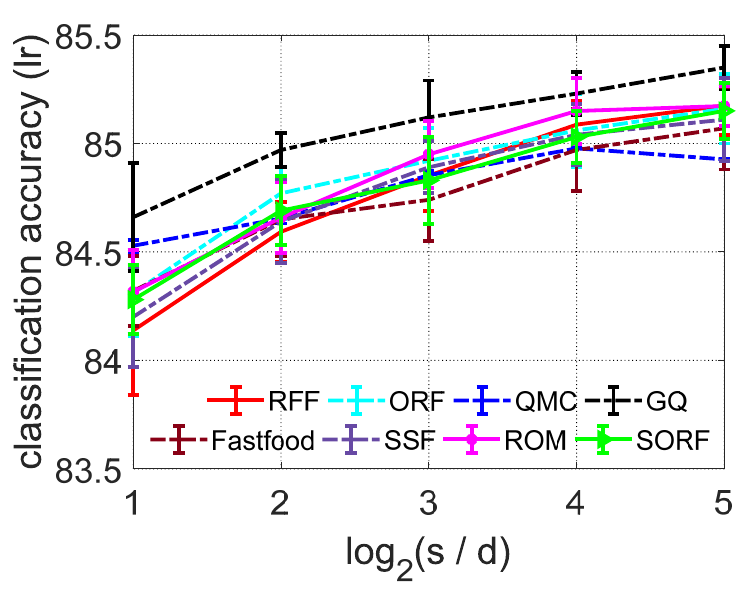}}\\
	\emph{(g) skin} \hspace{6cm} \emph{(h) a8a}

	\caption{Results on eight datasets across the zero-order arc-cosine kernel.}	\label{figarccos0}
	\vspace{-0.05cm}
\end{figure*}

\begin{figure*}[t]
	\centering
	\subfigure{
		\includegraphics[width=0.21\textwidth]{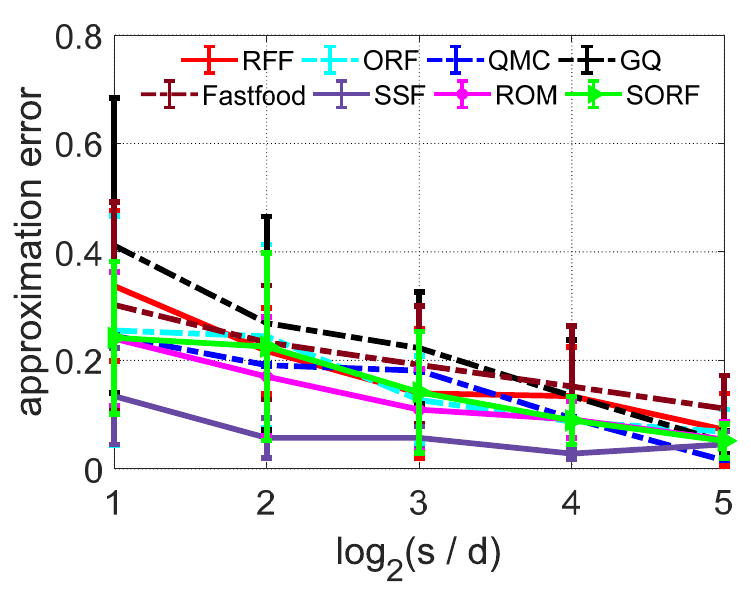}}
	\subfigure{
		\includegraphics[width=0.21\textwidth]{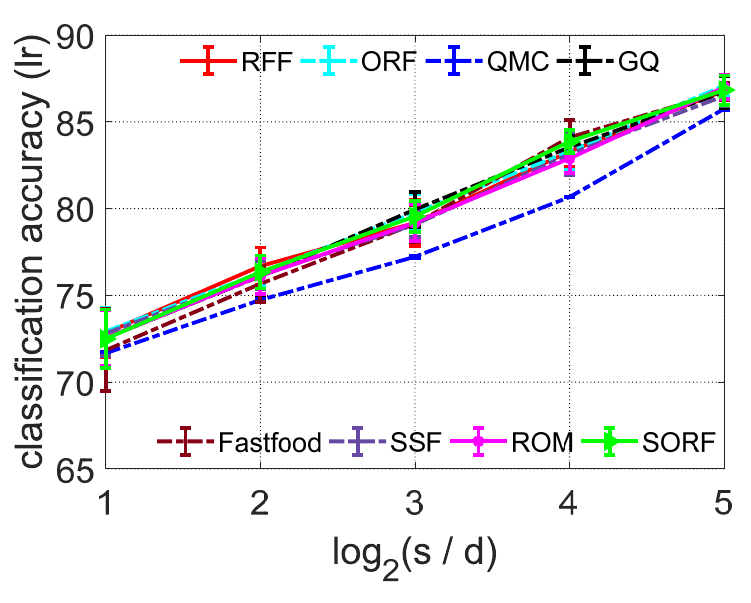}}
	\subfigure{
		\includegraphics[width=0.21\textwidth]{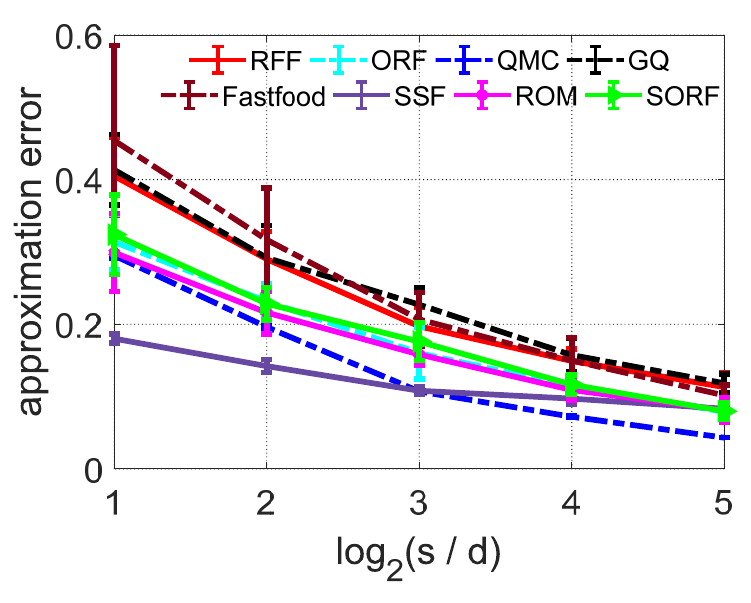}}
	\subfigure{
		\includegraphics[width=0.21\textwidth]{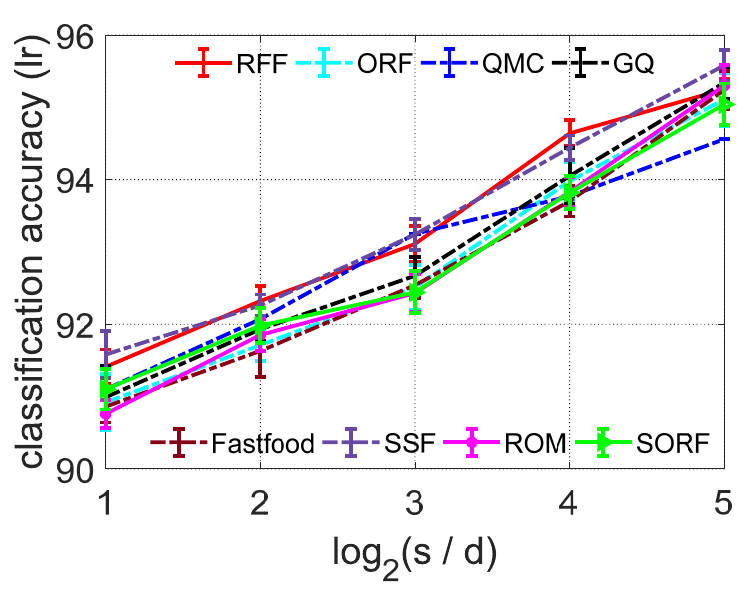}}\\
	\emph{(a) letter} \hspace{6cm} \emph{(b) ijcnn1}
	
	\subfigure{
		\includegraphics[width=0.21\textwidth]{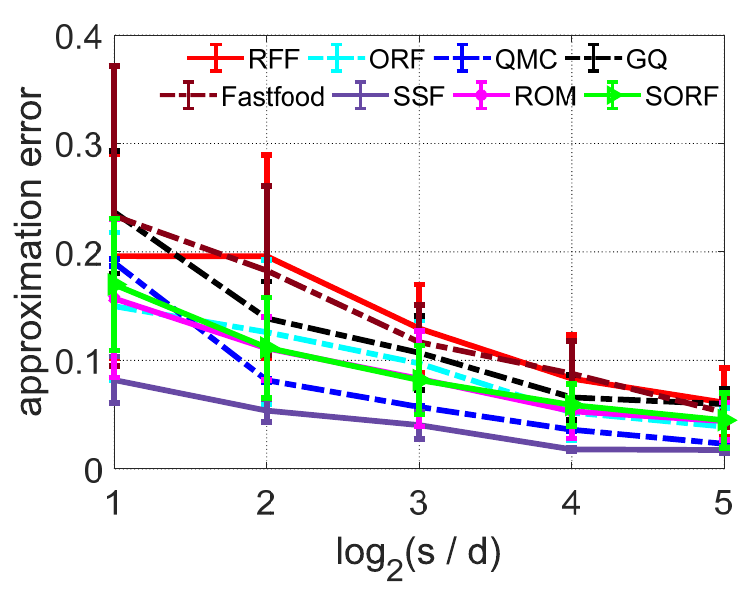}}
	\subfigure{
		\includegraphics[width=0.21\textwidth]{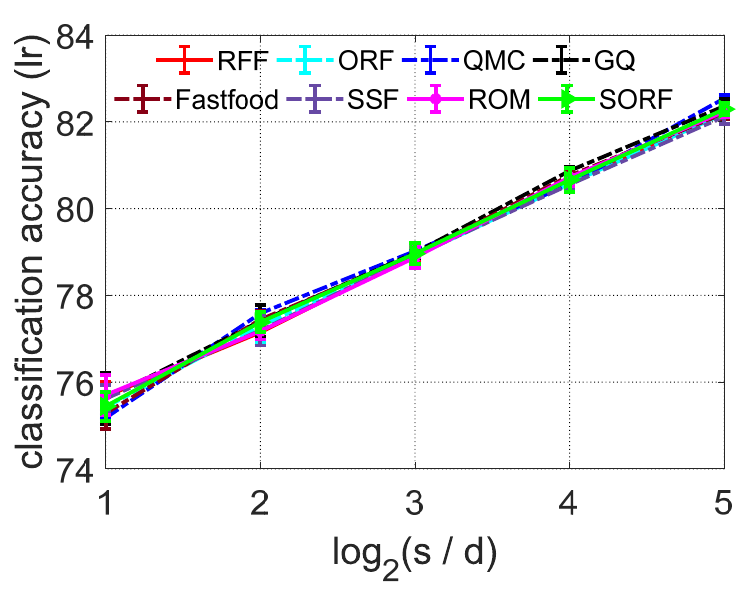}}
	\subfigure{
		\includegraphics[width=0.21\textwidth]{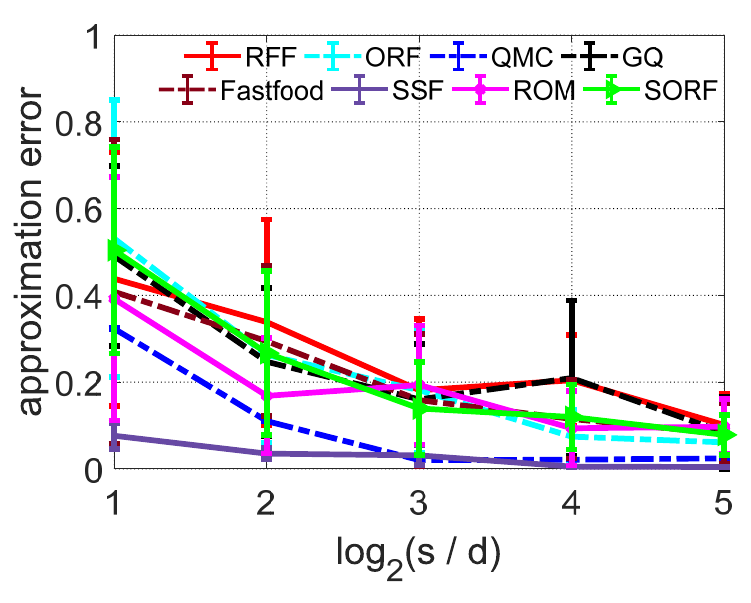}}
	\subfigure{
		\includegraphics[width=0.21\textwidth]{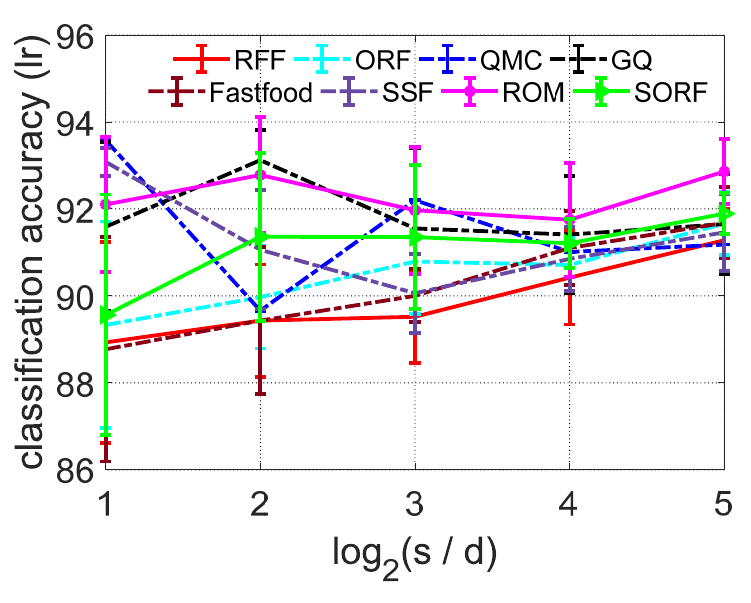}}\\
	\emph{(c) covtype} \hspace{6cm} \emph{(d) cod-RNA}
	
	\subfigure{
		\includegraphics[width=0.21\textwidth]{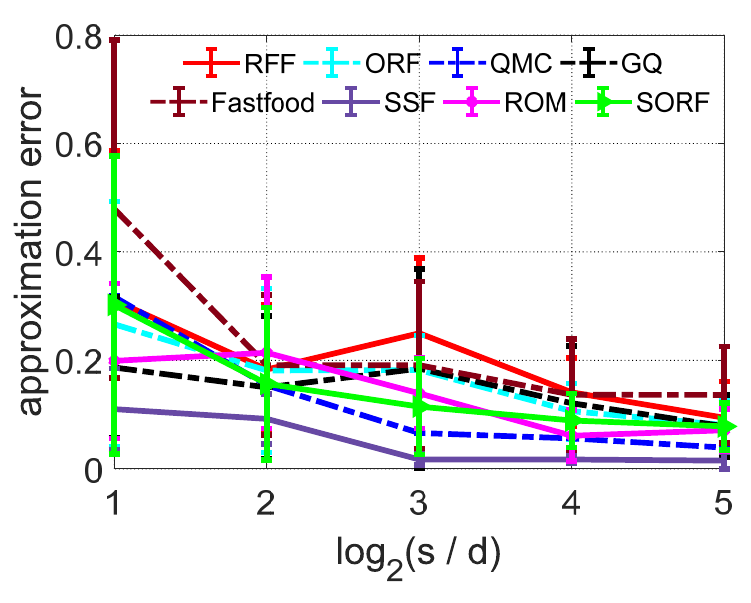}}
	\subfigure{
		\includegraphics[width=0.21\textwidth]{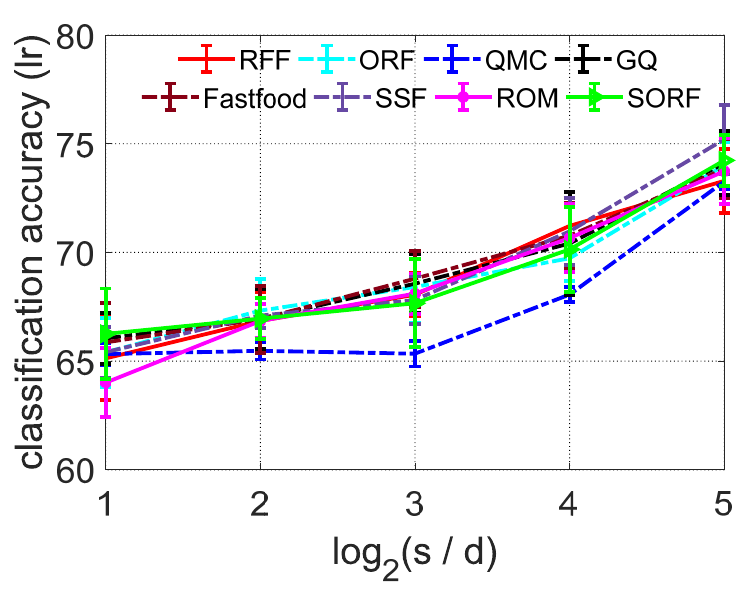}}
	\subfigure{
		\includegraphics[width=0.21\textwidth]{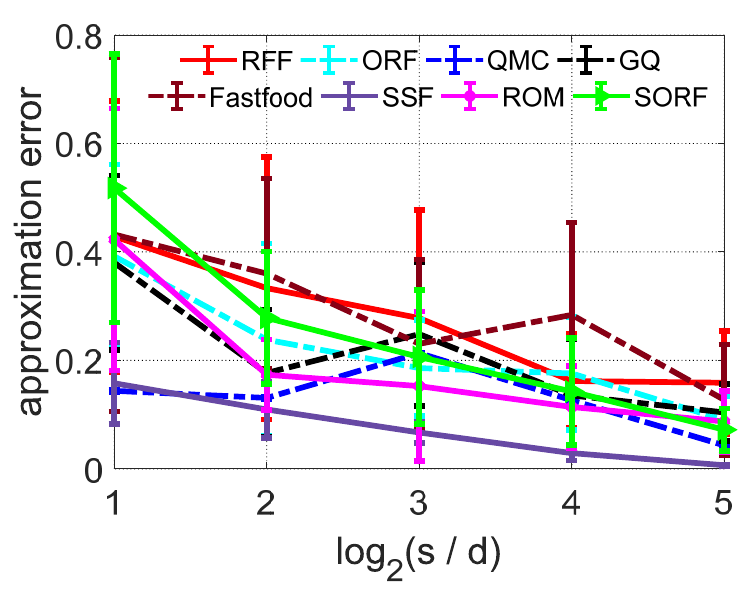}}
	\subfigure{
		\includegraphics[width=0.21\textwidth]{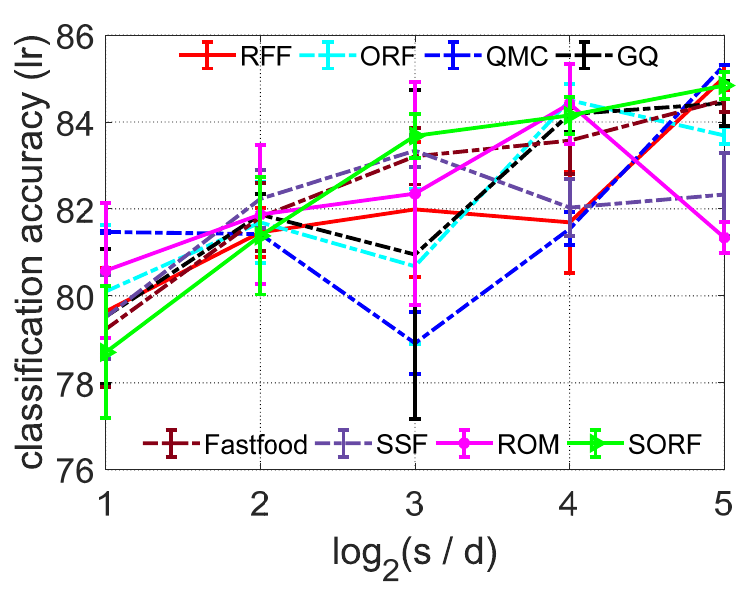}}\\
	\emph{(e) EEG} \hspace{6cm} \emph{(f) magic04}
	
	\subfigure{
		\includegraphics[width=0.21\textwidth]{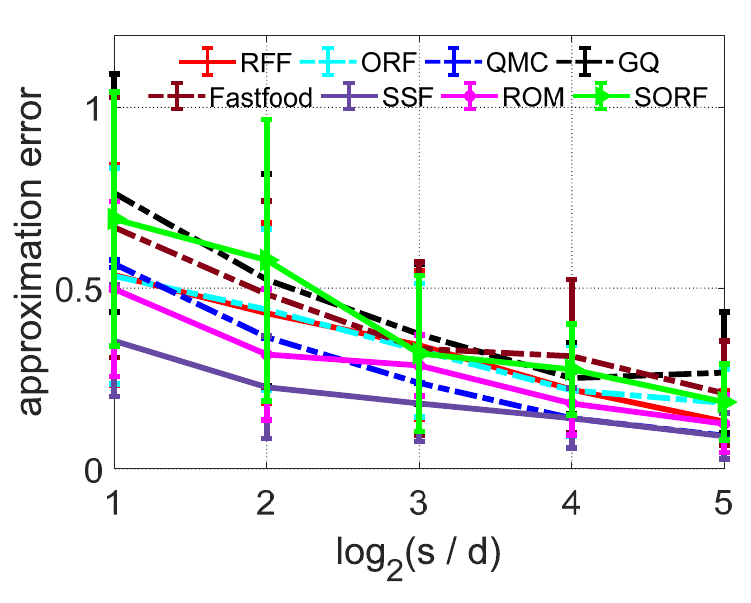}}
	\subfigure{
		\includegraphics[width=0.21\textwidth]{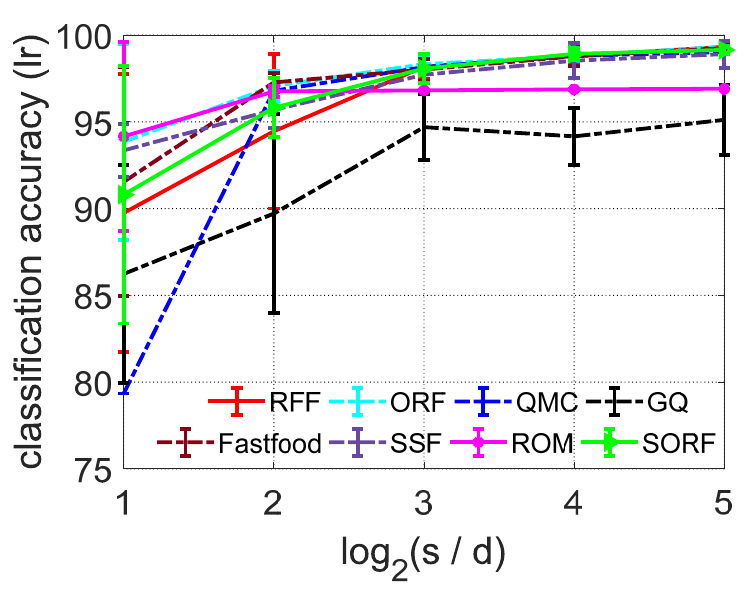}}
	\subfigure{
		\includegraphics[width=0.215\textwidth]{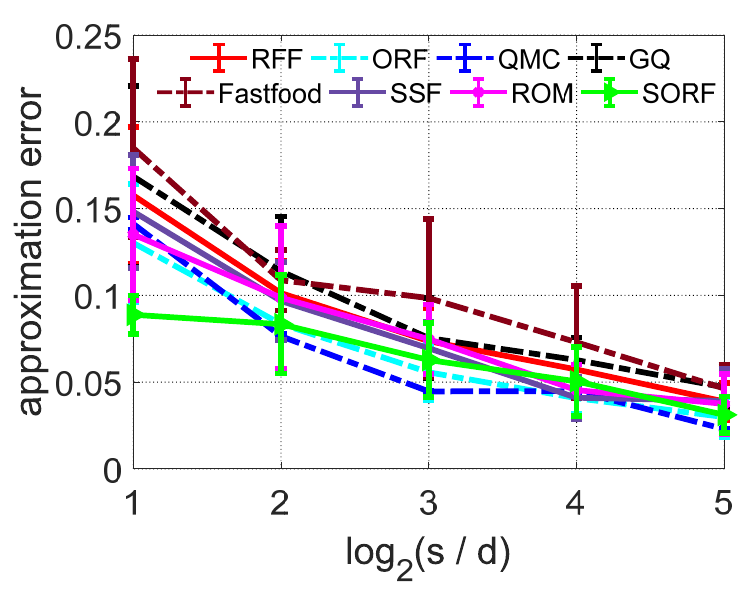}}
	\subfigure{
		\includegraphics[width=0.21\textwidth]{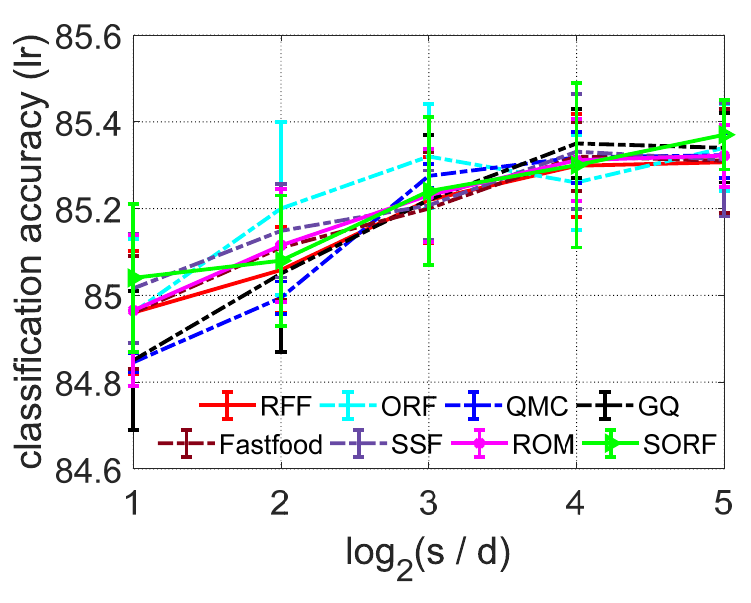}}\\
	\emph{(g) skin} \hspace{6cm} \emph{(h) a8a}

	\caption{Results on eight datasets across the first-order arc-cosine kernel.}	\label{figarccos1}
	\vspace{-0.05cm}
\end{figure*}

Figures~\ref{figarccos0},~\ref{figarccos1} show the approximation error and test accuracy across the zero/first-order arc-cosine kernels, respectively.
It can be observed that in most cases SSF and QMC achieve a lower approximation error than the other approaches, which corresponds to the theoretical findings. However, there is no distinct difference on approximation between RFF and ORF/SORF. In fact, the current theoretical results on ORF/SORF for variance reduction are only valid to the Gaussian kernel. Whether such results can be transferred to arc-cosine kernels are still unclear.
In general, the approximation performance and time cost (see Figure~\ref{figtimearc0} and \ref{figtimearc1}) of these algorithms on arc-cosine kernels are similar to that on the Gaussian kernel, though the approximation error value is often larger than that for the Gaussian kernel.
This is because, according to Eq.~\eqref{kernelfor}, we actually conduct a $d$-dimensional integration approximation, the smoothness of the integrand $\sigma(\bm \omega^{\top} \bm x) \sigma(\bm \omega^{\!\top} \bm x')$ would significantly effect the approximation performance as indicated by sampling theory.
In the term of classification performance, the difference in test accuracy of most algorithms is relatively small, which shows the similar tendency with that of the Gaussian kernel.

\subsection{Results on Polynomial kernels}
\label{app:exppoly}
\begin{figure*}[!htb]
	\centering
	\subfigure{
		\includegraphics[width=0.21\textwidth]{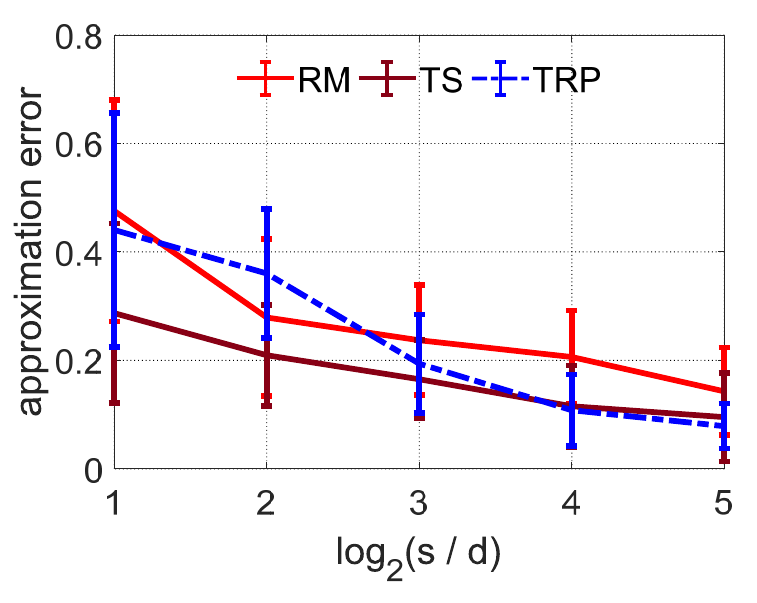}}
	\subfigure{
		\includegraphics[width=0.21\textwidth]{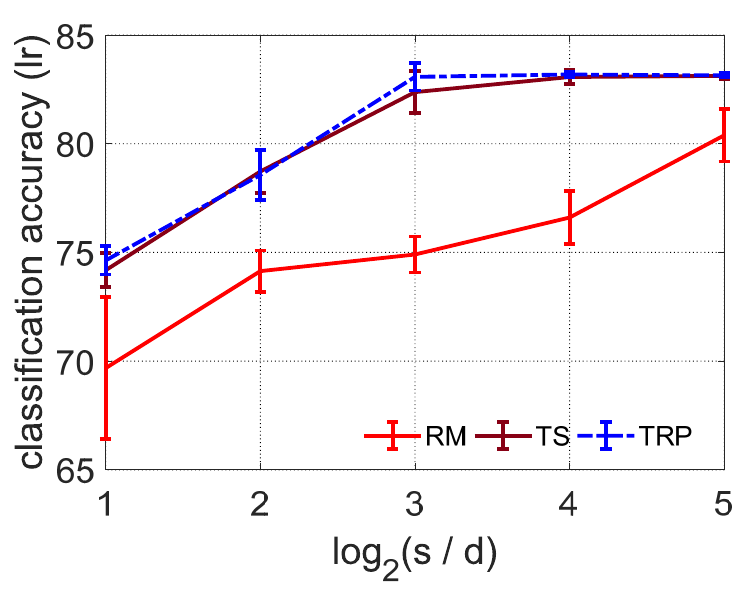}}
	\subfigure{
		\includegraphics[width=0.21\textwidth]{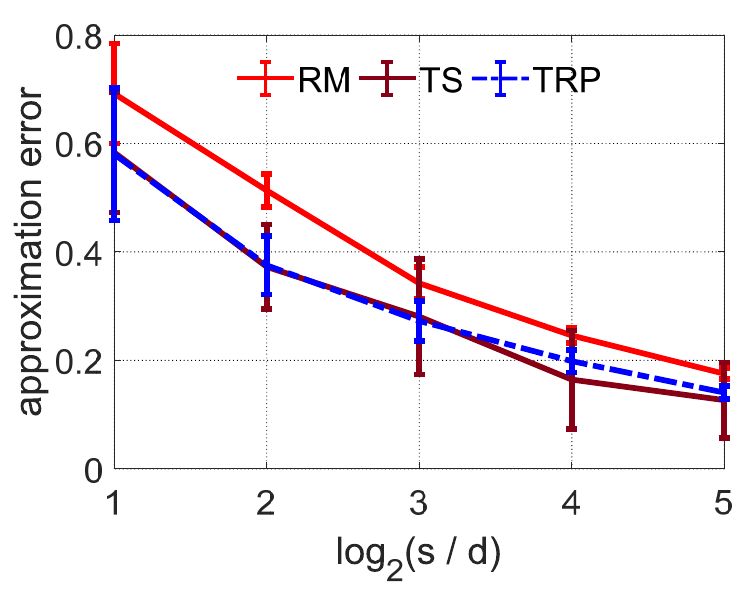}}
	\subfigure{
		\includegraphics[width=0.21\textwidth]{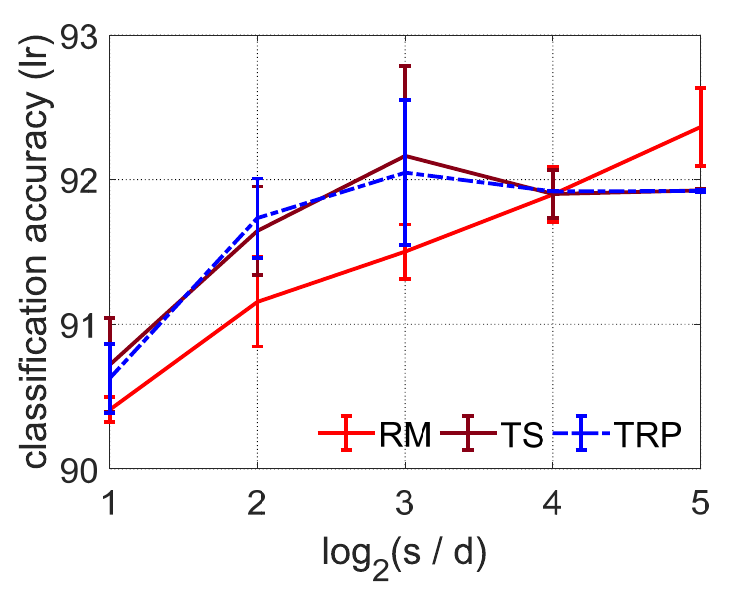}}\\
	\emph{(a) letter} \hspace{6cm} \emph{(b) ijcnn1}
	
		\subfigure{
		\includegraphics[width=0.21\textwidth]{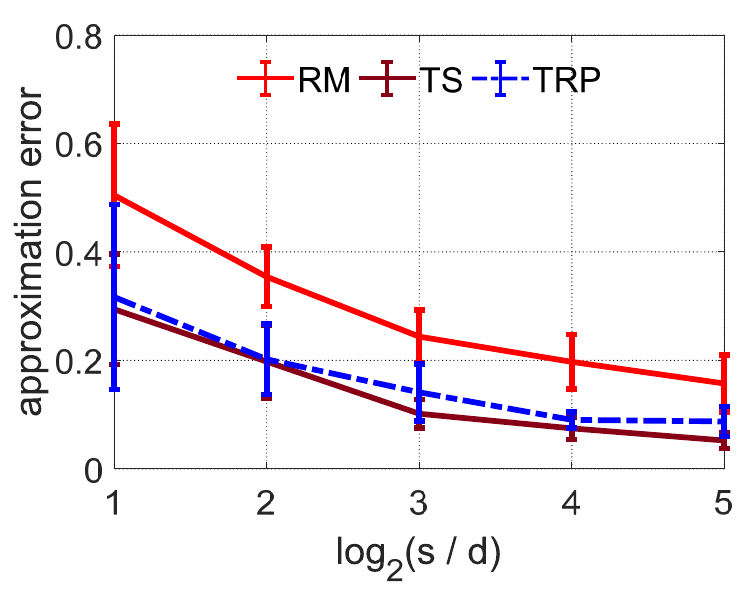}}
	\subfigure{
		\includegraphics[width=0.21\textwidth]{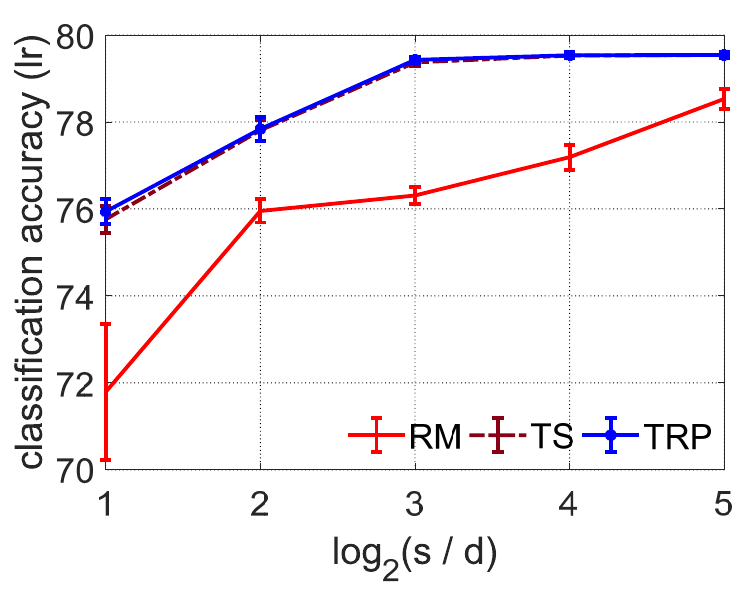}}
	\subfigure{
		\includegraphics[width=0.21\textwidth]{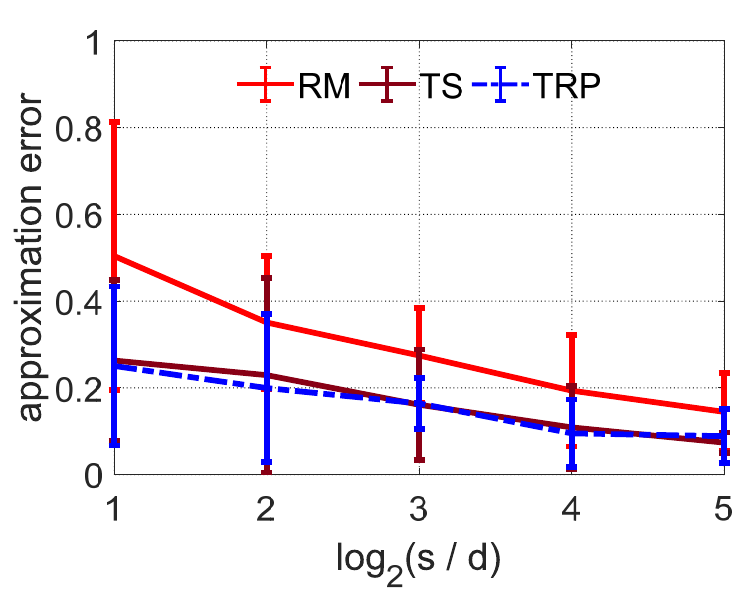}}
	\subfigure{
		\includegraphics[width=0.21\textwidth]{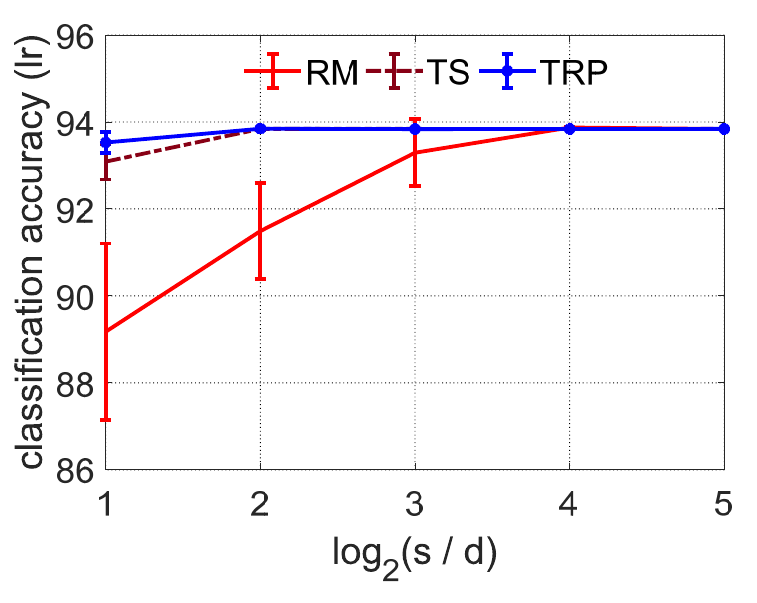}}\\
	\emph{(c) covtype} \hspace{6cm} \emph{(d) cod-RNA}
	
			\subfigure{
		\includegraphics[width=0.21\textwidth]{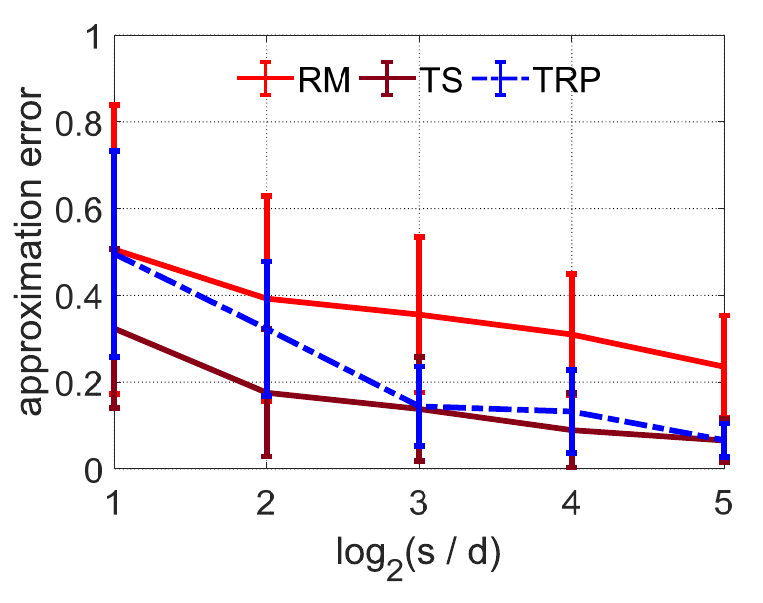}}
	\subfigure{
		\includegraphics[width=0.21\textwidth]{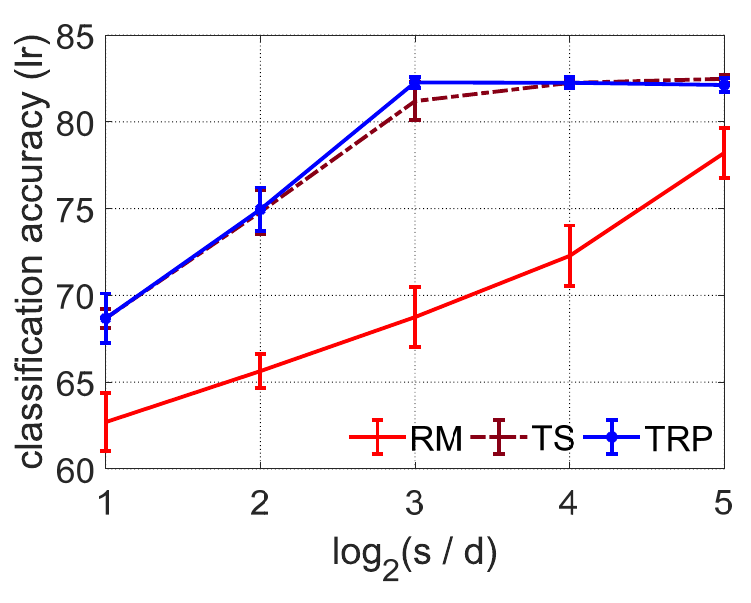}}
	\subfigure{
		\includegraphics[width=0.21\textwidth]{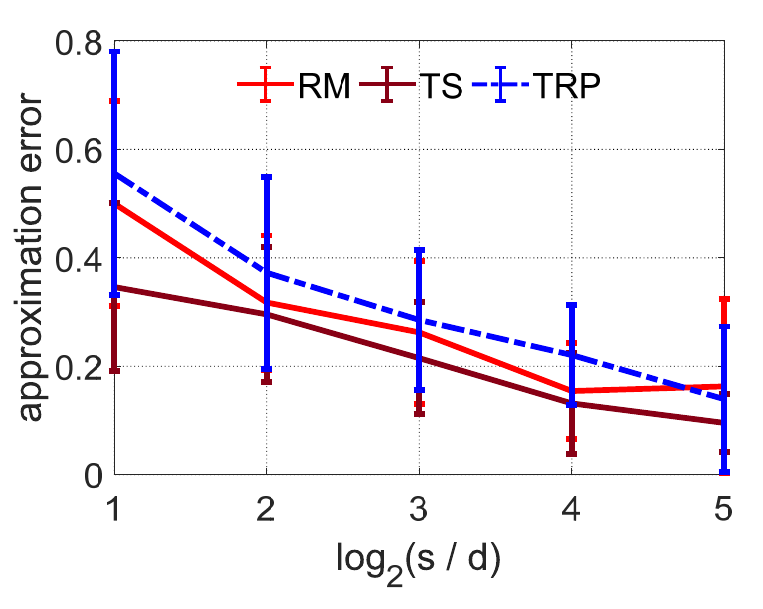}}
	\subfigure{
		\includegraphics[width=0.21\textwidth]{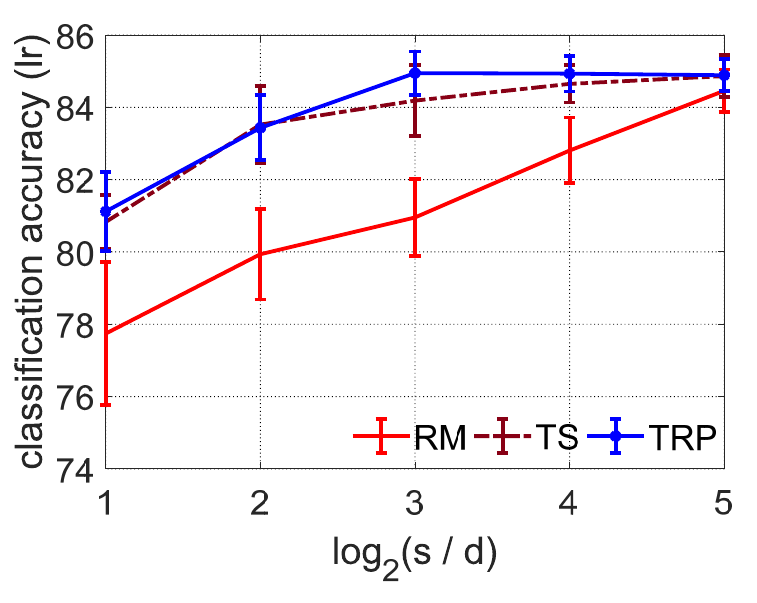}}\\
	\emph{(e) EEG} \hspace{6cm} \emph{(f) magic04}
	
				\subfigure{
		\includegraphics[width=0.21\textwidth]{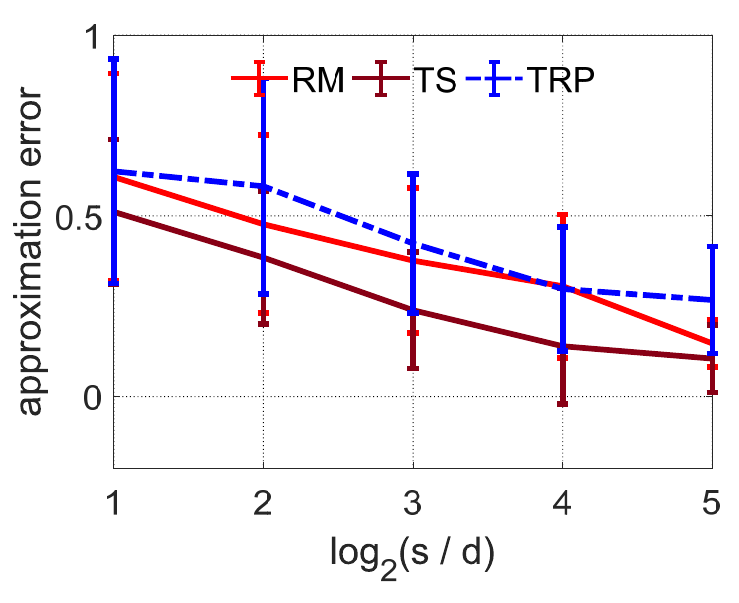}}
	\subfigure{
		\includegraphics[width=0.21\textwidth]{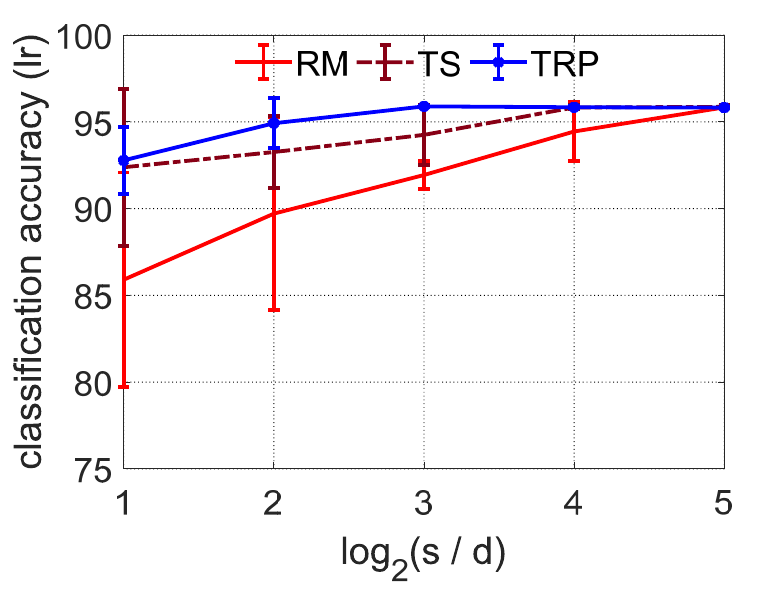}}
	\subfigure{
		\includegraphics[width=0.215\textwidth]{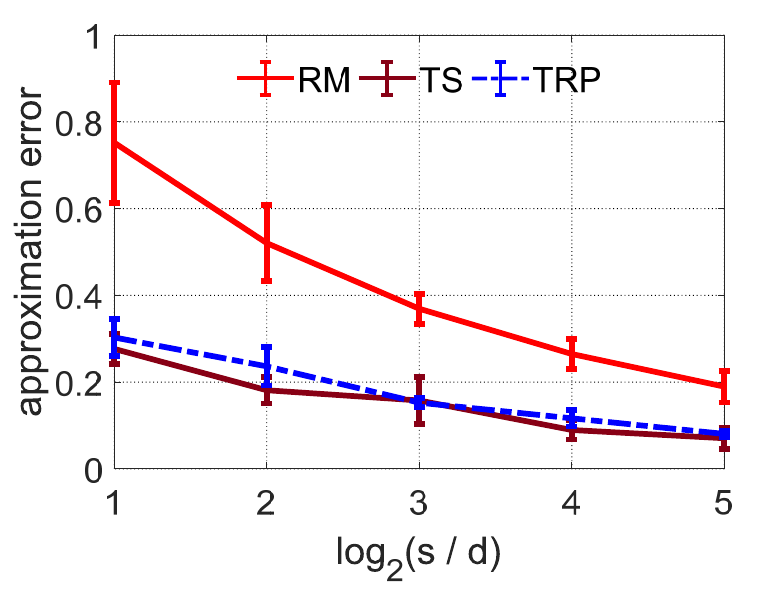}}
	\subfigure{
		\includegraphics[width=0.21\textwidth]{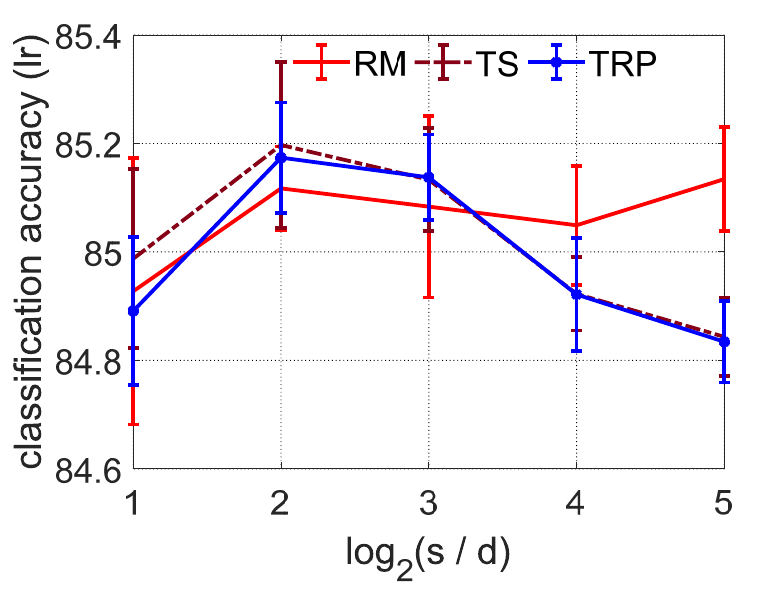}}\\
	\emph{(g) skin} \hspace{6cm} \emph{(h) a8a}

		\caption{Results on eight datasets across polynomial kernels.}	\label{figpoly1}
	\vspace{-0.05cm}
\end{figure*}

\begin{figure*}[t]
	\centering
	\subfigure[zero-order arc-cosine kernel]{\label{figtimearc0}
		\includegraphics[width=0.32\textwidth]{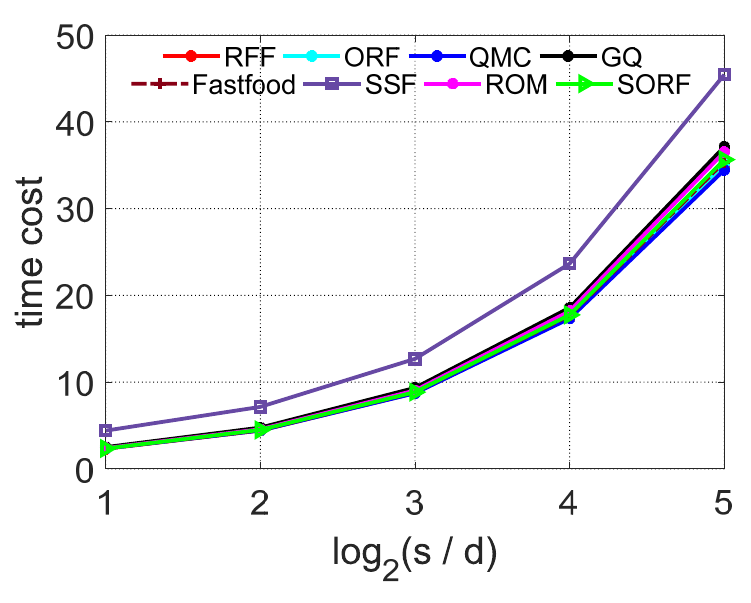}}
	\subfigure[first-order arc-cosine kernel]{\label{figtimearc1}
		\includegraphics[width=0.32\textwidth]{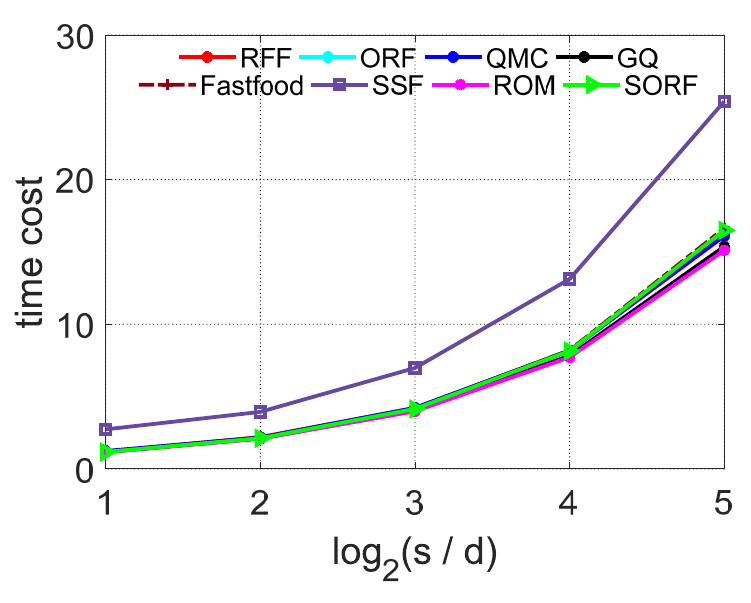}}
	\subfigure[polynomial kernel]{\label{figtimepoly}
		\includegraphics[width=0.32\textwidth]{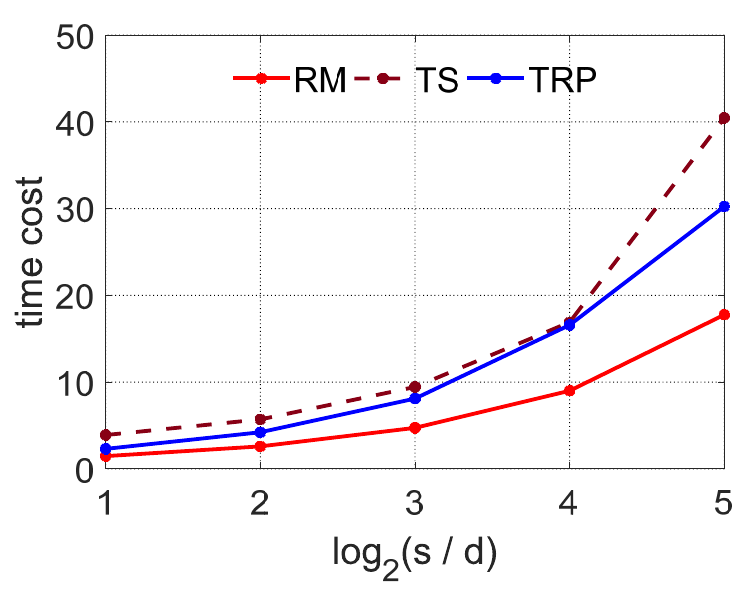}}
	\caption{Comparison of various algorithms on the \emph{covtype} dataset in terms of time cost for generating random features mappings.}	\label{figtimecost}
	\vspace{-0.05cm}
\end{figure*}

For polynomial kernel approximation, we include three representative approaches, tensorized random projections (TRP) \cite{meister2019tight},  TensorSketch (TS) \cite{Pham2013Fast}, and random Maclaurin (RM) \cite{kar2012random} sketch evaluated on eight datasets for approximation and prediction.
Since the polynomial kernel can be written as a special type of tensor product, TS and TRP work in this setting by sketching a tensor product of arbitrary vectors, which is different from RM using Maclaurin expansion.
Figure~\ref{figpoly1} shows that, TS and TRP have the similar test accuracy, but significantly perform better than RM, as RM's generality is not required for the polynomial kernel.
Besides, Figure~\ref{figtimepoly} shows that RM is quite computational efficient due to its Maclaurin expansion scheme; while TS takes much time on generating random features since it utilizes a fixed sampling probability to compute the tensor sketch; while TRP works in a flexible sampling strategy proportional to its Maclaurin coefficient.

\subsection{Results on the \emph{MNIST-8M} dataset}
\label{app:explarge}
Here we evaluate the compared ten algorithms across the Gaussian kernel and arc-cosine kernels on the \emph{MNIST-8M} dataset \cite{loosli2007training}.
Due to the memory limit, following the doubly stochastic framework \cite{dai2014scalable}, we incorporate these random features based approaches under the data streaming setting for the reduction of time and space complexity.
The experimental setting on this dataset follows with \cite{dai2014scalable}: the feature dimension $d=784$ is reduced to $100$ by PCA; the number of random features $s$ is set to $4096$; the used Gaussian RBF kernel with kernel bandwidth $\varsigma$ equaling to four times the median pairwise distance; logistic regression with the regularization parameter $\lambda = 0.0005$ for this multi-class classification task; the batch size is set to be $2^{20}$ and feature block to be $2^{15}$.
Besides, we report the total time cost of each algorithm on generating feature mapping, training process and test process for evaluation.

Table~\ref{Tabultra} reports the approximation error, training error, test error, and the total time cost of each algorithm across the Gaussian kernel and the zero/first-order arc-cosine kernels under $s=4096$.
It can be found that, ORF/SORF and SSF achieve the best approximation performance on the Gaussian kernel, but ORF fails to significantly improve the approximation ability on arc-cosine kernels.
This is consistent with previous discussion on medium datasets in Section~\ref{app:exparc}.

\begin{table*}[t]
	 \centering
	\fontsize{8.3}{8}\selectfont
	\begin{threeparttable}
		\caption{Comparison results of various algorithms across three kernels in terms of training error (\%), classification error (\%) and total time cost (sec.) on the ultra-large \emph{MNIST 8M} dataset.}
		\label{Tabultra}
		\begin{tabular}{ccccccccccccccccccccc}
			\toprule
			&kernel &metric &RFF &QMC &ORF & SORF & ROM & Fastfood & SSF & GQ & LS-RFF   \cr
			\midrule
			&\multirow{4}{1.2cm}{\emph{Gaussian}}
			& approximation error    & 0.0126 & 0.0065 & 0.0041 &0.0041 & 0.0046 &0.0159  & 0.0078 & 0.0121 &0.0147  \\
			&&training error    & 0.22\% & 0.21\% &0.19\% &0.22\% & 0.19\% &0.21\%  & 0.20\% & 0.21\% & 0.22\% \\
			&&test error   & 0.99\% & 1.07\% & 1.11\% & 1.13\% & 0.99\% & 1.16\%  & 1.09\% & 1.16\% & 0.97\%  \\
			&&time cost (sec.)  & 13669 & 13999 & 14296 &14526 & 13497 & 14343  & 14872 & 14322 & 15725  \\
			\cmidrule(lr){3-12}
			&\multirow{4}{1.2cm}{\emph{arccos0}}
			& approximation error    & 0.0209 & 0.0124 & 0.0224 & 0.0231 & 0.0199 & 0.0246  & 0.0448 &0.0383 &0.0612 \\
			&&training error    & 2.71\% & 2.70\% & 2.70\% & 2.70\% & 2.70\% &2.60\%  & 2.70\% &3.02\% & 2.64\% \\
			&&test error   & 2.76\% & 2.91\% & 2.75\% & 2.86\% & 2.73\% &2.94\%  & 2.89\% &3.00\% & 2.72\%  \\
			&&time cost (sec.)   & 10577 & 10266 & 10501 & 10558 & 10595 &10807  & 11235 &10330 & 12231  \\
			\cmidrule(lr){3-12}
			&\multirow{4}{1.2cm}{\emph{arccos1}}
			& approximation error    & 0.0394 & 0.0104 &0.0310 &0.0316 & 0.0259 & 0.0458  & 0.0198 & 0.0369 & 0.0357 \\
			&&training error    & 0.93\% &0.96\% &0.94\% & 1.00\% & 0.94\% & 0.98\%  & 0.95\% & 0.96\% &0.93\%  \\
			&&test error   & 1.64\% & 1.59\% & 1.52\% &1.57\% &1.62\% &1.27\% &1.34\% & 1.51\% & 1.62\%  \\
			&&time cost (sec.)   & 9243.7 & 9170.3 & 9187.4 &8861.6 & 8870.8 &8824.1  &9455.3 &9188.1 & 9742.3  \\
			\bottomrule
		\end{tabular}
	\end{threeparttable}
\end{table*}

\vspace{-0.2cm}
\bibliographystyle{IEEEbib}
\bibliography{E:/Me/refs.bib}
\end{document}